%% file: main.tex
\documentclass[11pt,a4paper,logo]{googledeepmind}

\setleftlogo[120pt]{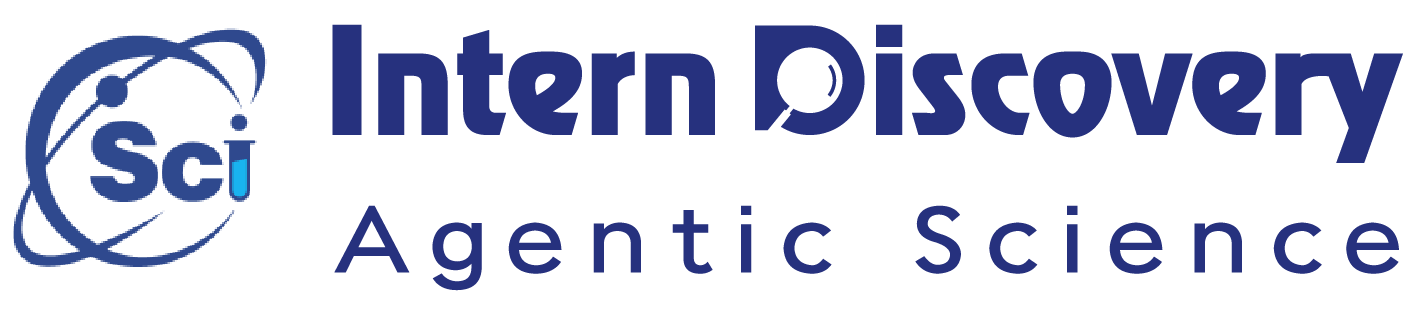} 
\setrightlogo[180pt]{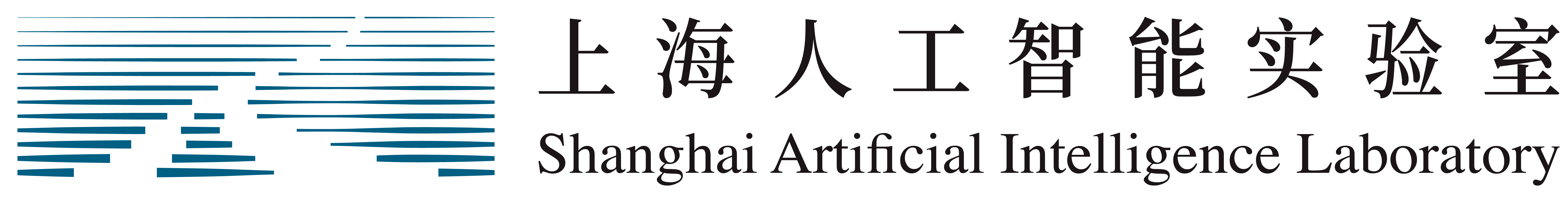}

\usepackage{placeins}
\usepackage{subcaption}

\usepackage[
    natbib=true,
    backend=biber,
    style=numeric, % Changed from 'alphabetic' to 'numeric' as per common technical report styles
    sorting=none % To keep references in order of appearance or as defined in .bib
]{biblatex}
\AtEveryBibitem{\clearfield{month}}
\AtEveryBibitem{\clearfield{day}}

\usepackage{csquotes}

\newcommand{\ProjectName}{InternAgent-1.5}

\newcommand{\github}{\raisebox{-1.5pt}{\includegraphics[height=1em]{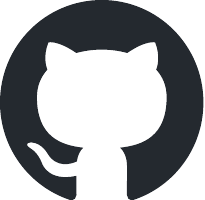}}}

\title{\ProjectName: A Unified Agentic Framework for Long-Horizon Autonomous Scientific Discovery}

\author[1]{InternScience Team, Shanghai Artificial Intelligence Laboratory \\
\github\ \texttt{\url{https://github.com/InternScience/InternAgent}}
}

\usepackage{pdflscape}

\usepackage{textcomp}
\usepackage{rotating}

\usepackage{setspace}
\usepackage{soul} 
\usepackage{multicol}
\usepackage{subcaption}
\usepackage{caption}

\sethlcolor{green!14}

\usepackage{array}
\usepackage{multirow}
\usepackage[justification=centering]{subcaption}
\usepackage{amsmath}
\usepackage{siunitx}

\usepackage{enumitem}
\usepackage{float}
\usepackage{seqsplit}
\usepackage{framed}

\usepackage{tikz}
\usepackage{listings}
\usepackage{colortbl}
\usepackage{wrapfig}
\usepackage[most]{tcolorbox}
\usepackage{pdfpages}

\tcbuselibrary{listingsutf8}
\definecolor{mycolor}{RGB}{50,80,150}

\usepackage{ragged2e}
\usepackage{makecell}
\usepackage{adjustbox}

\usepackage[symbol]{footmisc}
\newcolumntype{Y}{>{\RaggedRight\arraybackslash}X}
\newcommand{\cmark}{\ding{51}}
\newcommand{\xmark}{\ding{55}}

\hypersetup{
    colorlinks=true,
    linkcolor=blue, % Changed for better visibility of links
    citecolor=blue,  % Changed for better visibility of citations
    filecolor=black,
    urlcolor=blue    % Changed for better visibility of URLs
}

\usepackage{tocloft}
\usepackage{etoolbox}
\usepackage{arydshln}
\makeatletter
\patchcmd{\@tocline}
    {\hfil}
    {\leaders\hbox{\hfil}\hfil}
    {}{}
\makeatother

\begin{document}
\sloppy
\thispagestyle{firststyle}% 在文档开始就设置首页样式

\input{sections/abstract}

\maketitle

\tableofcontents
\newpage

\input{sections/introduction}
\input{sections/method}
\input{sections/experiment}
\input{sections/related}

\input{sections/conclusion}

\clearpage
\begingroup
\sloppy
\printbibliography[heading=bibintoc]
% \bibliography{references}
% \input{main.bbl}
\endgroup

\clearpage
\input{sections/appendix}

\end{document}

%% file: sections/abstract.tex
\begin{abstract}
Artificial intelligence is rapidly emerging as a powerful engine for scientific discovery. Modern machine learning and large language models support literature analysis, hypothesis generation, experimental planning, and data interpretation across biology, chemistry, and earth science. These advances have inspired AI Scientist systems that coordinate computational modeling, laboratory experimentation, and cross disciplinary reasoning to accelerate scientific progress.
However, existing AI Scientist systems remain limited by domain specific designs, incomplete reasoning abilities, naive optimization pipelines, and insufficient support for long horizon autonomous operation.
We introduce {\ProjectName}, a unified system designed for end-to-end scientific discovery across computational and empirical domains. The system is built on a structured architecture composed of three coordinated subsystems for generation, verification, and evolution. These subsystems are supported by foundational capabilities for deep research, solution optimization, and long horizon memory. The architecture allows {\ProjectName} to operate continuously across extended discovery cycles while maintaining coherent and improving behavior. It also enables the system to coordinate computational modeling and laboratory experimentation within a single unified system.
We evaluate {\ProjectName} on scientific reasoning benchmarks such as GAIA, HLE, GPQA, and FrontierScience, and the system achieves leading performance that demonstrates strong foundational capabilities. Beyond these benchmarks, we further assess two categories of discovery tasks. In algorithm discovery tasks, {\ProjectName} autonomously designs competitive methods for core machine learning problems. In empirical discovery tasks, it executes complete computational or wet lab experiments and produces scientific findings in earth, life, biological, and physical domains. Overall, these results show that {\ProjectName} provides a general and scalable framework for autonomous scientific discovery.

\end{abstract}

%% file: sections/introduction.tex
\definecolor{headergray}{RGB}{242, 242, 242}  % 表头灰
\definecolor{rowgray}{RGB}{250, 250, 250}     % 外部方法斑马纹
\definecolor{ia10color}{RGB}{242, 249, 255}   % InternAgent 1.0 (极淡蓝)
\definecolor{ia15color}{RGB}{225, 240, 255}   % InternAgent 1.5 (高亮蓝)

\section{Introduction}
\label{sec:introd}

\begin{table}[h]
    \centering
    \caption{\textbf{Comparison with state-of-the-art frameworks for autonomous scientific discovery.}}
    \label{tab:sota_compare}
    
    \renewcommand{\arraystretch}{1.4} 
    \setlength{\tabcolsep}{1.2mm}
    
    \newcommand{\logoimg}[1]{\includegraphics[height=9pt]{#1}\hspace{3pt}}
    \begin{tabular}{>{\raggedright\arraybackslash}p{4.5cm} || c c || c c c c}
        \hline
        \hline
        
        \multirow{2}{*}{\makecell{\textbf{Method}}} & \multicolumn{2}{c||}{\textbf{Domains}} & \multicolumn{4}{c}{\textbf{Capabilities}} \\
        \cline{2-7}
        & \makecell{\textbf{Algorithm}\\\textbf{Discovery}} & \makecell{\textbf{Empirical}\\\textbf{Discovery}} & \makecell{\textbf{Deep}\\\textbf{Research}} & \makecell{\textbf{Solution}\\\textbf{Refinement}} & \makecell{\textbf{Wet}\\\textbf{Lab}} & \makecell{\textbf{Persistence}\\\textbf{Running}} \\ 
        \hline
        \hline
        
        \logoimg{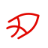} AI Scientist~\cite{lu2024ai, yamada2025ai} & \cmark & \xmark & \xmark & \cmark & \xmark & \xmark \\        
                
        \rowcolor[RGB]{250, 250, 250}
        \logoimg{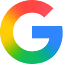} AlphaEvolve~\cite{novikov2025alphaevolve} & \cmark & \xmark & \xmark & \cmark & \xmark & \xmark \\
        
        \logoimg{imgs/logos/google_logo.png} AI Co-Scientist~\cite{gottweis2025towards} & \xmark & \cmark & \cmark & \xmark & \cmark & \xmark \\

        \rowcolor[RGB]{250, 250, 250}
        \logoimg{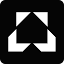} Robin~\cite{ghareeb2025robin} & \xmark & \cmark & \cmark & \xmark & \cmark & \xmark \\
        
        \logoimg{imgs/logos/futurehouse_logo.png} Kosmos~\cite{mitchener2025kosmos} & \xmark & \cmark & \cmark & \xmark & \cmark & \xmark \\
                
        \hline
        \hline
        
        \rowcolor[RGB]{242, 249, 255} 
        \logoimg{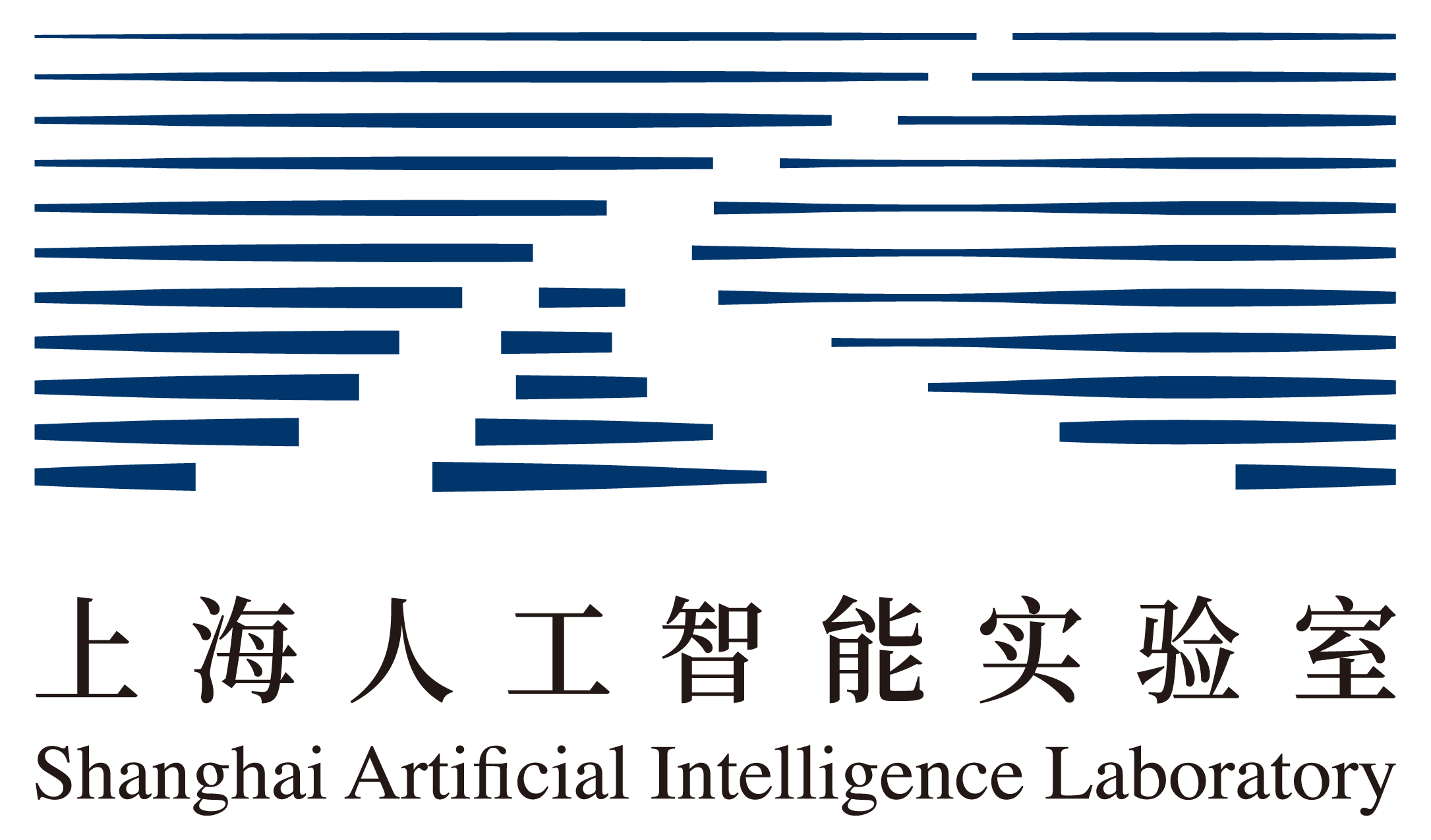} InternAgent 1.0~\cite{team2025novelseek} & \cmark & \xmark & \xmark & \xmark & \xmark & \xmark \\
        
        \rowcolor[RGB]{225, 240, 255} 
        \logoimg{imgs/logos/ailab_logo.png} \textbf{InternAgent 1.5} & \textbf{\cmark} & \textbf{\cmark} & \textbf{\cmark} & \textbf{\cmark} & \textbf{\cmark} & \textbf{\cmark} \\ 
        
        \hline
        \hline
    \end{tabular}
\end{table}

% [P1: The Rise of AI for Science]

Artificial intelligence is rapidly reshaping the landscape of scientific research, giving rise to the emerging paradigm of AI for Science ~\cite{wang2023scientific,van2023ai}. Recent progress in machine learning has driven advances across biology~\cite{abramson2024accurate,jumper2021highly,varadi2022alphafold}, chemistry~\cite{m2024augmenting,ghareeb2025robin}, and the physical and environmental sciences~\cite{guo2025earthlink}. Large language models have expanded this frontier by supporting literature analysis~\cite{jansen2023employing,yan2025surveyforge}, hypothesis generation~\cite{qi2023large,qi2024large,team2025novelseek}, experimental planning~\cite{team2025novelseek,du2025automlgen,hu2025flowsearch}, and data interpretation~\cite{yuan2025dolphin,bazgir2025agentichypothesis}. These capabilities have motivated a shift toward autonomous scientific systems capable of coordinating complex workflows that span computational modeling, wet‑lab experimentation, and cross-disciplinary reasoning.

% [P2: Related Work I]
A series of recent systems have demonstrated the potential of automated scientific agents. In algorithm optimization, AI Scientist~\cite{lu2024ai,yamada2025ai} and AlphaEvolve~\cite{novikov2025alphaevolve} integrate literature analysis, coding, and experimental evaluation into end-to-end research loops. In biomedicine, AI Co-Scientist~\cite{gottweis2025towards} generates hypotheses and designs therapeutic experiments. In chemistry, systems such as ChemCrow~\cite{m2024augmenting} and Robin~\cite{ghareeb2025robin} connect large language models with domain specific toolchains for synthesis planning and molecular design. In earth science, EarthLink~\cite{guo2025earthlink} integrates multisphere data and literature to support mechanism level reasoning. These systems have shown impressive domain-specific performance but operate as isolated verticals with architectures that embed strong domain assumptions. To move beyond single domain expertise, systems such as Kosmos~\cite{mitchener2025kosmos} introduce structured scientific world models to organize research across metabolomics, materials science, and genetics.

% [P4: Limitations and Motivation]
Despite substantial progress, current systems of AI4S exhibit several characteristics that limit their ability to support autonomous cross‑disciplinary discovery:
\begin{itemize}
    \item \textbf{Domain‑Specific Architectures:} Many systems are organized around discipline‑focused designs, which makes it difficult to perform unified reasoning across scientific fields.
    
    \item \textbf{Partial Foundational Capabilities:} Existing frameworks vary in their support for the use of heterogeneous dry‑lab and wet‑lab experiments, leading to uneven coverage of core scientific competencies.
    
    \item \textbf{Linear Optimization Pipelines:} Optimization procedures are often based on trajectory‑local updates and therefore do not integrate information across broader search processes when refining scientific proposals.
    
    \item \textbf{Limited Long‑Horizon Operation:} Most systems do not maintain persistent memory over extended research cycles, which restricts iterative refinement and long‑term autonomous operation.
\end{itemize}
A comparative overview of existing systems is presented in Table~\ref{tab:sota_compare}, which summarizes these characteristics across domains and foundational capabilities.

To address these challenges, we adopt an epistemological perspective grounded in the philosophy of science~\cite{bunge1967scientific,hey2009fourth} and categorize tasks into two fundamental domains: \textbf{Algorithm Discovery}, which \textit{transforms objectives into solutions in formal systems}, and \textbf{Empirical Discovery}, which \textit{transforms observations into generalizations about the physical world}. A framework capable of supporting both domains requires unified architectural principles, strong foundational capabilities, long‑horizon iterative optimization, and the ability to operate across computational and experimental environments.

\begin{figure*}[t]
    \centering
    
    \newlength{\imgw}
    \setlength{\imgw}{0.21\textwidth}  % 5 × 0.19 = 0.95，留点边距
    
    \begin{minipage}[t]{\imgw}
        \centering
        \includegraphics[width=\linewidth]{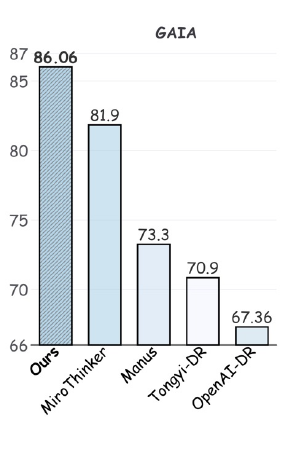}
    \end{minipage}\hspace{-5mm}
    \begin{minipage}[t]{\imgw}
        \centering
        \includegraphics[width=\linewidth]{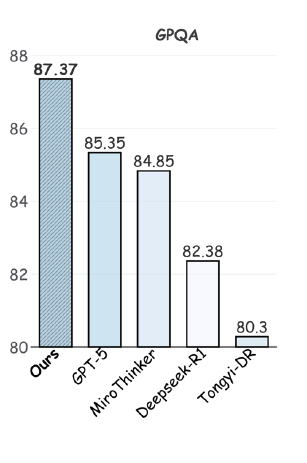}
    \end{minipage}\hspace{-5mm}
    \begin{minipage}[t]{\imgw}
        \centering
        \includegraphics[width=\linewidth]{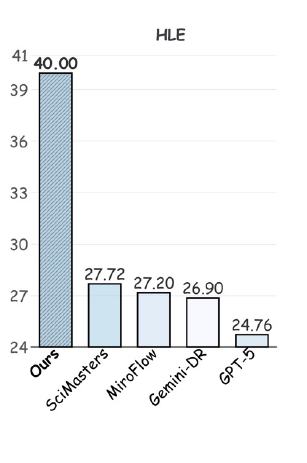}
    \end{minipage}\hspace{-5mm}
    \begin{minipage}[t]{\imgw}
        \centering
        \includegraphics[width=\linewidth]{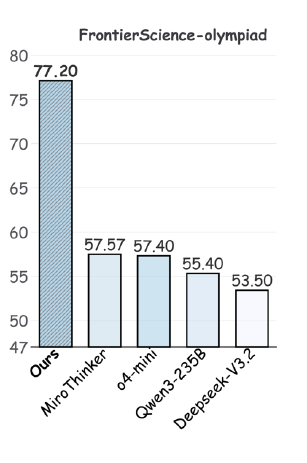}
    \end{minipage}\hspace{-5mm}
    \begin{minipage}[t]{\imgw}
        \centering
        \includegraphics[width=\linewidth]{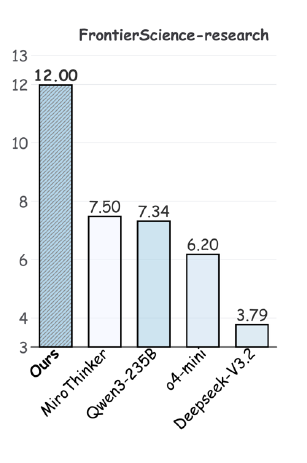}
    \end{minipage}
    \vspace{-20pt}
    \caption{Performance comparison of {\ProjectName} across GAIA~\citep{mialon2023gaia}, GPQA~\citep{rein2024gpqa}, HLE-full~\citep{phan2025humanity}, and FrontierScience~\citep{frontierscience}.}
    \label{fig:five_benchmarks}
\end{figure*}

\begin{figure}[t]
\vspace{-2pt}
\begin{center}
\includegraphics[width=0.95\textwidth]{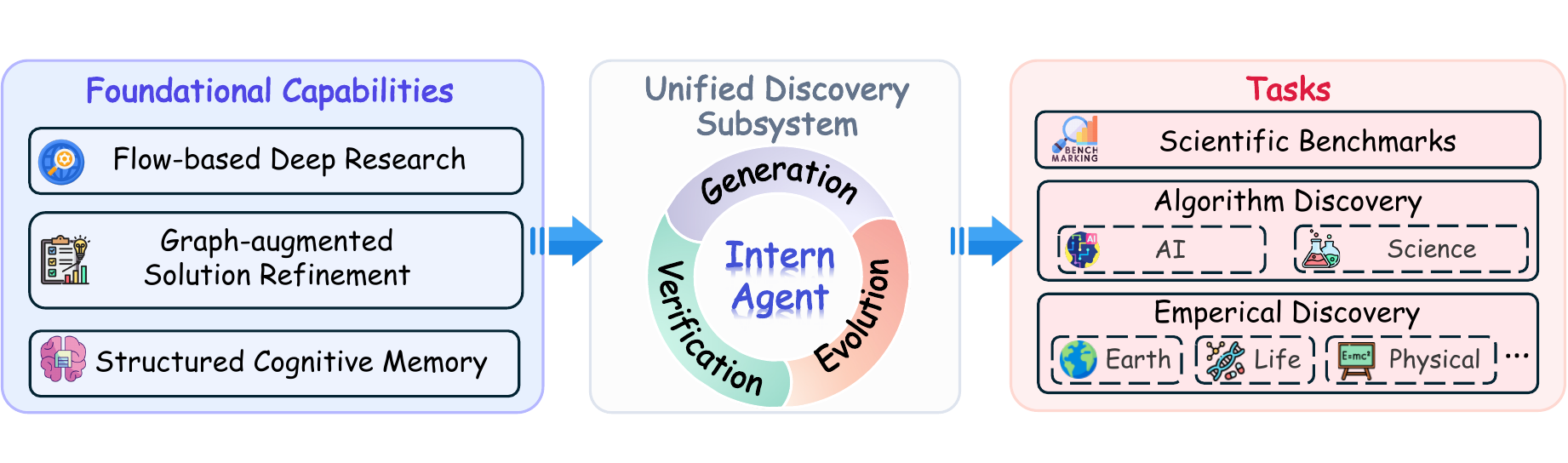} 
\end{center}
\vspace{-10pt}
\caption{Overview of {\ProjectName} that summarizes its foundational capabilities, unified discovery pipeline, and supported scientific tasks in a high‑level manner.}
\vspace{-4pt}
\label{img:fig_h-overview}
\end{figure}

% [P6: Our Work [IA15]]
Building on InternAgent 1.0~\citep{team2025novelseek}, we introduce {\ProjectName}, a unified system designed for end-to-end scientific discovery. The system follows the observation that scientific inquiry across domains can be organized into a common structure that includes literature based hypothesis construction, methodological evaluation, and evidence driven refinement. {\ProjectName} operationalizes these processes through three coordinated subsystems, namely \textbf{Generation}, \textbf{Verification}, and \textbf{Evolution}. Each subsystem is driven by a foundational capability: \textit{deep research} supports the \textbf{Generation} subsystem, \textit{solution refinement} supports the \textbf{Verification} subsystem, and \textit{long horizon memory} supports the \textbf{Evolution} subsystem. This design moves beyond structures restricted to single domain algorithm discovery and establishes a general framework suitable for both computational and empirical scientific tasks. A high level overview of {\ProjectName}, including its core capabilities, subsystem organization, and supported discovery tasks, is presented in Fig.~\ref{img:fig_h-overview}.

% [P7: Core Results]
{\ProjectName} is evaluated across standard benchmarks and open ended scientific discovery tasks. The system attains leading performance on agentic reasoning abilities, demonstrating the effectiveness of the foundational capabilities that drive the Generation and Verification subsystems. These capabilities, together with long horizon memory in the Evolution subsystem, support stable extended operation and enable consistent iterative refinement throughout long discovery cycles. Building on this capability structure, {\ProjectName} further performs competitively in both \textbf{algorithm discovery} and \textbf{empirical discovery} tasks, indicating that the unified framework scales from benchmark level reasoning to practical scientific workflows.

% [P8: Summary of Contributions] 
In summary, the main contributions of this work are as follows:

\begin{itemize}
    \item \textbf{A Unified Architecture for End-to-end Scientific Discovery:} {\ProjectName} organizes the scientific discovery process into three coherent subsystems for Generation, Verification, and Evolution. These subsystems support the full cycle of hypothesis formulation, methodological evaluation, and evidence driven refinement through foundational capabilities for deep research, solution refinement, and long horizon memory. This organization provides a robust basis for reliable and scalable scientific discovery.
    
    \item \textbf{State-of-the-Art Foundational Capabilities:} {\ProjectName} delivers strong foundational capabilities in deep research and solution refinement, supported by structured long horizon memory. Across benchmarks that measure cross disciplinary retrieval, structured reasoning, and scientifically grounded problem solving, the system achieves leading performance on HLE~\citep{phan2025humanity}, GAIA~\citep{mialon2023gaia}, GPQA~\citep{rein2024gpqa}, FrontierScience~\citep{frontierscience}, and SGI bench~\citep{xu2025probing}. These results confirm that the foundational capabilities of {\ProjectName} are sufficiently reliable to support complex scientific workflows.

    \item \textbf{Sustained Autonomous Optimization:} {\ProjectName} integrates a structured memory architecture with an iterative optimization process centered on the Verification and Evolution subsystems. This design supports the accumulation of contextual knowledge, the sustained refinement of hypotheses, and the stable improvement of methodological plans across many discovery cycles, moving toward scientific agents capable of extended self-improvement.
    
    \item \textbf{Breakthroughs in Algorithmic and Empirical Discovery:} {\ProjectName} demonstrates strong performance in both algorithm discovery and empirical scientific discovery. It produces competitive algorithmic solutions in domains such as reinforcement learning and test time methodology, and generates high quality outputs for data oriented scientific tasks. In empirical settings, the system executes complete experimental workflows and identifies new insights in fields that include biology and earth sciences. These results illustrate the generality and practical effectiveness of the framework across computational and physical scientific environments.
\end{itemize}

With these capabilities and results, we now present the design principles and technical foundations that enable {\ProjectName} to operate as a unified system for scientific discovery.

%% file: sections/method.tex
\section{{\ProjectName}}

\begin{figure}[h]
\begin{center}
\includegraphics[width=\textwidth]{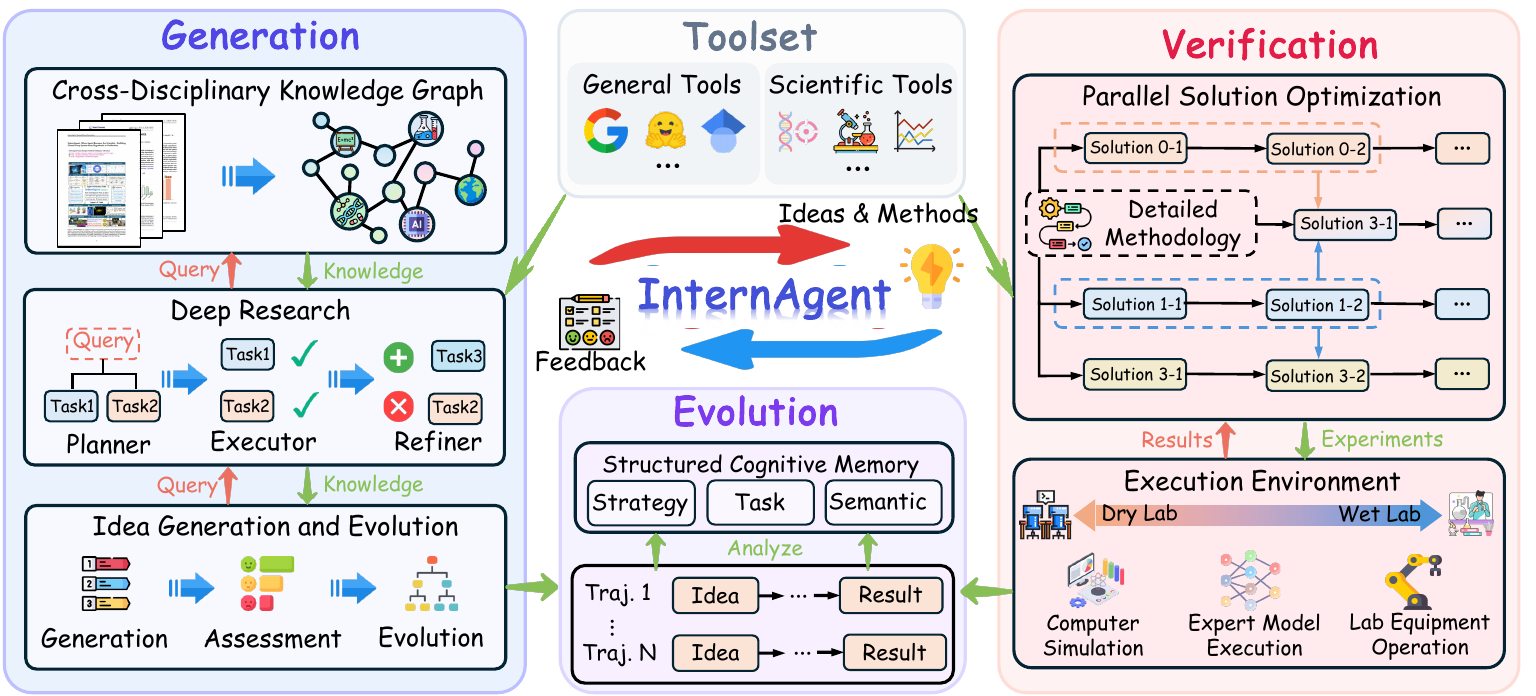} 
\end{center}
\vspace{-10pt}
\caption{Overview of {\ProjectName}, illustrating its unified scientific discovery pipeline organized around the Generation, Verification, and Evolution subsystems. The system operates through foundational capabilities for deep research, solution refinement, and long horizon memory, which together enable sustained autonomous scientific discovery.}
\label{img:framework}
\end{figure}
%%%%%%%%%%%%%%%%%%%%%%%%%%%%%%%%%%%%%%%%%%%%%%%%%%%%%%%%%%%%%%%%%%%%%%%%%%%%%%%%
\subsection{System Overview}

In this section, we present the system overview of {\ProjectName} as illustrated in Fig.~\ref{img:framework}. The system automates the full research cycle by integrating hypothesis formulation, methodological evaluation, and evidence driven refinement into a unified and continuously improving process. Its operation relies on foundational capabilities that support deep research, solution refinement, and long horizon memory. These capabilities are realized through agent driven reasoning and system level infrastructure and allow the system to maintain contextual continuity across iterations. With this capability structure, {\ProjectName} coordinates multiple subsystems to support autonomous, scalable, and sustained scientific discovery.

\subsubsection{Architecture}
The architecture of {\ProjectName} is organized into three core subsystems, namely the \textbf{Generation}, the \textbf{Verification}, and the \textbf{Evolution}. These subsystems form an integrated and iterative workflow. The Generation subsystem formulates hypotheses and methodological plans, the Verification subsystem evaluates these plans through computational or empirical procedures, and the Evolution subsystem incorporates the resulting evidence to update internal knowledge, strategies, and long term memory. This organization maintains a coherent flow of information and enables multi cycle autonomous operation.

\begin{itemize}
    \item \textbf{Generation:} The Generation subsystem constructs the conceptual foundation of scientific inquiry. It follows the generation and reflection paradigm of InternAgent 1.0~\citep{team2025novelseek} and is driven by the foundational capability of deep research. It conducts large scale literature analysis, scientific reasoning, and contextual integration and may invoke scientific tools when processing domain specific data. It produces structured hypotheses and methodological plans and records key reasoning traces for subsequent processing.

    \item \textbf{Verification:} The Verification subsystem evaluates the hypotheses and methodological plans produced by the Generation subsystem. Its operation is driven by the foundational capability of solution refinement, which structures the iterative search for improved procedures. It performs computational analyses, simulations, and laboratory style assessments as needed and uses historical information to guide evaluation choices. It supports parallel assessment of methodological variants and generates structured evidence for downstream refinement.

    \item \textbf{Evolution:} The Evolution subsystem refines system understanding and long term strategies based on outcomes from the Generation and Verification subsystem. It is driven by the foundational capability of memory and unifies analytical feedback with persistent knowledge management. It interprets verification results, identifies methodological limitations, updates procedural, episodic, and semantic information, and produces refined priors that guide subsequent cycles of the Generation and Verification subsystem.
\end{itemize}

These subsystems rely on the foundational capabilities introduced above in order to function coherently across extended discovery horizons. The next section presents these capabilities in detail and describes the technical methods that implement them.

\subsubsection{Foundational Capabilities}

The operation of {\ProjectName} relies on a set of foundational capabilities that allow the \textbf{Generation}, \textbf{Verification}, and \textbf{Evolution} subsystems to function coherently across extended discovery cycles. These capabilities support literature based hypothesis construction, methodological evaluation, iterative refinement, and long horizon continuity. They are implemented through the technical methods introduced in Sections \ref{sec:dr_method} to \ref{sec:memory_method} and provide the requirements for end-to-end scientific discovery.

\textbf{\textit{The first capability is deep research.}} It supports the Generation subsystem by enabling large scale retrieval, integration, and structuring of cross disciplinary scientific knowledge. Section \ref{sec:dr_method} introduces the search mechanisms and structured representations that realize this capability.

\textbf{\textit{The second capability is solution refinement.}} It supports the Verification subsystem by guiding the refinement of methodological plans and structuring the multi round search for improved procedures. Section \ref{sec:optim_method} presents the optimization strategies that implement this capability. Scientific tools may be invoked within this subsystem when computational or empirical assessment is required.

\textbf{\textit{The third capability is long horizon memory.}} It supports the Evolution subsystem by maintaining persistent storage and retrieval of contextual information, reasoning traces, and experimental outcomes. Section \ref{sec:memory_method} describes its structured organization and interaction rules.

Across these capabilities, {\ProjectName} maintains the continuity, adaptability, and scalability needed for reliable and continuously improving scientific discovery.

%%%%%%%%%%%%%%%%%%%%%%%%%%%%%%%%%%%%%%%%%%%%%%%%%%%%%%%%%%%%%%%%%%%%%%%%%%%%%%%%
% \subsection{Cross Disciplinary Deep Search and Graph Construction \\
% \normalsize \color{gray} \textbf{\textit{Deep Research Capability within the Generation Subsystem}}}
\subsection{Cross Disciplinary Graph Construction and Knowledge Capturing}
\vspace{-0.5em}
{\raggedleft \normalsize \color{gray} \textbf{\textit{Deep Research Capability within the Generation Subsystem}}\par}

\label{sec:dr_method}
To enable cross disciplinary knowledge construction and utilization, our design operates on both data and methodological levels. On the data side, the system integrates diverse scientific sources with the assistance of domain specific tools to parse, normalize, and structure scientific information into a large scale multidisciplinary knowledge graph. On the methodological side, it identifies relations and dependencies across domains through a structured extraction workflow that combines model driven analysis with tool assisted processing of specialized scientific data, enabling deep and effective cross disciplinary knowledge integration.

\subsubsection{Cross-Disciplinary Knowledge Graph}
To support accurate and comprehensive deep research, we maintain a cross-disciplinary knowledge graph (KG). Notably, it differs from traditional KGs \cite{DBLP:conf/acl/ChamiWJSRR20,DBLP:conf/kdd/CaoLWL24}, which represent knowledge as triples including simple entities and relations; our KG instead captures a richer set of scientific elements. 

\begin{figure}[t]
\vspace{-2pt}
\begin{center}
\includegraphics[width=\textwidth]{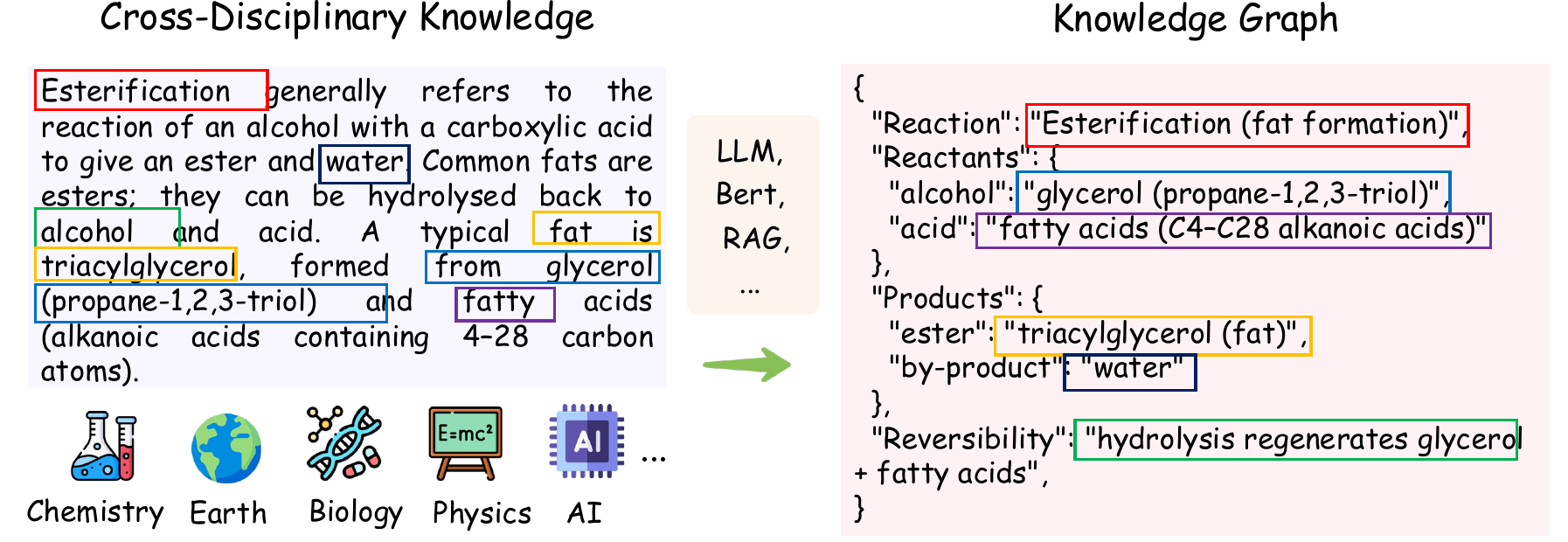} 
\end{center}
\vspace{-10pt}
\caption{The illustration for our cross-disciplinary knowledge graph.}\label{img:kg-illustration}
\vspace{-4pt}
\end{figure}

\paragraph{Graph Construction} 
From parsed outputs of papers, survey articles, technical reports, and domain notes, we construct a heterogeneous graph with nodes representing documents, key concepts, methods, datasets, empirical settings, and problem statements. The parsing process incorporates domain specific scientific tools to assist in identifying specialized entities and scientific attributes that are difficult to extract through general text analysis alone. Edges encode typed relations such as “cites’’ and “by product.’’ This design lets a single research idea sit at the intersection of multiple methodological and application communities and converts a flat corpus into a structured map where cross field dependencies emerge as paths rather than isolated points. For example, given the text
\textit{
``Esterification generally refers to the reaction of an alcohol with a carboxylic acid to give an ester and water. Common fats are esters; they can be hydrolysed back to alcohol and acid. A typical fat is triacylglycerol, formed from glycerol (propane 1,2,3 triol) and fatty acids (alkanoic acids containing 4–28 carbon atoms).’’}
the corresponding structured representation is shown in Fig.~\ref{img:kg-illustration}.

\begin{figure}[h]
\vspace{-2pt}
\begin{center}
\includegraphics[width=\textwidth]{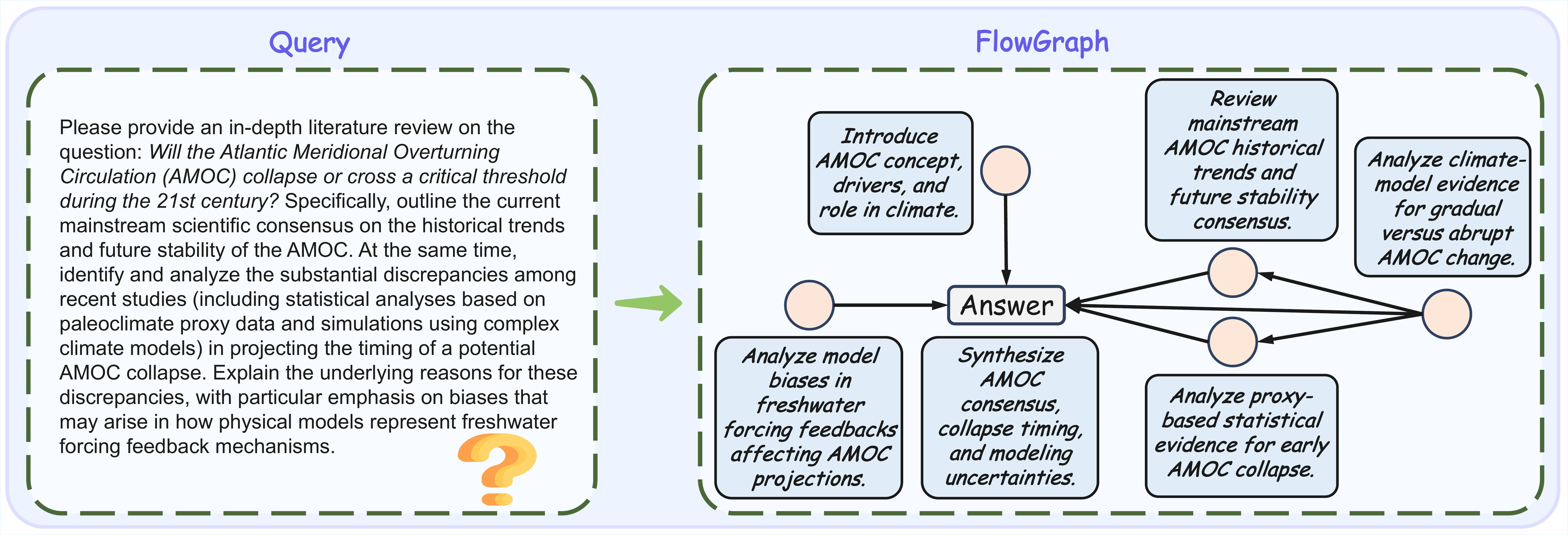} 
\end{center}
\vspace{-10pt}
\caption{The illustration for our flow graph.}\label{img:flow_graph}
\vspace{-4pt}
\end{figure}

\paragraph{Knowledge Extraction and Retrieval} We employ a schema-guided extraction workflow to build a knowledge graph from noisy, cross-domain text. First, candidate entities are identified via domain-agnostic named entity recognition and noun-phrase mining, and document-level co-citation and co-usage relations are used to establish initial concept links. A subsequent consolidation step refines node types and edge semantics, aligning textual evidence with citation evidence into a compact cross-disciplinary graph. To answer deep research queries, we integrate graph search with dense vector retrieval: graph search uncovers the nodes and paths that connect the query to relevant methods and domains, while dense retrieval captures semantically related items not directly linked in the graph. Finally, a ranking step merges these results and outputs path-structured evidence chains, which the deep research module then analyzes to reveal cross-disciplinary connections.

\subsubsection{Flow Graph}
In real scientific deep research tasks, knowledge often exhibits highly non-linear and dynamic dependencies. Conventional sequential research process struggle to capture these relationships effectively, which can lead to redundant information, over-reliance on early hypotheses, and inflexible reasoning processes. To address these challenges, we introduce Dynamic Structured Knowledge Flow as a core principle of deep research system, enabling systematic and adaptive organization of knowledge throughout the research process. Specifically, we capture the knowledge in research process as a directed acyclic graph (DAG) that explicitly represents tasks, subtasks, and their dependencies. 

\paragraph{Structured Knowledge Flow}
The research process is organized as a directed graph, which provides a structured
representation of the reasoning process. Formally, the research process is
defined as:
\begin{equation}
G = (V, E),
\end{equation}
where $V$ denotes the set of nodes and $E$ denotes the set of directed edges encoding
dependencies among nodes. Each node $v_i \in V$ corresponds to a subtask or a key conceptual unit arising during the reasoning process. To explicitly capture its functional role and execution status, each node is represented as a tuple:
\begin{equation}
v_i = (t_i, d_i, s_i, c_i),
\end{equation}
where $t_i \in \{\text{search}, \text{solve}, \text{answer}\}$ specifies the task type
associated with the node, $d_i$ describes the task content, $s_i$ tracks the execution
state of the node, and $c_i$ stores the resulting knowledge context upon successful
completion of the task. Directed edges in the graph encode structural dependencies or reasoning constraints between nodes. Specifically, each edge is defined as:
\begin{equation}
e_{ij} = (v_i, v_j, r_{ij}) \in E,
\end{equation}
where $e_{ij}$ indicates a directed relationship from node $v_i$ to node $v_j$, and $r_{ij}$
characterizes the type of dependency between the two nodes, such as \emph{requires
result from}, \emph{provides evidence for}, or \emph{constrains reasoning of}. These
relational edges ensure that information and intermediate results are propagated in a
dependency-aware manner throughout the reasoning graph.

\paragraph{Dynamic planning and refinement}
The knowledge flow is constructed incrementally: starting from a root query node, a planner identifies nodes that require further decomposition or context enrichment, generates successor nodes, and updates dependency edges accordingly. This iterative expansion continues until the flow sufficiently covers all subproblems necessary to address the research objective. This design not only enables efficient multi-agent collaboration but also supports adaptive refinement of the knowledge flow as new evidence emerges, ensuring coherent, systematic, and verifiable reasoning throughout the research process.

\subsubsection{Graph-Guided Output Synthesis}
Building upon the dynamically constructed knowledge flow, we describes how the abstract graph structure is instantiated into concrete research outputs. Once the planner completes the incremental construction of the flow graph, executable nodes are activated according to their dependency states and progressively populated with domain knowledge and intermediate reasoning results. Through iterative node execution, state updating, and context propagation along graph edges, the system refines the structured knowledge flow and synthesizes the final research answer.

\paragraph{Cross-Disciplinary Knowledge Collector.}
To facilitate cross disciplinary insight, the Knowledge Collector gathers information from a diverse set of sources across multiple domains. These sources include outputs obtained through scientific tools and remote resources accessed via the Science Context Protocol (SCP)~\citep{jiang2025scp}. By integrating multi domain knowledge, the system can uncover unexpected connections and inspire ideas that may not emerge within a single discipline. Executable nodes with satisfied dependencies are assigned to agents, which decompose each subtask into a sequential reasoning and information retrieval process, iteratively assembling the knowledge required to resolve it. After a node is executed, its state is updated and the resulting knowledge context is propagated to dependent nodes, ensuring that subsequent reasoning benefits from the most up to date and contextually rich information. This design enables structured, adaptive, and collaborative knowledge acquisition throughout the research process.

\paragraph{Reasoning capability enhancement}
We adopt a reasoning capability enhancement strategy that enables reasoning along multiple complementary pathways. For a given query, the model generates three forms of responses: a direct answer based solely on the input, a search augmented answer that incorporates evidence retrieved from external sources and scientific tools, and a self driven answer obtained through internal retrieval and refinement. These complementary outputs are aggregated to form the final response, balancing intrinsic reasoning, external evidence, and self consistency. This multi path strategy improves answer completeness and factual reliability while reducing reliance on any single reasoning pathway.

%%%%%%%%%%%%%%%%%%%%%%%%%%%%%%%%%%%%%%%%%%%%%%%%%%%%%%%%%%%%%%%%%%%%%%%%%%%%%%%%
% \subsection{Experiment Execution and Multi-round Parallel Optimization \\
% \normalsize \color{gray} \textbf{\textit{solution refinement Capability within the Verification Subsystem}}}
\subsection{Experiment Execution and Multi-round Parallel Optimization}
\vspace{-0.5em}
{\raggedleft \normalsize \color{gray} \textbf{\textit{Solution Refinement Capability within the Verification Subsystem}}\par}

\label{sec:optim_method}

% \textbf{[UNDER REORGANIZATION]: MLE-bench removed}
The transformation of a refined methodology into a verifiable scientific result requires an efficient and reliable validation loop. In both computational algorithm design and physical wet‑lab experimentation, the search space of possible configurations is extremely large, and linear trial‑and‑error procedures often converge prematurely. This section introduces the multi‑round parallel experiment optimization and execution framework, which enables {\ProjectName} to explore this space autonomously and progressively converge toward high‑quality experimental proposals.

\subsubsection{Generative Design for Experimental Optimization}

Efficiently exploring a large and unstructured design space is a central challenge in automated scientific experimentation. Traditional strategies \cite{aider, automind} based on linear refinement or tree‑structured search often face three fundamental limitations: %~\cite{xx}~\cite{xx}
\textbf{Isolated Trajectories} arise when insights discovered in one search path cannot inform parallel explorations.
\textbf{Unexploited Search History} occurs when informative patterns across longer trajectories are not captured or reused.
\textbf{Limited Idea Composition} restricts the integration of promising elements from different branches into improved solutions.

We formalize the experimental optimization problem as identifying the optimal solution within a search space $\mathcal{S}$, where each candidate solution $s \in \mathcal{S}$ represents a complete experimental proposal, including code logic, parameter configurations, and physical operation protocols. The objective is to find:
\begin{equation}
s^* = \arg\max_{s \in \mathcal{S}} h(\mathcal{T}, s),
\label{eq:optimization_objective}
\end{equation}
where $h(\mathcal{T}, s)$ denotes the evaluation metric of solution $s$ on a given task $\mathcal{T}$.

To address the limitations of conventional search, {\ProjectName} adopts a \textit{Graph‑Augmented Monte Carlo Search} framework. This approach preserves the exploration–exploitation balance of Monte Carlo Tree Search while replacing its rigid tree structure with a dynamic solution graph that aggregates information across all prior experiments. The search still follows the classical loop of selection, expansion, simulation, and backpropagation, but its effectiveness comes from a strengthened expansion phase powered by four graph‑based operators:
\begin{itemize}
    \item \textbf{Primary Expansion.} Generates a new proposal using only its immediate parent. It performs localized adjustments such as parameter refinement or correction of logical inconsistencies, creating the core backbone of parent and child edges used in credit assignment.
    
    \item \textbf{Intra‑branch Evolution.} Conditions proposal generation on both the parent and the historical trajectory of ancestors within the same branch. By analyzing recent successes and failures, it reinforces productive design changes and avoids repeatedly exploring unpromising strategies, formalizing a localized form of self‑reflection.
    
    \item \textbf{Cross‑branch Reference.} Introduces targeted transfer of design elements across different branches. When a branch stagnates, the system identifies high‑performing nodes elsewhere in the solution graph and uses them as references, allowing the new proposal to incorporate robust structural patterns or methodological modules discovered in parallel explorations.
    
    \item \textbf{Multi‑branch Aggregation.} Synthesizes complementary strengths from multiple top‑performing nodes across the solution graph. By decomposing these proposals into their essential components and recombining them, the operator produces hybrid designs that integrate successful ideas from previously independent search trajectories.
\end{itemize}

Once a new proposal is generated through one of these operators, it is executed in the corresponding environment, either a computational simulator or a physical experimental system, to obtain an evaluation score. This score is backpropagated through the proposal’s ancestral path to guide subsequent exploration. By integrating graph‑based information flow into the Monte Carlo search process, {\ProjectName} transforms experimental design into a collaborative and cumulative optimization pipeline, enabling rapid convergence toward high‑quality scientific solutions.

\subsubsection{Scenario}
\noindent \textbf{Code Opimization for Algorithm Discovery}
In algorithm discovery tasks, each proposal is an executable program specifying data‑processing steps, model components, and evaluation settings. The search module generates new variants by refining computational logic or integrating effective structures identified in other branches. Each candidate is executed in a controlled environment that compiles the code and evaluates its performance on benchmark datasets. Quantitative metrics such as accuracy, runtime, and resource usage are returned to the optimization module and backpropagated through the proposal’s lineage, enabling systematic refinement of algorithmic designs.

\noindent \textbf{Experimental Optimization for Empirical Discovery}
In empirical discovery tasks, each proposal specifies a full experimental procedure that may be executed either through computational simulation or on physical laboratory systems. The search process refines these procedures by adjusting parameters, modifying operational steps, or incorporating effective sub‑protocols identified across branches. When a proposal is simulated, domain models predict experimental outcomes such as reaction yield or protein stability. When it is executed physically, SCP \cite{jiang2025scp} coordinates automated devices to perform the protocol and collect measurements such as fluorescence intensity or assay signal quality. All quantitative results, whether simulated or physically measured, are returned to the optimization module for backpropagation, enabling iterative improvement of empirical workflows.

%%%%%%%%%%%%%%%%%%%%%%%%%%%%%%%%%%%%%%%%%%%%%%%%%%%%%%%%%%%%%%%%%%%%%%%%%%%%%%%%
% \subsection{Structured Cognitive Memory for Long Horizon Scientific Discovery \\
% \normalsize \color{gray} \textbf{\textit{Long Horizon Memory Capability within the Evolution Subsystem}}}
\subsection{Structured Cognitive Memory for Long Horizon Scientific Discovery}
\vspace{-0.5em}
{\raggedleft \normalsize \color{gray} \textit{\textbf{Long Horizon Memory Capability within the Evolution Subsystem}}\par}

\label{sec:memory_method}

\begin{figure}[h]
\vspace{-2pt}
\begin{center}
\includegraphics[width=\textwidth]{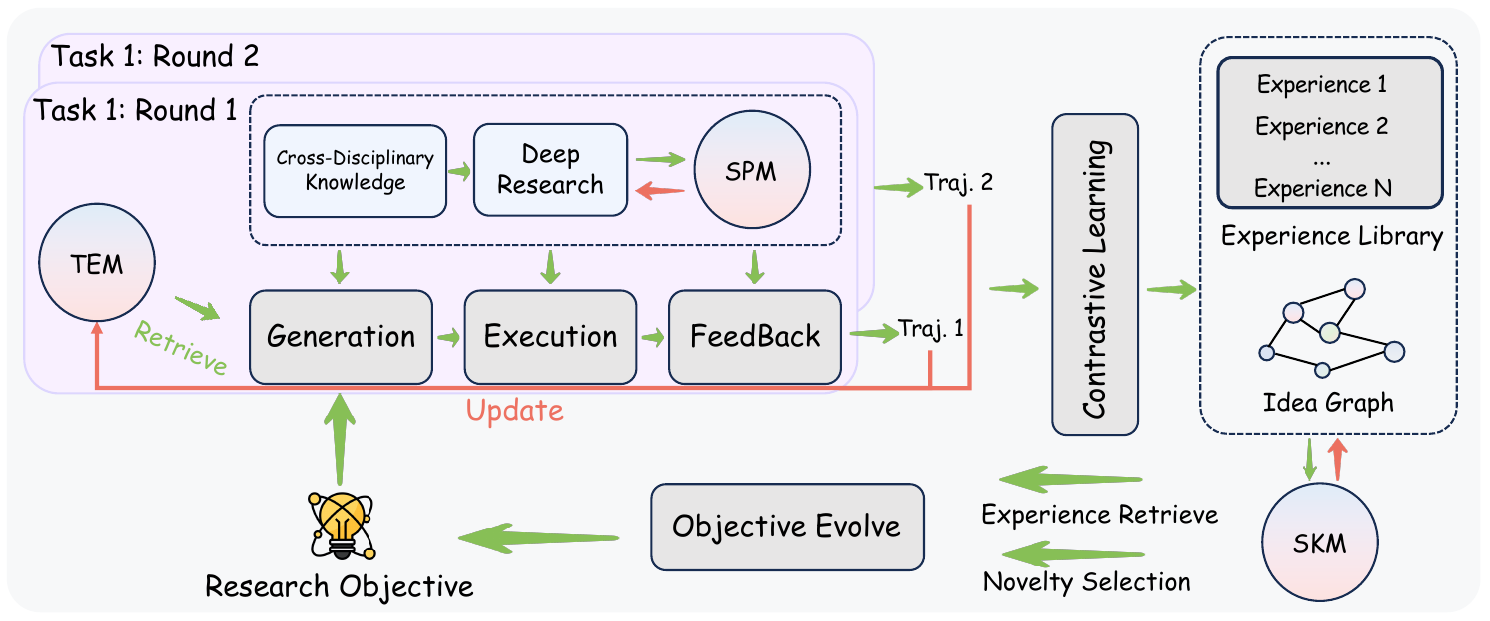} 
\end{center}
\vspace{-10pt}
\caption{The illustration for our Structured Cognitive Memory.}\label{img:mem_structure}
\vspace{-4pt}
\end{figure}

To support adaptive, efficient, and long horizon scientific discovery, {\ProjectName} incorporates a hierarchical memory subsystem referred to as Structured Cognitive Memory. This subsystem is engaged throughout the entire discovery loop and maintains continuity across cycles, allowing the agent to operate coherently over extended durations. It consolidates procedural, episodic, and semantic information into a unified structure that enables short term refinement, mid term adaptation, and long term conceptual development. The overall framework of Structured Cognitive Memory is shown in Fig.~\ref{img:mem_structure}.

\subsubsection{Strategy-Procedural Memory}
Strategy Procedural Memory (SPM) supports the deep research capability that InternAgent-1.5 invokes throughout the entire scientific discovery process whenever complex analytical reasoning is required. This capability involves integrating literature evidence, synthesizing contextual knowledge, constructing coherent multi-step reasoning plans, correcting failed strategies from earlier research workflows, and analyzing the root causes behind those failures. Instead of storing raw trajectories, the system distills reusable procedural structures from past reasoning processes, including both validated effective patterns and lessons learned from unsuccessful attempts. These procedural structures capture the key decision pivots, organizational patterns that have proven effective across earlier research workflows, as well as the diagnostic insights into failed strategies and their underlying reasons, serving as strategic priors that can be applied flexibly at different stages of the pipeline to avoid recurring pitfalls and optimize reasoning paths.

Given a historical trajectory $T$, SPM constructs a compact representation as follows:
\begin{equation}
\mathbf{z}_{T} = f_{\text{proc}}(T),
\end{equation}
which captures essential procedural states extracted from the full execution trace.  
When a new deep research query $q$ arrives, {\ProjectName} retrieves trajectories with procedurally aligned structures:
\begin{equation}
\mathcal{S}(q) =
\operatorname*{topk}_{T \in \mathcal{M}_{\text{SPM}}}
\mathrm{sim}\!\left( f_{\text{proc}}(q),\, \mathbf{z}_T \right).
\end{equation}
These retrieved strategic priors guide the planner toward globally coherent reasoning graphs, while constraining the executor to avoid redundant execution steps and unnecessary tool calls, thereby providing a stable and efficient procedural foundation for the downstream discovery process.

\subsubsection{Task-Episodic Memory}
Task Episodic Memory (TEM) supplies fine grained, within trajectory evidence that enables rapid adaptation during iterative experimentation. After each experiment, the system stores an episodic unit containing the attempted method $m$, extracted metrics $y$, and an improvement judgment. Each unit is encoded using a hybrid representation that combines semantic embeddings with sparse lexical features to support both conceptual and literal alignment.

During hypothesis refinement, relevant episodes are retrieved through the following formula:
\begin{equation}
\mathcal{R}(q) =
\operatorname*{topk}_{e \in \mathcal{E}}
\mathrm{sim}\!\left( f_{\text{enc}}(q),\, f_{\text{enc}}(e) \right),
\end{equation}
where $q$ denotes the current hypothesis.  
The retrieved episodes are injected directly into the generation context, helping the system avoid unsuccessful methodological directions, exploit successful ones, and refine hypotheses efficiently within each research trajectory.

\subsubsection{Semantic-Knowledge Memory}

Semantic Knowledge Memory (SKM) consolidates conceptual information across sessions and supports the long horizon evolution of research objectives. It consists of a Long Term Experience Library, which stores distilled methodological insights, and an Idea Graph that tracks the semantic topology of previously explored research directions.Specifically, upon the end of each experimental batch, the system employs a pairwise combination strategy for the generated methods to maximize the utilization of existing information. By leveraging contrastive learning between each combination according to their methods and experimental results, {\ProjectName} extracts both high-level methodological principles and low-level experimental heuristics to construct Long-term Experience Library.
Given a research goal $G$, long term knowledge entries are retrieved via the following formula:
\begin{equation}
\mathcal{K}(G) =
\operatorname*{topk}_{k \in \mathcal{L}}
\mathrm{sim}\!\left( f_{\text{enc}}(G),\, f_{\text{enc}}(k) \right).
\end{equation}

To promote continued exploration, each candidate objective $c$ is assigned a novelty score:
\begin{equation}
\mathrm{nov}(c) =
1 - \max_{x \in \mathcal{G}}
\mathrm{sim}\!\left( f_{\text{enc}}(c),\, f_{\text{enc}}(x) \right),
\end{equation}
which encourages the selection of objectives that extend beyond previously explored conceptual regions.  
In this way, SKM provides the semantic continuity and innovation pressure required for sustained multi-session scientific discovery.

%% file: sections/experiment.tex
\section{Experiments}
To evaluate {\ProjectName}'s capabilities in the full process of scientific discovery from multiple aspects, we verify its effectiveness through cross-disciplinary benchmarks, autonomous algorithm development, and scientific mechanism discovery, as elaborated in Sec.~\ref{sec:benchmarks},~\ref{sec:disc_algo}, and~\ref{sec:sci_mech}, respectively. In Sec.~\ref{sec:sci_mech},  {\ProjectName} demonstrates its applications in scenarios such as Earth Science~\ref{sec:earth_task}, Life Science~\ref{sec:life_science}, Biological Science~\ref{sec:bio_sci}, and Physical Science Tasks~\ref{sec:exp_chem}.

\subsection{Experiments Setup}
\label{sec:setup}

\subsubsection{General Scientific Reasoning Abilities}
\noindent \textbf{SGI-Bench~\citep{xu2025probing}.}
SGI-Bench is a scientist-aligned benchmark for Scientific General Intelligence (SGI), defined as the ability of an AI system to autonomously navigate the full scientific inquiry cycle of Deliberation, Conception, Action, and Perception. It operationalizes this goal with four task families spanning 10 scientific disciplines and 1,000+ expert-curated samples: Scientific Deep Research, Idea Generation, Dry/Wet Experiments, and Experimental Reasoning. Our results are reported on the DeepResearch and IdeaGeneration subsets.

\noindent \textbf{GAIA~\citep{mialon2023gaia}.}
GAIA is a benchmark of real-world tasks that require coordinated abilities in reasoning, multimodal understanding, web navigation, and tool use. We report results on its 165-question validation set.

\noindent \textbf{HLE~\citep{phan2025humanity}.}
Humanity’s Last Exam (HLE) is a large-scale multimodal benchmark of 2,500 expert-written questions covering mathematics, humanities, and the natural sciences. It is designed to probe frontier-level academic reasoning, where current LLMs still fall far short of human performance.

\noindent \textbf{Frontier Science~\citep{frontierscience}.}
FrontierScience is a benchmark that evaluates whether AI systems can perform expert‑level scientific tasks, including study design, data interpretation, and hypothesis assessment. Following the protocol in the original paper, our results are averaged over 20 runs on the Olympiad subset and 30 runs on the Research subset using its standard evaluation set.

\noindent \textbf{GPQA-diamond~\citep{rein2024gpqa}.}
GPQA is a collection of 448 expert-written multiple-choice questions in biology, chemistry, and physics, designed to test deep scientific reasoning rather than surface knowledge. We use its 198-question GPQA-diamond subset for evaluation.

\subsubsection{Algorithm Discovery}
\noindent \textbf{Scientific Algorithm.}
To validate the ability of {\ProjectName} to discover algorithms in scientific data domains and to demonstrate its improvements over InternAgent‑1.0~\cite{team2025novelseek}, we conducted experiments on six scientific data oriented algorithm discovery tasks. Notably, due to the limited capabilities of the coding agent in InternAgent‑1.0~\cite{team2025novelseek}, the baseline repositories for all tasks were first manually consolidated into single‑file implementations before being optimized by our system. In contrast, \textbf{\textit{{\ProjectName} supports an end‑to‑end algorithm optimization workflow directly on the original codebases.}}

\begin{itemize}
    \item \textbf{AutoRYP:} The AutoRYP task is built on the Suzuki–Miyaura reaction dataset containing 5,760 reaction entries~\citep{perera2018platform}. A LoRA‑finetuned LLaMA3‑8B embedding model~\citep{dubey2024llama} with an MLP predictor serves as the baseline. Model performance is assessed using the coefficient of determination (R²).
    \item \textbf{AutoTPPR:} The AutoTPPR task operates on the Perturb‑seq single‑cell transcription‑response dataset~\citep{norman2019exploring}. GEARS~\citep{roohani2024predicting}, a GNN‑ and MLP‑based framework for multi‑omics representation learning, is adopted as the baseline. The Top‑20 DE MSE is used as the evaluation metric.
    \item \textbf{AutoPower:} The AutoPower task relies on the IEEE 39‑Bus benchmark for power‑flow estimation~\citep{zimmerman2010matpower}. SenseFlow~\citep{zhao2025senseflow}, a physics‑informed self‑ensembling model, serves as the baseline method. Evaluation is performed using RMSE on PQ nodes.
    \item \textbf{AutoTSF:} The AutoTSF task is defined on the ETTh1 multivariate time‑series dataset from the ETT benchmark~\citep{haoyietal-informer-2021}. DLinear~\citep{zeng2023transformers}, an MLP‑based decomposition and forecasting model, is used as the baseline. Mean Absolute Error (MAE) averaged over horizons {96, 192, 336, 720} serves as the metric.
    \item \textbf{AutoMD:} The AutoMD task uses the MD17 dataset~\citep{chmiela2017machine}, which provides molecular energies and forces for seven small organic molecules. VisNet~\citep{ wang2024enhancing}, an equivariant geometry‑enhanced GNN, is adopted as the baseline. Force‑MAE is used as the evaluation metric.
    \item \textbf{AutoEAP:} The AutoEAP task is constructed from the UMI‑STARR‑seq enhancer‑activity dataset~\citep{arnold2013genome}. DeepSTARR~\citep{de2022deepstarr} provides the baseline for sequence‑based enhancer‑activity prediction. Housekeeper Pearson Correlation Coefficient (HK‑PCC) is used for evaluation.
\end{itemize}

\noindent \textbf{AI Algorithm.}
To further evaluate the capabilities of {\ProjectName} on AI algorithm discovery, we assembled a diverse suite of tasks that span model training pipelines, memory optimization strategies, reinforcement learning methods, and data processing routines, which collectively represent several of the most active directions in current AI algorithm research. \textbf{\textit{For each domain, we selected papers accepted by top AI conferences in 2025 as comparative baselines to validate whether {\ProjectName} can outperform the latest AI algorithms.}}

\begin{itemize}
    \item \textbf{AutoTTS:} The AutoTTS task is constructed on a benchmark evaluating test‑time scaling strategies for enhancing LLM reasoning. Atom ~\citep{tengAtomThoughtsMarkov2025a}, a Markov‑structured test‑time scaling framework that refines reasoning through iterative candidate exploration and denoising, serves as the baseline. Model performance is assessed using standard accuracy‑based reasoning metrics defined by the benchmark.
    \item \textbf{AutoMem:} The AutoMem task is defined on long‑term interaction and memory‑management benchmarks for LLM agents. A‑MEM~\citep{xuAMemAgenticMemory2025a}, an agentic memory system inspired by the Zettelkasten method and designed for dynamic note construction, semantic linking, and memory evolution, serves as the baseline. Evaluation focuses on long‑horizon agent performance using metrics such as retrieval accuracy, contextual relevance, and behavioral consistency under extended interaction.
    \item \textbf{AutoLM:} The AutoLM task examines self‑distillation based data synthesis for mathematical reasoning. For comparison, we implement a complete self‑distillation pipeline that performs synthetic question creation through few‑shot prompting, reasoning‑trajectory generation, and answer verification through majority voting. The resulting synthetic data are then used to train Intern-S1-mini~\citep{cai2024internlm2}. The evaluation measures the mathematical‑reasoning ability of the resulting model, using standard question‑answering accuracy as the primary metric.
    \item \textbf{AutoTTRL:} The AutoTTRL task is designed to autonomously discover reinforcement learning algorithms that do not require ground truth annotations on reasoning tasks (\textit{i.e.}, AIME 2024~\citep{li2024numinamath}). We employ Test-Time Reinforcement Learning (TTRL)~\citep{zuoTTRLTestTimeReinforcement2025a} as the baseline method, which utilizes a majority voting mechanism to provide effective reward estimation. Following TTRL, we generate 16 responses per question and calculate the average pass rate $\text{pass@1}=\frac{1}{k}\sum^{k}_{i=1}p_i$ for evaluation, where $p_i$ denotes the correctness of the $i$-th response.
\end{itemize}

\subsubsection{Empirical Discovery}

\noindent \textbf{Earth Science.}
To evaluate {\ProjectName} in the Earth Science domain, which involves petabyte-scale, multi-dimensional datasets and complex physical processes, we constructed two representative tasks:
\begin{itemize}
    \item \textbf{Automated Climate Diagnostics:} This task assesses the system's ability to integrate multi-source knowledge for data analysis. The benchmark utilizes historical Surface Air Temperature (TAS) data (1970–2010) from 20 Global Climate Models (GCMs) in the CMIP6 archive~\cite{cmip6} (including ACCESS-ESM1-5, CanESM5, etc.) and the ERA5 reanalysis dataset~\cite{era5} as the observational ground truth. The goal is to autonomously identify global warming trends and regional biases.
    \item \textbf{Climate Downscaling Optimization:} This task evaluates the system's ability to innovate scientific methods. The objective is to enhance surface-temperature fields over China from coarse-resolution NCEP-NCAR-R1 data ($2^\circ$)~\cite{ncep} to fine-scale ERA5 resolution ($0.25^\circ$). We employ standard Kriging interpolation~\cite{kriging} and linear Bias-Corrected Spatial Disaggregation (BCSD)~\cite{bcsd} as baselines to test if the system can autonomously design a superior deep-learning-based solution.
\end{itemize}

\noindent \textbf{Life Science.}
To demonstrate the broad applicability of {\ProjectName} in early-stage drug discovery, we evaluate its capabilities to therapeutic target identification, a domain characterized by heterogeneous multi-omics evidence, complex disease mechanisms, and strong requirements for mechanistic interpretability and experimental verifiability. We construct two representative target-discovery tasks that stress graph-structured planning, multi-modal tool orchestration, and iterative reflection:

\begin{itemize}
    \item \textbf{Automated Biological Evidence Synthesis:} The agent orchestrates end-to-end analyzes (expression, genomic alteration, survival, pathway, and tractability) by integrating resources such as TCGA~\cite{TCGA}, OpenTargets~\cite{opentargets} and KEGG~\cite{kegg}, and synthesizes a coherent evidence chain. We reproduce OriGene’s discovery of \textit{GPR160} as an HCC target.
    \item \textbf{Hypothesis Generation and Target Prioritization:} The agent constructs a multi-modal evidence graph (cohorts, proteomics, annotations, pathways, and literature) and iteratively refines mechanistic hypotheses to rank actionable candidates. We reproduce the identification of \textit{ARG2} as a mechanistically grounded target in CRC, together with experiment-ready validation suggestions.
\end{itemize}

\noindent \textbf{Biological Sciences.}
As a key capability in the Biological Sciences domain, the fluorescent‑protein engineering task targets the improvement of existing fluorescent proteins for imaging applications. The system identifies the parent sequence and relevant structural context through literature‑driven analysis, then performs dry‑lab design by combining sequence inspection, folding prediction with ESMFold \cite{rives2019biological}, and mutational‑effect evaluation using sequence–function and stability predictors such as ProSST \cite{li2024prosst} to generate candidate variants. These designs are translated into wet‑lab protocols through SCP \cite{jiang2025scp}, which coordinates automated DNA assembly, expression, purification, and fluorescence‑intensity measurement. The resulting data are analyzed and fed back into the design layer, producing a structured experimental report that integrates predictions, protocols, and measured performance.

\noindent \textbf{Physical Science.}
To evaluate {\ProjectName} in chemical synthesis and drug discovery, we define two tasks requiring the integration of physical constraints and structural design:
\begin{itemize}
    \item \textbf{Automated Reaction Outcome Prediction:} Evaluated on the ChemCoTBench~\citep{li2025beyond} forward prediction dataset, this task requires the agent to predict both the target major product and stoichiometric by-products. The system must analyze reactant properties and strictly apply atomic conservation logic. To ensure rigorous evaluation, we explicitly revised 26 problematic entries in this benchmark, providing a corrected ground truth for synthesis planning.
    \item \textbf{Generative Scaffold Hopping:} This task aims to discover novel molecular entities that circumvent patent barriers while preserving bioactivity. The agent is tasked with replacing the core scaffold of a molecule while maintaining key 3D shape and electrostatic features. The system must employ generative algorithms to propose bioisosteres and filter candidates based on medicinal chemistry metrics, such as Synthetic Accessibility and LogP, to ensure the proposed analogs are viable drug candidates.
\end{itemize}

\subsection{Evaluating Agentic Reasoning Abilities} 
\label{sec:benchmarks}

\begin{table}[t]
\vspace{-6pt}
    \centering
    \caption{Performance comparison on SGI-bench. The best results are \textbf{bolded} and the second best results are \underline{underlined}.}
    \vspace{-6pt}
    \label{tab:sgi-bench}
    \setlength{\tabcolsep}{6mm}
    \renewcommand{\arraystretch}{1.2}
    \begin{tabular}{l c c}
        \toprule
        \textbf{Method} & \textbf{Deep Research} & \textbf{Idea Generation} \\
        \midrule
        Gemini-3-pro~\citep{gemini-3-pro} & \underline{18.48} & 39.68 \\
        GPT-5~\citep{chatgpt2025} & 14.47 & \underline{55.40} \\
Claude-Sonnet-4.5~\citep{claude-sonnet-45} & 13.84 & 43.20 \\
Qwen3-Max~\citep{yang2025qwen3} & 15.38 & 39.83 \\
o4-mini~\citep{o4-mini} & 11.95 & 40.78 \\
        \rowcolor{ia15color} \textbf{{\ProjectName} (Gemini-3-pro+o4-mini)} & \textbf{37.74} & \textbf{58.11} \\
        \bottomrule
    \end{tabular}
\end{table}

\begin{table}[t]
\centering
\caption{Performance comparison on GAIA and HLE benchmarks. The best results are \textbf{bolded} and the second best results are \underline{underlined}. Results not reported in the original papers are denoted as `` - ".}
\label{tab:main_results}
\vspace{-6pt}

\setlength{\tabcolsep}{3.2pt}
\resizebox{1.0\linewidth}{!}{
\begin{tabular}{cccccccc}
\toprule
             &     & \multicolumn{4}{c}{\textbf{GAIA val}}                            & \multicolumn{2}{c}{\textbf{HLE}}                                   \\ 
\cmidrule{3-8}
\multicolumn{1}{l}{\textbf{Agent}}     &\multicolumn{1}{c}{\textbf{Base Model}}               & Level 1 & Level 2 & Level 3 & \multicolumn{1}{c}{Avg.}& Text only & All \\ 

% \rowcolor{pink!20}
% \multicolumn{7}{c}{\textit{\textbf{No Agency}}}                                                                                                                                                                         \\ \hline
% \multicolumn{1}{c|}{Qwen-3-8B}            &    11.32     &   2.32      &   0.00      &   \multicolumn{1}{c|}{4.85}       & -                                                    &  -   \\
% \multicolumn{1}{c|}{Qwen3-32B}            &     13.21    &   3.49      &    3.84     & \multicolumn{1}{c|}{6.67}      &      -                                               &  -   \\
% \multicolumn{1}{c|}{Qwen3-235B}            &    15.09     &   3.49      &     3.84    & \multicolumn{1}{c|}{7.27}      &   9.18                                                  & 8.60    \\
% \multicolumn{1}{c|}{Intern-S1}            &28.30    &   9.30      &    7.69     & \multicolumn{1}{c|}{15.15}      & 8.90                                                    & 8.30    \\
% \multicolumn{1}{c|}{Deepseek-R1}          &   33.96      &   13.95      &   3.84      & \multicolumn{1}{c|}{18.78}        & 8.60                                                    &-     \\
% \multicolumn{1}{c|}{o4-mini}              &   28.30      &   12.79      &    7.69     & \multicolumn{1}{c|}{16.97}   &14.50                                                     & 14.28    \\
% \multicolumn{1}{c|}{GPT-5}                &  -       &   -      &     -    & \multicolumn{1}{c|}{-}    & 25.85                                                    & 24.76    \\ \hline
% \rowcolor{orange!10}  
\midrule
% \rowcolor{blue!10}
\multicolumn{8}{l}{\textit{\textbf{React Model with Tools}}}                                                                                                                                                            \\ \midrule

\multicolumn{1}{l}{WebDancer~\citep{wu2025webdancer}}   &\multicolumn{1}{c}{QwQ-32B}           & 61.5        & 50.0        & 25.0       & \multicolumn{1}{c}{51.5}     &  -                                                   & -    \\
\multicolumn{1}{l}{WebShaper~\citep{tao2025webshaper}}  &\multicolumn{1}{c}{Qwen2.5-72B}           & 69.2        & 63.4       &  16.6      & \multicolumn{1}{c}{60.1}       &  -                                                  & -    \\
\multicolumn{1}{l}{MiroThinker~\citep{team2025mirothinker-v1-5}} &\multicolumn{1}{c}{MiroThinker-v1.5-30B}              & -        & -        &  -       & \multicolumn{1}{c}{72.0}     &  31.0                                         & \\ 
\multicolumn{1}{l}{MiroThinker~\citep{team2025mirothinker-v1-5}} &\multicolumn{1}{c}{MiroThinker-v1.5-235B}              & -        & -        &  -       & \multicolumn{1}{c}{80.8}     &  \underline{39.2}                                         &  -  \\ 
\multicolumn{1}{l}{Tongyi-DR~\citep{tongyidr}} &\multicolumn{1}{c}{Tongyi-DR-30B}              & -        & -        &  -       & \multicolumn{1}{c}{70.9}     &  32.9                                          & - \\ \midrule
\multicolumn{8}{l}{\textit{\textbf{DeepResearch Agents}}}    \\ \midrule
\multicolumn{1}{l}{OpenAI DR~\citep{openai2025deepresearch}}     &\multicolumn{1}{c}{-}        & 74.29        &  69.06       &  47.60       & \multicolumn{1}{l}{67.36}    & -                                                    & 26.60    \\
\multicolumn{1}{l}{ChatGPT-Agent~\citep{chatgpt2025}}     &\multicolumn{1}{c}{-}        & -       &  -      &  -       & \multicolumn{1}{l}{-}    & -                                                   & \textbf{41.60}  \\
\multicolumn{1}{l}{Kimi-Researcher~\citep{moonshot_kimi_researcher_2025}}     &\multicolumn{1}{c}{-}        & -       &  -      &  -       & \multicolumn{1}{l}{-}    & -                                                  & 26.90   \\
\multicolumn{1}{l}{Manus~\citep{manus2025}}    &\multicolumn{1}{c}{-}             & \underline{86.50}        & \underline{70.10}        & \underline{57.70}        & \multicolumn{1}{c}{73.30}    &-                                                     & -    \\
\multicolumn{1}{l}{Gemini DR~\citep{google2024geminidr}} &\multicolumn{1}{c}{-}   & -        & -        &   -      & \multicolumn{1}{c}{-}    & -                                                    & 26.90    \\    
\multicolumn{1}{l}{OWL~\citep{hu2025owl}}     &\multicolumn{1}{c}{Gemini 2.5 Pro}               & 84.90        & 68.60        & 42.30        & \multicolumn{1}{c}{69.70}     &  -                                                   & -    \\
\midrule
\multicolumn{8}{l}{\textit{\textbf{Our Method}}}  \\ \midrule
\multicolumn{1}{l}{{\ProjectName} }  &\multicolumn{1}{c}{Qwen3-235B}       &    69.81     &   60.47      &     30.77    & \multicolumn{1}{c}{58.79}      & 15.04                                                    &14.84     \\ 

\multicolumn{1}{l}{{\ProjectName}} &\multicolumn{1}{c}{o4-mini}        &  88.68     &   81.39      &  61.54       & \multicolumn{1}{c}{80.61}     & 36.10                                                    & 34.52    \\ 
\rowcolor{ia15color}
\multicolumn{1}{l}{{\ProjectName}} & \multicolumn{1}{c}{Gemini-3-pro+o4-mini} & \textbf{92.45} & \textbf{89.53} & \textbf{61.54} & \multicolumn{1}{c}{\textbf{86.06}} & \textbf{40.87} & \underline{40.00} \\

\bottomrule
\end{tabular}
}
\end{table}

\noindent \textbf{SGI-Bench.} As shown in Table~\ref{tab:sgi-bench}, {\ProjectName} (Gemini-3-pro+o4-mini) achieves the best performance on two SGI-Bench tracks, Deep Research and Idea Generation, significantly outperforming strong frontier models. On Deep Research track, {\ProjectName} reaches 37.74\%, surpassing the second-best Gemini-3-pro 18.48\%) by a large margin (+19.26\%). On Idea Generation, {\ProjectName} attains 58.11\%, exceeding the prior best GPT-5 55.40\% (+2.71\%). Overall, these results suggest that {\ProjectName}’s iterative deep-research workflow that integrate structured planning, targeted information gathering, and refinement yields substantial gains in evidence-driven research capability while also improving creative yet grounded idea generation.

\begin{table}[t]
\vspace{-6pt}
\centering
\caption{Domain-wise performance comparison on the Humanity's Last Exam (HLE). The best results are \textbf{bolded} and the second best results are \underline{underlined}.}
\label{tab:hle_results}
\vspace{-6pt}

\setlength{\tabcolsep}{3.2pt}
\resizebox{1.0\linewidth}{!}{
\begin{tabular}{llccccccccc}
\toprule
 &  & \multicolumn{9}{c}{\textbf{Humanity's Last Exam}} \\
\cmidrule{3-11}
\textbf{Setting} & \textbf{Model}
& Math & Bio/Med & CS/AI & Physics & Human. & Chem. & Engineer. & Other & Avg. \\
\midrule

\multirow{3}{*}{\textbf{Text-Only}}
& Deepseek-R1~\cite{guo2025deepseek}
& 9.30 & 8.60 & 7.40 & 5.80 & 11.00 & 5.60 & 10.30 & 7.50 & 8.60 \\

& Gemini-3-pro-preview~\cite{gemini-3-pro}
& \underline{45.08} & \underline{26.13} & \underline{26.79} & \underline{32.67}
& \underline{44.04} & \textbf{34.65} & \textbf{29.69}
& \underline{32.39} & \underline{38.00} \\

\rowcolor{ia15color}
& \textbf{InternAgent-1.5}
& \textbf{48.96} & \textbf{30.63} & \textbf{29.46} & \textbf{34.16}
& \textbf{44.56} & \underline{30.69} & \underline{28.13}
& \textbf{37.50} & \textbf{40.87} \\
\midrule

\multirow{4}{*}{\textbf{All-Set}}
& o4-mini~\cite{o4-mini}
& 19.00 & 11.40 & 12.90 & 12.60 & 9.10 & 12.70 & 12.60 & 6.90 & 14.30 \\

& GPT-5~\cite{chatgpt2025}
& 31.00 & 22.10 & 24.90 & 21.70 & 20.60 & 16.40 & 14.40 & 18.00 & 24.80 \\

& Gemini-3-pro-preview~\cite{gemini-3-pro}
& \underline{44.76} & \underline{27.14} & \underline{29.05} & \underline{31.30}
& \textbf{42.92} & \textbf{40.00}
& \textbf{32.43} & \underline{34.33} & \underline{38.04} \\

\rowcolor{ia15color}
& \textbf{InternAgent-1.5}
& \textbf{48.09} & \textbf{30.36} & \textbf{30.71} & \textbf{33.04}
& \underline{42.47} & \underline{34.55}
& \underline{30.63} & \textbf{38.63} & \textbf{40.00} \\
\bottomrule
\end{tabular}
}
\end{table}

% %%%%%%%%%%%%%%% FrontierScience %%%%%%%%%%%%%%%

\begin{table}[t]
\centering
\caption{Performance comparison on FrontierScience of olympiad and research tasks across bio, chem, and phy domains. The best results are \textbf{bolded} and the second best results are \underline{underlined}.}
\label{tab:performance_comparison_olympiad_research}
\vspace{-6pt}

\setlength{\tabcolsep}{3.2pt}
\resizebox{1.0\linewidth}{!}{
\begin{tabular}{lcccccccc}
\toprule
& \multicolumn{4}{c}{\textbf{Olympiad (avg N=20)}} & \multicolumn{4}{c}{\textbf{Research (avg N=30)}} \\
\cmidrule(lr){2-5} \cmidrule(lr){6-9}
\textbf{Method} & \textbf{Bio} & \textbf{Chem} & \textbf{Phy} & \textbf{All} & \textbf{Bio} & \textbf{Chem} & \textbf{Phy} & \textbf{All} \\
\midrule
o4-mini~\cite{o4-mini} & \textbf{47.00±14.90} & 65.00±6.40 & 53.40±4.50 & 57.40±3.30 & 9.67±5.47 & 8.17±4.37 & 0.83±2.27 & 6.20±2.54 \\
InternS1-235B~\cite{bai2025intern} & 17.00±12.69 & 52.88±4.05 & 50.40±3.88 & 48.05±2.84 & 4.50±4.35 & 11.00±3.74 & 2.67±3.35 & 6.06±2.30 \\
Mirothinker-v1.5-30B-A3B~\cite{team2025mirothinker-v1-5} & 22.86±4.52 & 69.64±7.49 & 54.86±3.18 & 57.57±3.66 & 8.17±6.39 & 8.50±6.21 & 5.83±4.10 & 7.50±3.77 \\
DeepSeek-V3.2-Thinking~\cite{liu2025deepseek-v3-2} & 26.50±7.26 & 72.25±3.25 & 66.30±2.63 & 64.70±2.41 & 2.50±3.10 & 16.33±4.64 & 1.40±2.70 & 6.84±1.88 \\
Qwen3-235B-A22B-Thinking~\cite{yang2025qwen3} & 24.00±9.17 & 61.13±6.05 & 57.10±4.79 & 55.40±3.68 & 10.17±5.08 & 10.00±6.32 & 1.58±2.41 & 7.34±3.37 \\
Qwen3-30B-A3B-Thinking~\cite{yang2025qwen3} & 13.50±9.10 & 47.25±4.47 & 42.70±3.65 & 41.60±2.94 & 1.50±2.93 & 2.00±3.32 & 0.70±1.79 & 1.41±1.52 \\
\midrule
\multicolumn{9}{l}{\textit{\textbf{Our Method}}} \\
\midrule
\rowcolor{ia15color}
\textbf{InternAgent-1.5} & \underline{46.00±8.00} & \textbf{85.50±3.67} & \textbf{76.80±2.99} & \textbf{77.20±3.06} & \textbf{10.33±4.64} & \textbf{22.00±6.00} & \textbf{3.67±2.87} & \textbf{12.00±2.49} \\
\bottomrule
\end{tabular}
}
\end{table}

\begin{table}[t]
\vspace{-2pt}
\centering
\small
\caption{Performance comparison on GPQA-diamond benchmark. The best results are \textbf{bolded} and the second best results are \underline{underlined}. Results not reported in the original papers are denoted as `` - ".}
\label{tab:gpqa_results}
\vspace{-6pt}

\setlength{\tabcolsep}{8pt}
% \resizebox{1.0\linewidth}{!}{
\renewcommand{\arraystretch}{0.85}
\begin{tabular}{ccccc}
\toprule
             & \multicolumn{4}{c}{\textbf{GPQA-diamond}}                                                       \\ 
\cmidrule{2-5}
\multicolumn{1}{l}{\textbf{Agent}}               & Bio & Chem & Phys & \multicolumn{1}{c}{Avg.}\\ 

\midrule
% \rowcolor{blue!10}
% \multicolumn{5}{l}{\textit{\textbf{React Model with Tools}}}      
\multicolumn{5}{l}{\textit{\textbf{Base Models}}}   \\ \midrule
\multicolumn{1}{l}{Qwen-3-8B~\cite{yang2025qwen3}}     &  -   &  -    &  -    & \multicolumn{1}{c}{44.44}     \\
\multicolumn{1}{l}{Qwen3-32B~\cite{yang2025qwen3}}     &   -  &  -    &  -    & \multicolumn{1}{c}{49.49}   \\
\multicolumn{1}{l}{Qwen3-235B~\cite{yang2025qwen3}}    & -    &  -    &  -    & \multicolumn{1}{c}{47.47}   \\
\multicolumn{1}{l}{Intern-S1~\cite{bai2025intern}}     &  \textbf{89.47}   &  59.49    &  93.02    & \multicolumn{1}{c}{78.26}  \\
\multicolumn{1}{l}{Deepseek-R1~\cite{guo2025deepseek}} &  63.16   &  \underline{76.34}    &  91.86  & 82.32  \\
\multicolumn{1}{l}{o4-mini~\cite{o4-mini}}             &  78.95   &  63.44    &  94.19    & \multicolumn{1}{c}{78.28} \\
\multicolumn{1}{l}{GPT-5~\cite{chatgpt2025}}           & \underline{84.21}    & \underline{76.34}     & \underline{95.35}     & \multicolumn{1}{c}{\underline{85.35}}   \\ \midrule
\multicolumn{5}{l}{\textit{\textbf{React Model with Tools}}}   \\ \midrule
\multicolumn{1}{l}{WebShaper~\cite{tao2025webshaper}}             & 47.37    &  52.69    &  81.40 & \multicolumn{1}{c}{64.65}   \\
\multicolumn{1}{l}{MiroThinker~\citep{2025mirothinker}}            &\underline{84.21}     & 75.27   &  \underline{95.35} &\multicolumn{1}{c}{84.85}    \\
\multicolumn{1}{l}{Tongyi DR~\cite{tongyidr}}         & 78.95     & 67.74  &  \underline{95.35} & \multicolumn{1}{c}{80.30}   \\

\midrule
\multicolumn{5}{l}{\textit{\textbf{Our Method}}}  \\ \midrule
\rowcolor{ia15color}
\multicolumn{1}{l}{InternAgent-1.5}         & \underline{84.21}     & \textbf{79.57}  &  \textbf{96.51} & \multicolumn{1}{c}{\textbf{87.37}}   \\

\bottomrule
\end{tabular}
% }
\end{table}

\noindent \textbf{GAIA.} As shown in Table~\ref{tab:main_results}, {\ProjectName} outperforms both closed-source Manus (73.30\%) and leading open-source agentic models Mirothinker (80.8\%) and Tongyi-DR (70.9\%), even though they are specifically trained and evaluated only on the GAIA text-only subset. \ProjectName~ also shows strong robustness on Level 3 questions (61.54\%). These results indicate that \ProjectName's iterative workflow combining knowledge planning, collection, and refinement is particularly effective for multi-hop and compositional reasoning.

\noindent \textbf{HLE.} As shown in Table~\ref{tab:main_results} and \ref{tab:hle_results}, \ProjectName~ achieves the best overall performance among all compared systems. It reaches 40.87\% accuracy in the text-only setting and 40.00\% on the full benchmark, outperforming strong baselines such as Gemini-3-pro-preview and GPT-5. The improvements are consistent across most HLE sub-domains, highlighting the robustness of \ProjectName~ on long-horizon, cross-disciplinary reasoning tasks.

\noindent \textbf{FrontierScience.} Table~\ref{tab:performance_comparison_olympiad_research} compares the performance of various methods on Olympiad and Research tasks across biology, chemistry, and physics domains.  {\ProjectName} achieves the best overall results in both Olympiad (77.20\%) and Research (12.00\%) settings, with particularly strong performance in Chemistry and Physics. It outperforms all baselines, including DeepSeek-V3.2-Thinking (64.70\% Olympiad) and Mirothinker-v1.5 (7.50\% Research), demonstrating its superiority in both structured problem-solving and open-ended scientific reasoning. 

\noindent \textbf{GPQA.} As shown in Table~\ref{tab:gpqa_results}, \ProjectName~ achieves state-of-the-art performance on the GPQA-diamond benchmark with an average accuracy of 87.37\%. It outperforms both strong base models and prior tool-augmented agents, with particularly strong results in Chemistry and Physics. These results demonstrate the effectiveness of our method for expert-level scientific reasoning.

\subsection{Results for Algorithm Discovery Tasks}
\label{sec:disc_algo}

\subsubsection{Scientific Algorithm}
\label{sec:sci_algo}

We evaluated {\ProjectName} across six scientific domains and compared it against our previous work~\citep{team2025novelseek,yuan2025dolphin}, and state-of-the-art domain-specific baselines. As summarized in Table~\ref{tab:performance_comparison_sci}, {\ProjectName} consistently achieves superior performance across all tasks and demonstrates the efficacy of our architectural improvements.

\noindent \textbf{Chemical and Molecular Analysis.} 
In the domain of chemical synthesis, the model demonstrates a strong capacity to interpret structured reaction information. For the AutoRYP task on the Suzuki-Miyaura dataset, {\ProjectName} achieves an $R^2$ of 36.6. This result significantly outperforms the LoRA finetuned LLaMA3 baseline of 27.6 and the Dolphin score of 31.8. Similarly, for the AutoMD task regarding Molecular Dynamics, our model effectively captures geometric features. It reduces the Energy MAE to 0.114 compared to 0.158 achieved by the equivariant GNN baseline VisNet.

\noindent \textbf{Physics and Engineering Systems.} 
Our framework exhibits robust performance in modeling complex physical systems and temporal dependencies. In the AutoPower task for Power Flow Estimation, {\ProjectName} achieves an RMSE of 0.00318 on the IEEE 39-Bus dataset. This surpasses the physics informed SenseFlow model score of 0.00473. For the AutoTSF task involving Time Series Forecasting, the DLinear baseline proves to be a strong competitor with an MAE of 0.4382. Our method further reduces the error to 0.423 and demonstrates effective handling of multivariate trends in the ETTh1 dataset.

\noindent \textbf{Biological and Genomic Prediction.} 
The most substantial improvements are observed in computational biology tasks. In the AutoTPPR task for Transcription Prediction, the model achieves an MSE of 0.143. This outperforms the GNN based GEARS framework score of 0.197. Notably, in the AutoEAP task for Enhancer Activity Prediction, {\ProjectName} reaches a Pearson Correlation Coefficient of 0.91. This constitutes a drastic improvement over the DeepSTARR baseline of 0.65 and highlights the exceptional ability of the agent to map DNA sequences to quantitative activity levels.

\begin{table}[tbp]
\centering
\renewcommand{\arraystretch}{1.2}
\caption{Performance comparison across six types of scientific algorithm tasks. }
\vspace{-6pt}
\label{tab:performance_comparison_sci}
\resizebox{0.97\textwidth}{!}{
\begin{tabular}{l@{\hspace{0.5em}}cccccc}
\toprule
& \multicolumn{6}{c}{\textbf{Tasks}} \\
\cmidrule(lr){2-7}
\textbf{Method} & \textbf{AutoRYP} & \textbf{AutoTPPR} & \textbf{AutoPower} & \textbf{AutoTSF} & \textbf{AutoMD} & \textbf{AutoEAP} \\
\cmidrule(lr){2-7}
& \textbf{$R^2$} & \textbf{MSE} & \textbf{RMSE} & \textbf{MAE} & \textbf{Energy-MAE} & \textbf{HK-PCC}  \\
\midrule
Baseline & 27.6 & 0.197 & 0.00473 & 0.438 & 0.158 & 0.65 \\
Dolphin~\citep{yuan2025dolphin} & 31.8 & 0.173 & 0.00455 & 0.463 & 0.152 & 0.76  \\
\textbf{InternAgent-1.0~\citep{team2025novelseek}}  & 35.4 & 0.146 &  0.00426 &  0.433 & 0.148 & 0.79 \\
\rowcolor{ia15color} \textbf{InternAgent-1.5}  & \textbf{36.6} & \textbf{0.143} & \textbf{0.00318} & \textbf{0.423} & \textbf{0.114} & \textbf{0.91} \\
\bottomrule
\end{tabular}
}
\end{table}

\begin{table}[t]
\centering
\renewcommand{\arraystretch}{1.0}
\caption{Results on complicated scientific algorithm design tasks. \textbf{Note that} our previous version Dolphin~\citep{yuan2025dolphin}  and InternAgent-1.0~\citep{team2025novelseek} cannot address these complicated tasks listed in the table.}
\vspace{-6pt}
\label{tab:performance_comparison_ai}
\resizebox{0.80\textwidth}{!}{
\begin{tabular}{l@{\hspace{0.5em}}cccc}
\toprule
& \multicolumn{4}{c}{\textbf{Tasks}} \\
\cmidrule(lr){2-5}
\textbf{Method} & \textbf{AutoTTS} & \textbf{AutoMem} & \textbf{AutoTTRL} & \textbf{AutoLM} \\
\cmidrule(lr){2-5}
& \textbf{Acc} & \textbf{F1} & \textbf{Acc} & \textbf{Acc} \\
\midrule
Baseline & 70.9  & 0.2338±0.3452 & 23.3 & 0.880 \\
\rowcolor{ia15color} \textbf{InternAgent 1.5} & \textbf{72.5} & \textbf{0.2785±0.3643} & \textbf{23.9} & \textbf{0.904} \\
\bottomrule
\end{tabular}
}
\end{table}

\subsubsection{AI Algorithm} 
\label{sec:ai_algo}

\noindent \textbf{Test-time Scaling.} On the MMLU-CF dataset~\citep{zhao2025mmlu}, we evaluate the reasoning capability using the architecture proposed by~\citep{tengAtomThoughtsMarkov2025a}. Our approach attains a score of 72.5, exceeding the baseline score of 70.9. This improvement indicates that \ProjectName effectively enhances performance in knowledge-intensive tasks, demonstrating robust reasoning capabilities and superior adaptability in complex domain scenarios.

\noindent \textbf{Memory Mechanism.} On the Locomo dataset~\citep{maharana2024evaluating}, we evaluate under the AutoMem setup using Qwen2.5‑3B~\citep{yang2025qwen3} as the base model to ensure alignment with the A‑MEM~\citep{xuAMemAgenticMemory2025a} baseline. Using F1 as the evaluation metric, our approach attains an F1 of 0.2785, exceeding the baseline score of 0.2338. This improvement indicates that the proposed memory architecture and interaction mechanism enable more reliable long‑horizon retention, retrieval, and integration of accumulated information.

\noindent \textbf{Reinforcement Learning.} 
On the AutoRL task, we evaluate our approach across the reinforcement‑learning control and decision‑making benchmarks used in prior work. Using the same environments and return‑based metrics as the TTRL baseline, our method achieves consistently higher returns and improved training stability. These results indicate that the proposed framework provides more effective trajectory refinement and decision‑making guidance across diverse RL settings.

\noindent \textbf{Large Language Model.} 
On the AutoLM task, we apply the full self distillation pipeline and fine tune Intern-S1-mini~\citep{bai2025intern} on the synthesized mathematical reasoning data. To validate algorithmic performance under a minimal system configuration, all experiments are conducted with a 16k token context. We evaluate our approach on the MATH500 reasoning benchmark used in prior work. Accuracy improves from 0.880 to 0.904, indicating that the enhanced self distillation pipeline produces higher quality trajectories and verified answers, providing effective supervision for strengthening the model’s mathematical reasoning ability.

\subsection{Discoveries of Scientific Mechanism}
\label{sec:sci_mech}

\subsubsection{Earth Science}
\label{sec:earth_task}

\begin{figure}[H]
    \centering
    \includegraphics[width=0.98\textwidth]{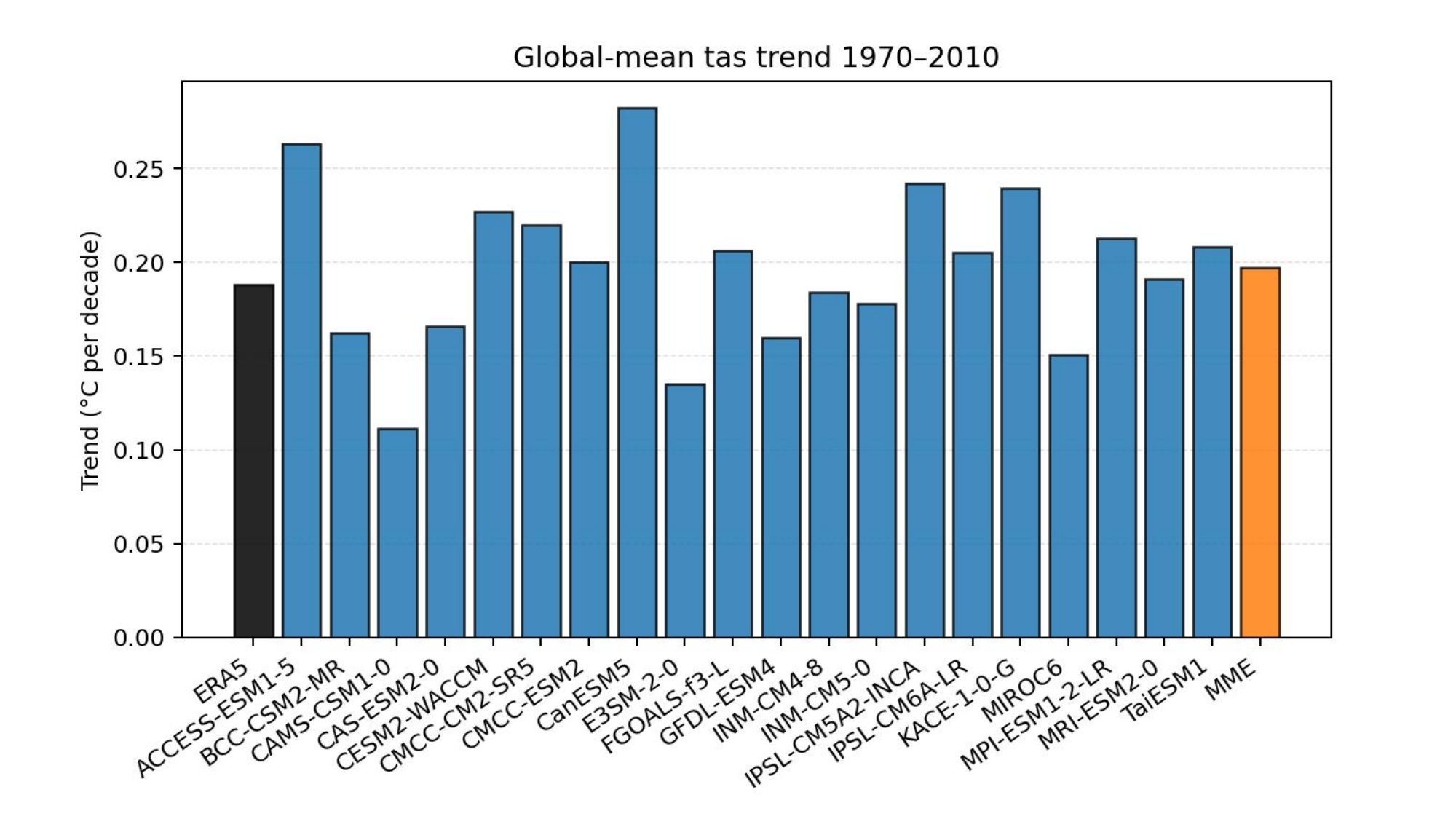}
    \vspace{-6pt}
    \caption{Automated Climate Analysis - Temperature Trends. {\ProjectName} autonomously generated this diagnostic for 20 CMIP6 models against ERA5 (1970-2010), showing the ranked bar chart of global-mean temperature trends ($^\circ$C/decade).}
    \label{fig:earth_viz_trends}
\end{figure}

Building on the setup described in Sec.~\ref{sec:setup}, we demonstrate how {\ProjectName} addresses the dual challenges of knowledge integration and high-fidelity modeling in Earth Science.

\noindent\textbf{Automated Data Analysis and Knowledge Integration.}
In the \textit{Automated Climate Diagnostics} task, the system was prompted to evaluate CMIP6 climate model simulations against the ERA5 reanalysis. Rather than simply calculating statistics, {\ProjectName}  employed its \textit{Knowledge Flow Planner} to integrate climate modeling literature and physical reasoning. Guided by widely adopted diagnostic conventions, the system selected key evaluation metrics, including global-mean surface temperature trends ($^\circ$C~decade$^{-1}$) and model–observation biases, and constructed an end-to-end analysis pipeline encompassing data retrieval, temporal alignment, and statistical estimation.

\begin{figure}[H]
    \centering
    \begin{subfigure}[b]{0.98\textwidth}
        \centering
        \includegraphics[width=\textwidth]{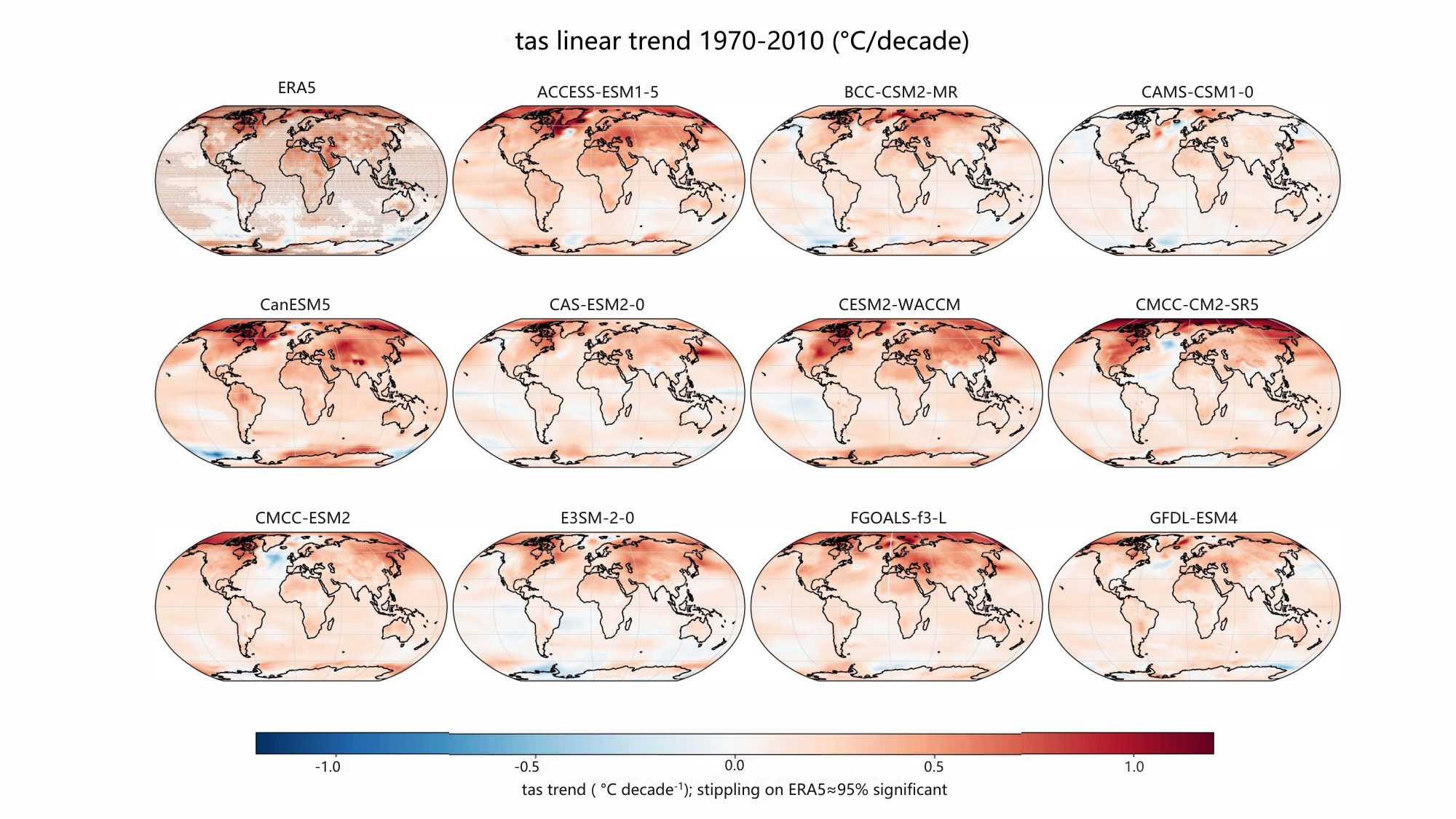}
        \caption{}
        \label{fig:earth_viz_map_a}
    \end{subfigure}
    \vspace{2ex}
    \begin{subfigure}[b]{0.80\textwidth}
        \centering
        \includegraphics[width=\textwidth]{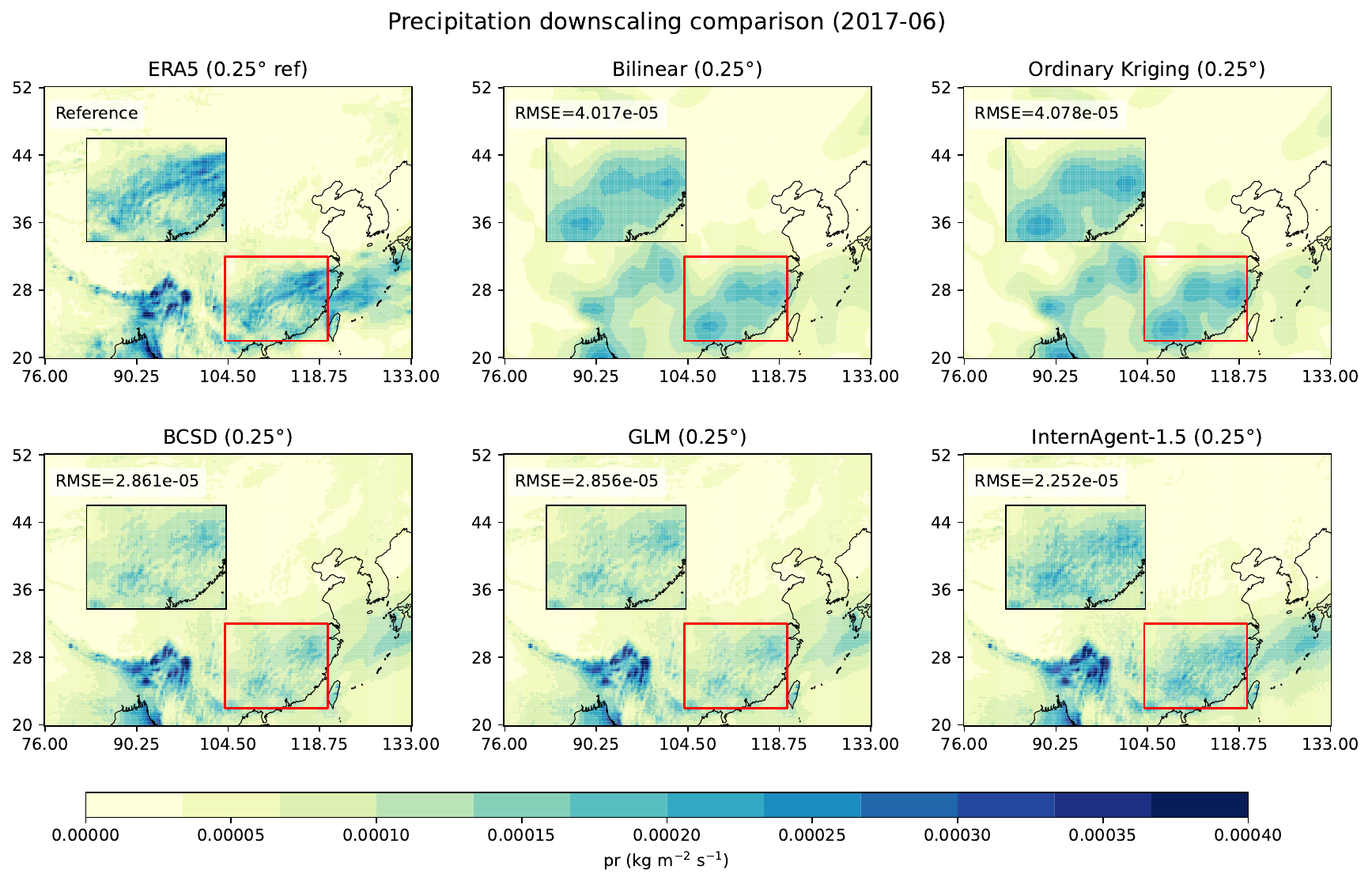}
        \caption{}
        \label{fig:earth_viz_map_b}
    \end{subfigure}
    \caption{(a) Automated Climate Analysis - Spatial Patterns. Spatial maps of linear temperature trends generated by {\ProjectName}, identifying regional warming patterns across different CMIP6 models. (b) Precipitation downscaling compariso across different methods.}
    \label{fig:earth_viz_maps}
\end{figure}

The system successfully processed the multi model ensemble and generated a ranked bar chart in Fig.~\ref{fig:earth_viz_trends} that contextualizes model performance. To further assess the physical consistency of the simulated trends, {\ProjectName} also produced spatial maps of linear temperature change in Fig.~\ref{fig:earth_viz_maps}, which enable interpretation at the regional scale. These diagnostics reveal canonical large scale warming patterns, including enhanced high latitude trends that match established characteristics of observed and simulated climate change. Taken together, the results show that {\ProjectName} supports climate analysis not only by automating computation but also by organizing diagnostics in a manner that aligns with domain specific interpretability.

\noindent\textbf{Algorithmic Innovation and Optimization.} 
For the \textit{Climate Downscaling Optimization} task, {\ProjectName} addressed known limitations of widely used baseline methods, including Kriging~\citep{kriging} and BCSD~\citep{bcsd}, which can struggle to represent non-stationary biases and fine-scale spatial variability in surface temperature fields.

The system autonomously conducted a literature review, recognizing that standard linear assumptions are insufficient for non-stationary biases. It proposed a deep-learning-based approach designed to capture complex spatial dependencies, generated the implementation code, and refined the architecture through iterative validation. 
As shown in Fig.~\ref{fig:earth_viz_maps} and summarized in Table \ref{tab:downscaling}, the model optimized by {\ProjectName} achieves improved performance relative to established statistical baselines. While the baseline Kriging and BCSD methods yielded RMSEs of 3.1658 and 0.9049 respectively, our system's solution reduced the RMSE to \textbf{0.8488}.

Beyond improvements in bulk error statistics, qualitative comparison of spatial fields indicates that bilinear interpolation and kriging introduce substantial spatial smoothing and attenuate high-intensity precipitation cores. In contrast, the {\ProjectName} more faithfully reproduces fine-scale spatial gradients and localized convective maxima present in the ERA5 reference data. This suggests that the model effectively captures nonlinear and scale-interactive processes that are not resolved by conventional interpolation or stationary bias-correction methods. Collectively, these results confirms that {\ProjectName} can independently conceive and optimize new scientific tools rather than merely applying existing ones.

\begin{table}[h]
    \centering
    \caption{\textbf{Performance Comparison on Climate Downscaling Task.} The deep learning method proposed and implemented by {\ProjectName} outperforms both traditional interpolation and statistical correction baselines in reconstructing high-resolution ($0.25^\circ$) surface temperature fields.}
    \label{tab:downscaling}
    \setlength{\tabcolsep}{6mm}
    \renewcommand{\arraystretch}{1.2}
    \begin{tabular}{l c c}
        \toprule
        \textbf{Method} & \textbf{Type} & \textbf{RMSE} \\
        \midrule
        Kriging Interpolation & Traditional Spatial & 3.1658 \\
        BCSD & Statistical Correction & 0.9049 \\
        \rowcolor{ia15color} \textbf{{\ProjectName}} & \textbf{AI-Optimized Deep Learning} & \textbf{0.8488} \\
        \bottomrule
    \end{tabular}
\end{table}

\subsubsection{Life Science}
\label{sec:life_science}

We present two representative case studies to illustrate how {\ProjectName} supports therapeutic target discovery in realistic biomedical scenarios.

\noindent\textbf{Automated Biological Evidence Synthesis.}
As a case study, we reproduced the discovery of \textit{GPR160} as a novel therapeutic target in hepatocellular carcinoma (HCC) reported by OriGene~\cite{zhang2025origene}. We prompted {\ProjectName} to “identify understudied yet mechanistically promising druggable targets in HCC using multi-modal evidence.”

Using the \textit{Knowledge Flow Planner}, the system autonomously decomposed the task into differential expression analysis, mutation and copy-number evaluation, survival association testing, pathway enrichment, and tractability assessment. It queried canonical resources including GEPIA, TCGA, GEO, and OpenTargets to generate an initial pool of 125 candidate genes, which was progressively narrowed to \textit{GPR160} through multi-round evidence compression and reflection. The system further produced expression profiles, Kaplan--Meier survival curves, and KEGG pathway maps, revealing tumor-specific overexpression of \textit{GPR160}, its association with poor recurrence-free survival, and its potential involvement in immune-related signaling. This case demonstrates {\ProjectName}'s ability to translate open-ended biomedical questions into structured and mechanistically grounded evidence chains.

\begin{figure}[H]
    \centering
    \includegraphics[width=0.8\textwidth]{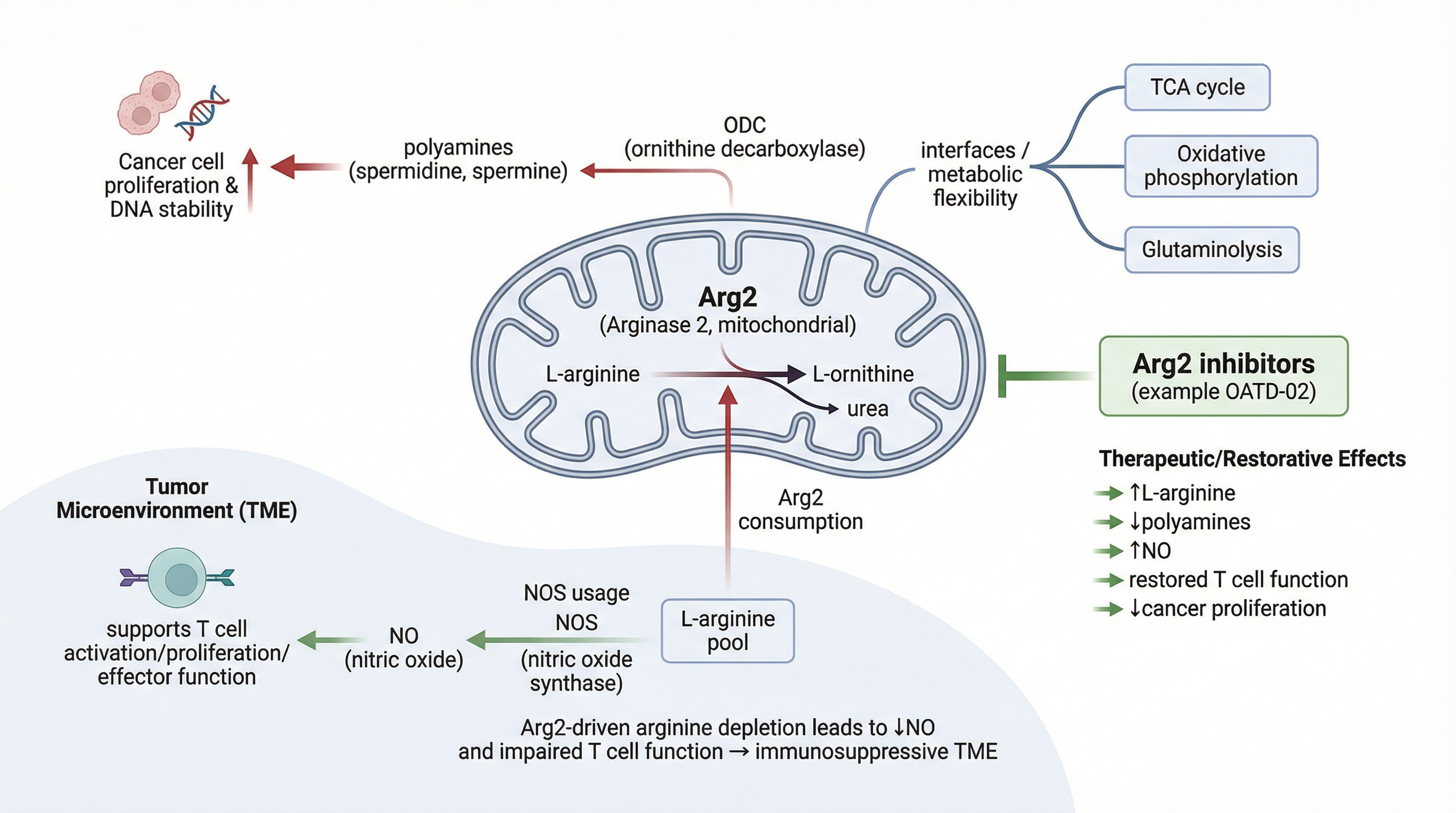}
    \caption{Mitochondrial Arg2 immunometabolic pathway and therapeutic intervention points}
    \label{fig:origene_image}
\end{figure}

\noindent\textbf{Hypothesis Generation and Target Prioritization.}
We further reproduced the identification of \textit{ARG2} as an overlooked yet mechanistically grounded target in colorectal cancer (CRC). The system constructed a multi-modal evidence graph integrating TCGA cohorts, Human Protein Atlas proteomics, UCSC genome annotations, pathway knowledge, and literature-derived molecular mechanisms. Built upon domain-specific reasoning templates, {\ProjectName} executed structured reasoning steps including disease gene consistency checks, pathway--phenotype alignment, pharmacological tractability analysis, and clinical differential expression testing.

Through iterative reflection cycles, \textit{ARG2} emerged as the top-ranked candidate, accompanied by mechanistic explanations involving metabolic reprogramming and immunosuppressive microenvironment remodeling. As illustrated in Fig.~\ref{fig:origene_image}, which is automatically generated by {\ProjectName}, mitochondrial ARG2-driven arginine depletion reduces nitric oxide (NO) availability, impairs T-cell effector function, and promotes tumor proliferation via enhanced polyamine biosynthesis, providing a unified metabolic--immunological rationale for therapeutic intervention. The complete report is released in our open-source repository.

{\ProjectName} further generated experiment-ready recommendations, including dose--response assays, patient-derived organoid (PDO) validation, and immune profiling protocols, consistent with those used in the original study. Notably, ARG2 inhibition exhibited dose-dependent anti-tumor effects in HCT116 cells and multiple CRC PDO models, supporting the validity of the generated hypotheses.

Together, these case studies show that {\ProjectName} can support end-to-end target discovery, bridging multi-omics evidence integration, mechanistic hypothesis generation, and experimental guidance.

\subsubsection{Biological Science}
\label{sec:bio_sci}
\begin{figure}[H]
    \centering
    \includegraphics[width=0.8\textwidth]{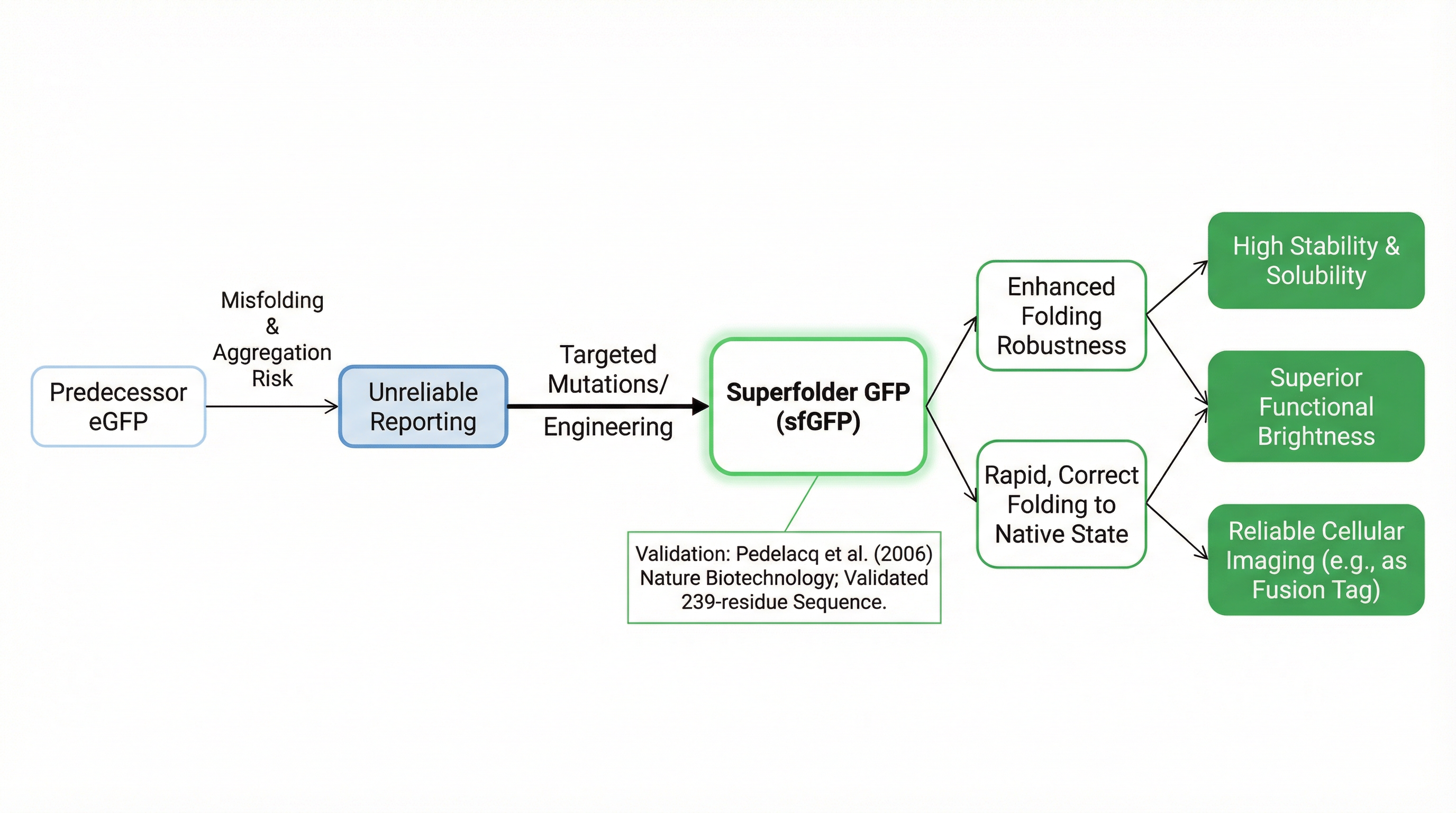}
    \caption{The engineering evolution from eGFP to sfGEP}
    \label{fig:scp_image}
\end{figure}

The experimental began with a targeted literature search by {\ProjectName} to identify fluorescent protein variants with strong brightness and folding stability. Evidence from peer reviewed studies pointed to sfGFP as a suitable candidate. This information, combined with predefined performance objectives, guided the design of the computational analyses and the experimental validation plan.

To evaluate these candidates, a series of dry lab and wet lab procedures were carried out using tools and devices coordinated through SCP~\citep{jiang2025scp}. The workflow included fluorescence assays, stability measurements, and quality control checks, paired with dry lab predictions of structural stability and sequence function relationships. The results show that sfGFP achieves high functional brightness and reliable folding efficiency, consistent with findings reported in the literature. Based on all data returned by SCP coordinated tools and instruments, {\ProjectName} generated a final experimental report that summarizes the empirical outcomes and identifies variants appropriate for downstream use. Figure~\ref{fig:scp_image}, which is automatically generated by InternAgent-1.5, presents an intermediate reasoning artifact automatically produced by {\ProjectName}, which outlines how evidence from the literature is transformed into an engineering rationale and target selection for sfGFP, and the complete report is available in our open source repository.

\subsubsection{Physical Science} 
\label{sec:exp_chem}

Building on the setup described in Sec.~\ref{sec:setup}, we demonstrate how {\ProjectName} addresses the dual challenges of strict atomic conservation and vast chemical search spaces in Physical Science. We report quantitative metrics on reaction prediction benchmarks and qualitative case studies in generative drug design.

\begin{table}[H]
    \centering
    \caption{Performance on Forward Major Product ($\text{Fwd}_{\text{major}}$) and By-product Prediction ($\text{Fwd}_{\text{by}}$). Top-1 accuracy and Fingerprint Tanimoto Similarity (FTS) are reported.}
    \label{tab:fwd_major_by}
    \begin{tabular}{lcccc}
        \toprule
        \multirow{2}{*}{Models} & \multicolumn{2}{c}{$\text{Fwd}_{\text{major}}$} & \multicolumn{2}{c}{$\text{Fwd}_{\text{by}}$} \\
        \cmidrule(lr){2-3} \cmidrule(lr){4-5}
         & Top-1 & FTS $\uparrow$ & Top-1 & FTS $\uparrow$ \\
        \midrule
        GPT-5.2 & 59 & 0.79 & 45 & 0.40\\   
        Claude4.5-sonnet-think & 0.74 & 0.90 & 0.49 & 0.43 \\
        o3-mini & 0.55 & 0.74 & 0.47 & \textbf{0.47}\\
        Gemini-3-Pro-Thinking & 0.81 & 0.91 & 0.45 & 0.36 \\
        {\ProjectName} &\textbf{ 0.86} & \textbf{0.94} & \textbf{0.62} & 0.42\\
        \bottomrule
    \end{tabular}
\end{table}

\paragraph{Automated Reaction Outcome Prediction.}
We evaluated the system on the ChemCoTBench~\citep{li2025beyond} forward prediction task. Unlike standard language models that treat molecular strings as mere text~\citep{weininger1988smiles}, {\ProjectName} adopts a physicochemical-aware approach by proactively invoking RDKit~\citep{landrum2013rdkit} to compute critical molecular descriptors (e.g., LogP, TPSA) and standardize SMILES representations. This descriptor-guided reasoning allows the system to accurately deduce the main product while simultaneously employing atomic conservation logic to infer by-products such as water or halides. As detailed in Table~\ref{tab:fwd_major_by}, {\ProjectName} achieves a Top-1 accuracy of 0.86 and a Fingerprint Tanimoto Similarity (FTS) of 0.94 for major product prediction. These results significantly outperform recent reasoning-enhanced models such as o3-mini (Top-1 0.55) and Gemini-3-Pro-Thinking (Top-1 0.81). Furthermore, in the challenging by-product prediction task ($\text{Fwd}_{\text{by}}$), our system achieves the highest Top-1 accuracy of 0.62, confirming its robustness in capturing complete reaction stoichiometry.

\begin{figure}[H]
    \centering
    \includegraphics[width=0.98\textwidth]{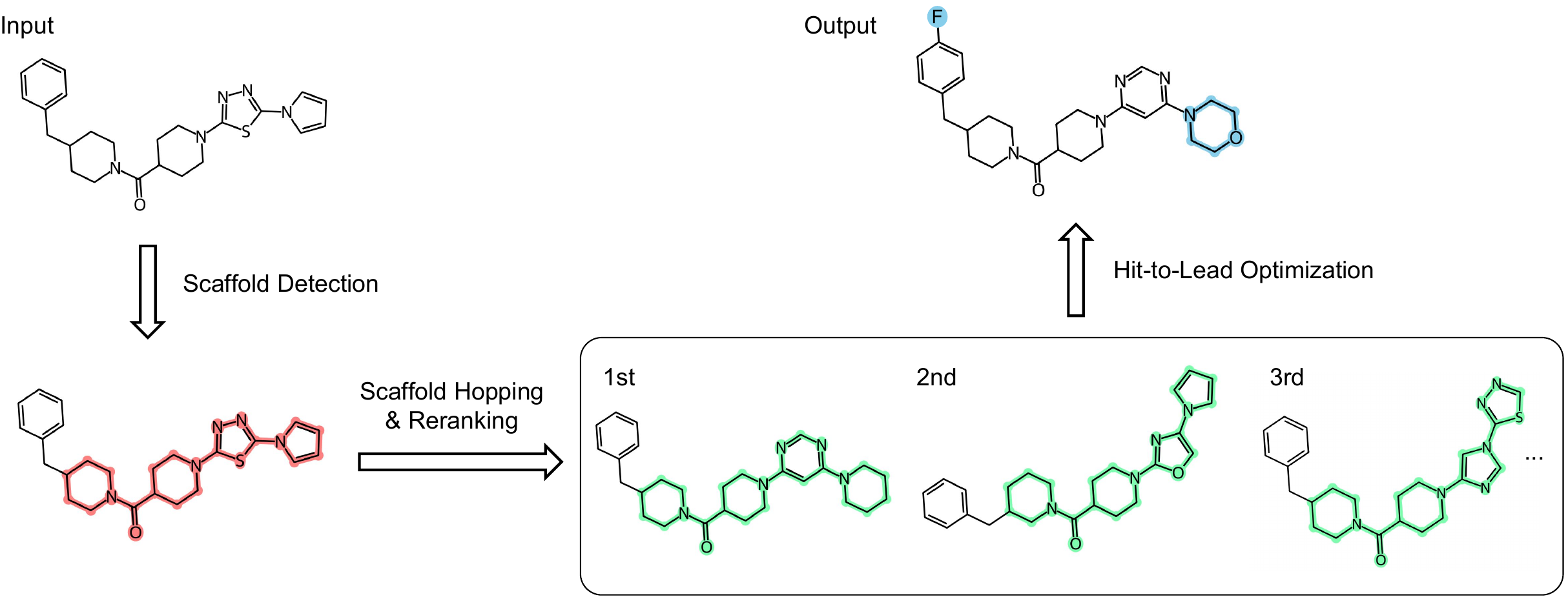}
    \caption{\textbf{Automated scaffold hopping and hit to lead optimization using {\ProjectName}.} The workflow begins by identifying the core scaffold in red from a DprE1 inhibitor hit. Through coordinated agent interaction, the system proposes structurally diverse bioisosteres and prioritizes piperidinopyrimidine variants shown in green. It then conducts a focused optimization step to address physicochemical limitations and generates a final candidate highlighted in blue, which features a modified heterocycle and fluorination. The resulting trajectory follows established medicinal chemistry practice and demonstrates the system’s ability to support rational drug design.}
    \label{fig:materail_scaffold}
\end{figure}

\paragraph{Generative Scaffold Hopping.} In the drug design domain, {\ProjectName} employs a generative multi-agent workflow that prioritizes 3D shape and electrostatic alignment over simple 2D topology. Crucially, the system integrates agent reasoning to refine candidates based on calculated metrics including Synthetic Accessibility (SA), Tanimoto Similarity, and LogP. When applied to a DprE1 inhibitor template known for solubility limitations~\citep{kovar2025scaffold}, the agent successfully navigated away from the original pyrrolothiadiazole core, proposing plausible bioisosteres based on piperidinopyrimidine scaffolds (Figure~\ref{fig:materail_scaffold}, Outputs 1st–3rd). Notably, the agent autonomously simulated a ``hit-to-lead'' optimization phase. It replicated expert-driven evolution by replacing the lipophilic piperidine side chain with a polar morpholine ring and introducing a fluorine atom at the para-position of the phenyl ring (Output), modifications critical for enhancing metabolic stability and solubility~\citep{kovar2025scaffold}.

\subsection{Effectiveness of Structured Cognitive Memory}
\begin{figure}[h]
\vspace{-2pt}
\begin{center}
\begin{subfigure}[b]{0.48\textwidth}
    \centering
    \includegraphics[width=\textwidth]{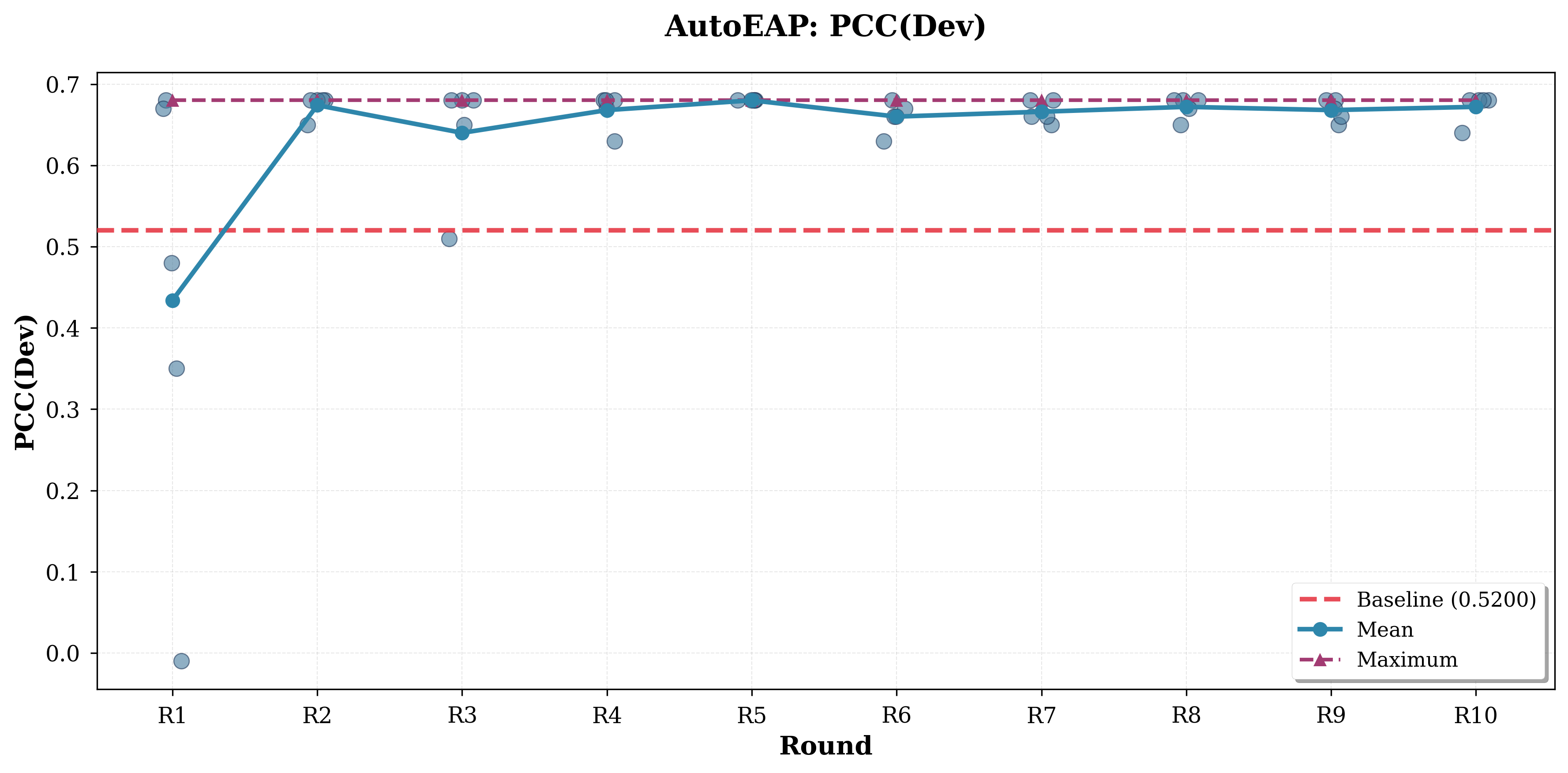}
    \label{fig:performance}
\end{subfigure}
\hfill
\begin{subfigure}[b]{0.48\textwidth}
    \centering
    \includegraphics[width=\textwidth]{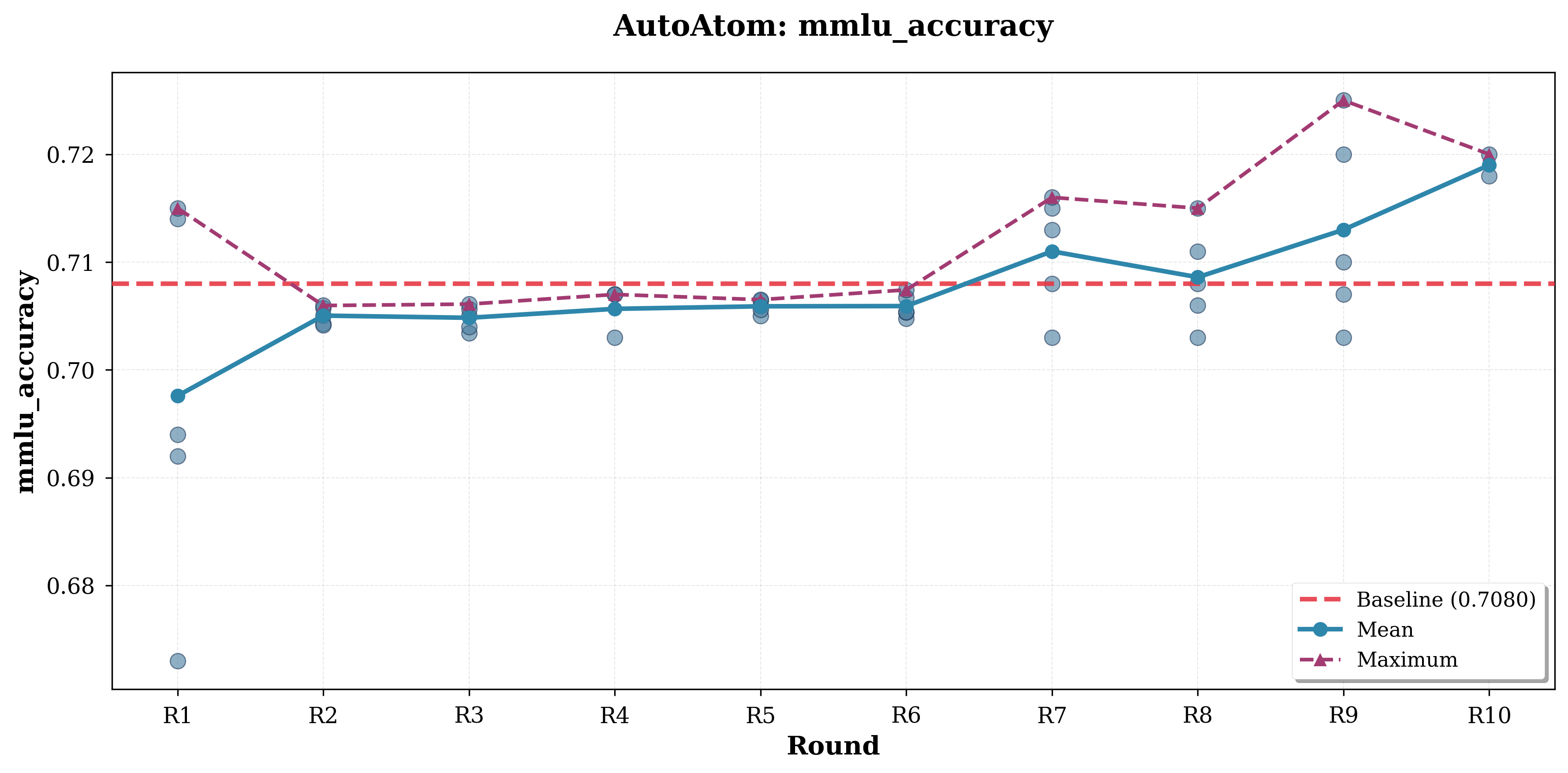}
    \label{fig:prompt_evolve}
\end{subfigure}
\end{center}
\vspace{-10pt}
\caption{Experimental validation of memory effectiveness on algorithm discovery tasks.}
\label{img:mem_line}
\vspace{-4pt}
\end{figure}

\begin{figure}[h]
\vspace{-2pt}
\begin{center}
\includegraphics[width=\textwidth]{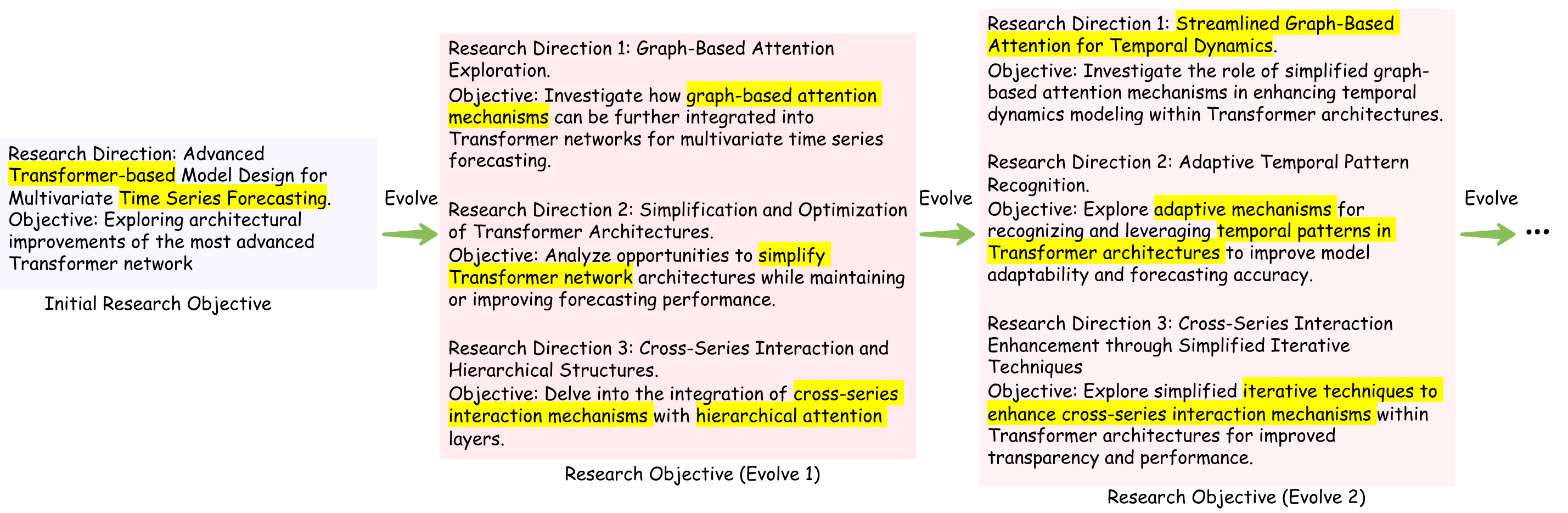} 
\end{center}
\vspace{-10pt}
\caption{The evolution process of research objectives on the AutoTSF task.}\label{img:prompt_evolve}
\vspace{-4pt}
\end{figure}
Our system is designed to operate continuously over extended periods, and the Structured Cognitive Memory subsystem is a core component that enables sustained improvement across diverse scientific discovery tasks. To isolate how each module contributes to long-horizon capability, we evaluate Task-Episodic Memory (TEM), Semantic-Knowledge Memory (SKM), and Strategy-Procedural Memory (SPM) using modalities aligned with their functional roles.

\noindent\textbf{Task-Episodic Memory.}
We analyze TEM through performance trajectories over iterative research steps. With TEM active, curves rise smoothly, indicating stable short-horizon adaptation. Retrieved episodes provide fine-grained evidence from earlier trials, helping the model avoid ineffective methodological choices and refine hypotheses efficiently. Removing TEM yields irregular or stagnant progression, with frequent revisiting of unproductive strategies due to missing within-session outcomes. This contrast shows episodic grounding is critical for robust, sample-efficient adaptation during sustained operation. Fig.\ref{img:mem_line} further presents long-horizon optimization experiments on scientific-discovery tasks, illustrating how TEM supports stable and persistent iterative improvement.

\noindent\textbf{Semantic-Knowledge Memory.}
To assess SKM, we use prompt-based case studies where the system proposes new research directions after multiple exploration batches. With SKM enabled, objectives reflect accumulated understanding of successful and unsuccessful methodological patterns. Retrieved long-term knowledge helps avoid saturated conceptual regions while preserving semantic continuity and measurable novelty. Fig.~\ref{img:prompt_evolve} illustrates iterative evolution of research objectives from an initial seed, showing that SKM maintains cross-batch coherence while allowing strategic redundancy to deepen promising sub-domains, balancing exploration breadth with exploitation depth.

\noindent\textbf{Strategy-Procedural Memory.}
As shown in Table~\ref{tab:memory_ablation}, we evaluate SPM on benchmark tasks requiring multi-step reasoning and coordinated planning. The full system achieves higher success rates and more coherent plans than the SPM-ablated baseline. SPM provides procedural priors that improve planning structure and execution-level tool selection, reducing unnecessary branching and redundant calls. Without SPM, plans become longer and fragmented, with error propagation and imprecise tool-call parameters. Overall, SPM supports transferable procedural structure and improves planning efficiency and execution rigor.

Overall, TEM, SKM, and SPM provide complementary support across short-term adaptation, long-term knowledge accumulation, and efficient reasoning execution for sustained improvement.

\begin{table}[t]
\vspace{-6pt}
\centering
\caption{Ablation study on the strategy-procedural memory on the GAIA benchmark. We report accuracy and the average of tool calls to evaluate both performance and efficiency.}
\label{tab:memory_ablation}
\vspace{-6pt}

\resizebox{\linewidth}{!}{
\begin{tabular}{lcccccccc}
\toprule
& \multicolumn{4}{c}{GAIA Accuracy (\%) $\uparrow$} 
& \multicolumn{4}{c}{Avg. \# Tool Calls $\downarrow$} \\
\cmidrule(lr){2-5} \cmidrule(lr){6-9}
Agent 
& Level 1 & Level 2 & Level 3 & Avg. 
& Level 1 & Level 2 & Level 3 & Avg. \\
\midrule
InternAgent 1.5 w/o SPM
& 92.45 & 84.88 & 53.85 & 82.42 
& 12.06 & 23.51 & 55.65 & 22.69 \\

InternAgent 1.5   
& \textbf{92.45} & \textbf{89.53} & \textbf{61.54} & \textbf{86.06} 
& \textbf{9.13} & \textbf{21.22} & \textbf{37.33} & \textbf{18.52} \\
\bottomrule
\end{tabular}
}
\end{table}

%% file: sections/related.tex
\section{Related Work}
\subsection{Agentic AI for Scientific Discovery}
Recent progress in agentic AI has produced systems capable of carrying out increasingly autonomous forms of scientific reasoning. The AI Scientist~\cite{lu2024ai} line of work demonstrates early examples of end to end research automation, with the initial system coordinating hypothesis generation and experiment design, and the later version~\cite{yamada2025ai} replacing fixed templates with a search based procedure that allows broader exploration of methodological space. AlphaEvolve~\cite{novikov2025alphaevolve}approaches scientific discovery from an evolutionary perspective by using language models to generate candidate algorithms and iteratively refine them through performance guided optimization. Other recent systems emphasize multi agent coordination within real scientific workflows. AI Co-Scientist~\cite{gottweis2025towards} distributes literature analysis, hypothesis refinement, and methodological planning across specialized agents directed by a central model, while Robin integrates planning, data analysis, and validation into a closed loop system capable of discovering new compound candidates without manual intervention. Kosmos~\cite{mitchener2025kosmos} further advances this direction by unifying literature retrieval, experiment design, and theory development into a continuously running discovery engine. Overall, these efforts illustrate the rapid emergence of autonomous scientific discovery systems and highlight the importance of long horizon reasoning, iterative experimentation, and persistent state management. These themes directly motivate the structured memory mechanisms developed in our work.

\subsection{Deep Research Agents}

Recent advances in Deep Research (DR) agents extend LLMs from retrieval-augmented generation to dynamic, tool-driven research workflows. Early systems such as WebGPT~\citep{nakano2021webgpt} and Toolformer~\citep{schick2023toolformer} explored web and API integration, demonstrating how models can reason over retrieved information while selectively invoking external tools. Building on these ideas, industrial solutions \textit{e.g.}, OpenAI DR~\citep{openai2025deepresearch}, Gemini DR~\citep{google2024geminidr}, Grok DR~\citep{xai2025grokdeepsearch}, and Perplexity DR~\citep{perplexity2025deepresearch}, incorporate adaptive planning, iterative retrieval, and multimodal reasoning to support long-horizon research tasks. Recently, single-agent designs (\textit{e.g.}, Search-o1~\citep{li2025search}, WebDancer~\citep{wu2025webdancer}, Tongyi DeepResearcher~\citep{qiao2025webresearcher}, MiroThinker~\citep{team2025mirothinker-v1-5}) enable end-to-end optimization within a unified reasoning loop, while multi-agent architectures (\textit{e.g.}, AI Scientist~\citep{lu2024ai}, Agent Laboratory~\citep{schmidgall2025agent}, and InternAgent~\citep{team2025novelseek}) offer greater modularity and scalability, which are particularly beneficial for complex research settings. Recent studies, \textit{e.g.}, GeAR~\citep{shen2024gear} and PANGU DeepDiver~\citep{shi2025pangudeepdiver}, further demonstrate the value of explicit structures and self-evolving mechanisms for multi-hop reasoning.

\subsection{Memory Mechanism}
Agent memory has become a central component of modern agent systems, enabling long‑horizon reasoning, continual adaptation, and interaction with complex environments~\citep{hu2025memory}. Recent advances cover token‑level mechanisms~\citep{wumemorizing} that extend contextual retention, parametric approaches that internalize accumulated experience into model parameters, and latent‑memory systems~\citep{wang2023augmenting} that store structured trajectories to guide future decisions. In parallel, short‑term interaction memory has been explored in conversational and agent‑simulation settings, where systems maintain ephemeral contextual traces to support local reasoning over brief episodes~\citep{park2023generative}. Long‑term episodic memory has also been investigated through architectures that accumulate environment interactions across extended horizons and retrieve them for subsequent decisions~\citep{xuAMemAgenticMemory2025a}, providing persistent records of agent experience. These techniques enhance an agent’s mechanisms for incorporating prior information, although they are typically designed for interaction settings with limited temporal scope and therefore remain orthogonal to the multi‑stage workflows considered in scientific discovery.

%% file: sections/conclusion.tex
\section{Conclusion}
In this work, we presented {\ProjectName}, a unified system for end‑to‑end scientific discovery. The framework integrates generation, verification, and evolution into a coherent architecture supported by foundational capabilities for deep research, solution refinement, and long horizon memory. This design enables consistent information flow across stages and provides a general substrate for cross‑disciplinary scientific workflows.

Comprehensive evaluations demonstrate that {\ProjectName} achieves exhibits strong performance in structured scientific reasoning. The system autonomously produces competitive algorithmic solutions, optimizes experimental proposals over extended trajectories, and executes multi‑step computational and empirical workflows. Across algorithmic and empirical domains, {\ProjectName} consistently generates outputs that align with established scientific principles and reproduce findings observed in real scientific studies.

Future work includes strengthening the coupling between computational reasoning and experimental validation, and accelerating the transition from generated hypotheses to verifiable results. Advancing these directions will further improve the efficiency and reliability of cross-disciplinary scientific discovery.

%\clearpage

%% file: sections/appendix.tex
\appendix
\section{Appendix}

\subsection{Contributions and Acknowledgments}

\definecolor{damaiblue}{RGB}{10, 102, 155}
\definecolor{damaiorange}{RGB}{180,50,50}
\definecolor{damaired}{RGB}{10, 50, 50}

\noindent
\textbf{\color{damaired} Lead Authors} \\ [1.5mm]
Shiyang Feng$^{1}$, Runmin Ma$^{1}$, Xiangchao Yan$^{1}$

\noindent
\textbf{\color{damaired} Core Authors} \\ [1.5mm]
Yue Fan$^{1}$, Yusong Hu$^{1,6}$, Songtao Huang$^{1,2}$, Shuaiyu Zhang$^{1,2}$, Zongsheng Cao$^{1}$, Tianshuo Peng$^{1,4}$, Jiakang Yuan$^{1,2}$, Zijie Guo$^{1,2}$, Zhijie Zhong$^{1}$, Shangheng Du$^{1,5}$, Weida Wang$^{1,2}$, Jinxin Shi$^{1,5}$, Yuhao Zhou$^{1}$

\noindent
\textbf{\color{damaired} Contributors} \\ [1.5mm]
Xiaohan He, Zhiyin Yu, Fangchen Yu, Bihao Zhan, Qihao Zheng, Jiamin Wu, Mianxin Liu, Chi Zhang, Shaowei Hou, Shuya Li, Yankai Jiang, Wenjie Lou, Lilong Wang, Zifu Wang, Jiong Wang, Wanghan Xu, Yue Deng, Dongrui Liu, Yiheng Wang

\noindent
\textbf{\color{damaired} Scientific Directors} \\  [1.5mm]
Wenlong Zhang$^{1}$, Fenghua Ling$^{1}$, Shufei Zhang$^{1}$, Xiaosong Wang$^{1}$, Shuangjia Zheng$^{3}$, Xun Huang$^{3}$, Siqi Sun$^{1,2}$, Shuyue Hu$^{1}$, Peng Ye$^{1,4}$, Chunfeng Song$^{1}$, Bin Wang$^{1}$, Conghui He$^{1}$, Yihao Liu$^{1}$, Xin Li$^{1}$, Qibin Hou$^{6}$, Tao Chen$^{2}$, Xiangyu Yue$^{1,4}$, Bin Wang$^{2}$, Liang He$^{1,5}$, Dahua Lin$^{1}$, Bowen Zhou$^{1}$

\noindent
\textbf{\color{damaired} Corresponding Authors} \\  [1.5mm]
Bo Zhang$^{1}$, zhangbo@pjlab.org.cn \\
Lei Bai$^{1}$, bailei@pjlab.org.cn \\

\textbf{Main Affiliations}

$^1$ Shanghai Artificial Intelligence Laboratory \\
$^2$ Fudan University \\
$^3$ Lingang Laboratory \\
$^4$ The Chinese University of Hong Kong  \\
$^5$ East China Normal University \\
$^6$ Nankai University  \\

%%%%%%%%%%%%%%%%%%%%%%%%%%%%%%%%%%%%%%%%%%%%%%%%%%%%%%%%%%%%

\clearpage

\subsection{Earth Science example}
\label{earth_science:example}
\vspace{-0.5em}
{\raggedleft \normalsize \color{gray} \textbf{\textit{Report Generated All by \ProjectName}}\par}

\begin{figure*}[b]
\vspace{-100pt}
\begin{center}
\fbox{
\includegraphics[page=1,width=0.9\textwidth]{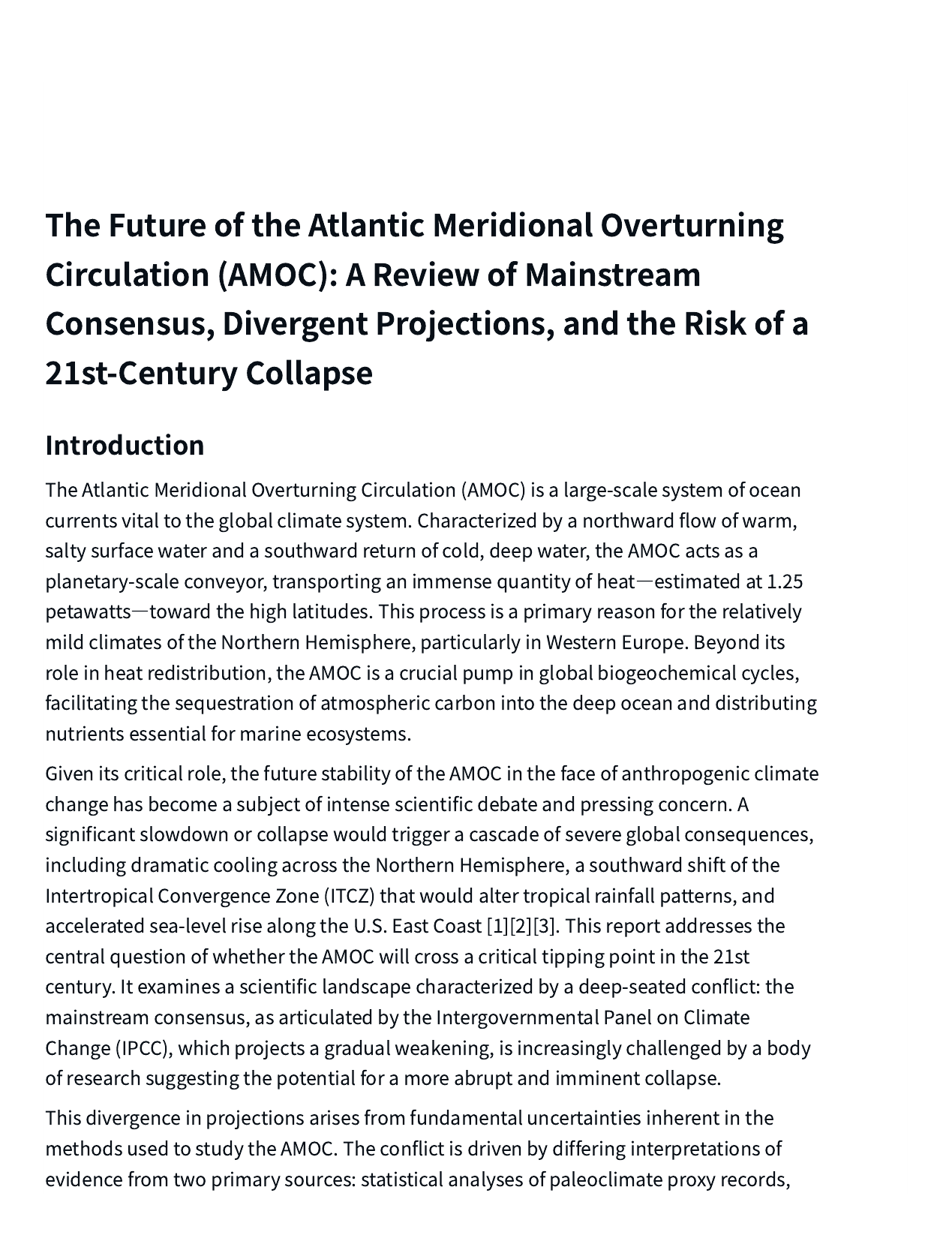} 
}
\end{center}
\vspace{-12pt}
\end{figure*}

\clearpage

\begin{figure*}[t]
\vspace{-6pt}
\begin{center}
\fbox{
\includegraphics[page=2,width=0.9\textwidth]{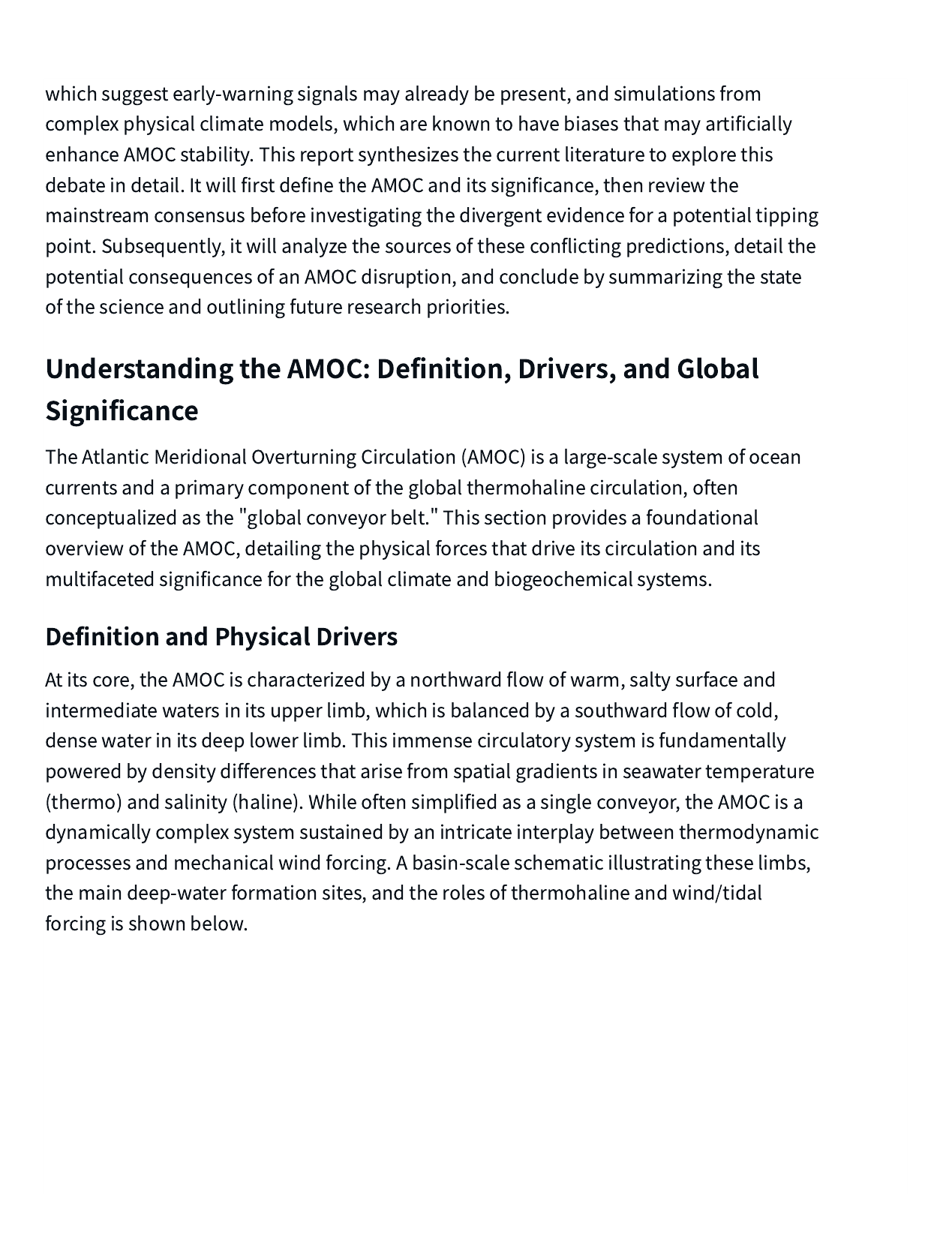} }
\end{center}
\vspace{-12pt}
\end{figure*}

\clearpage

\begin{figure*}[t]
\vspace{-6pt}
\begin{center}
\fbox{
\includegraphics[page=3,width=0.9\textwidth]{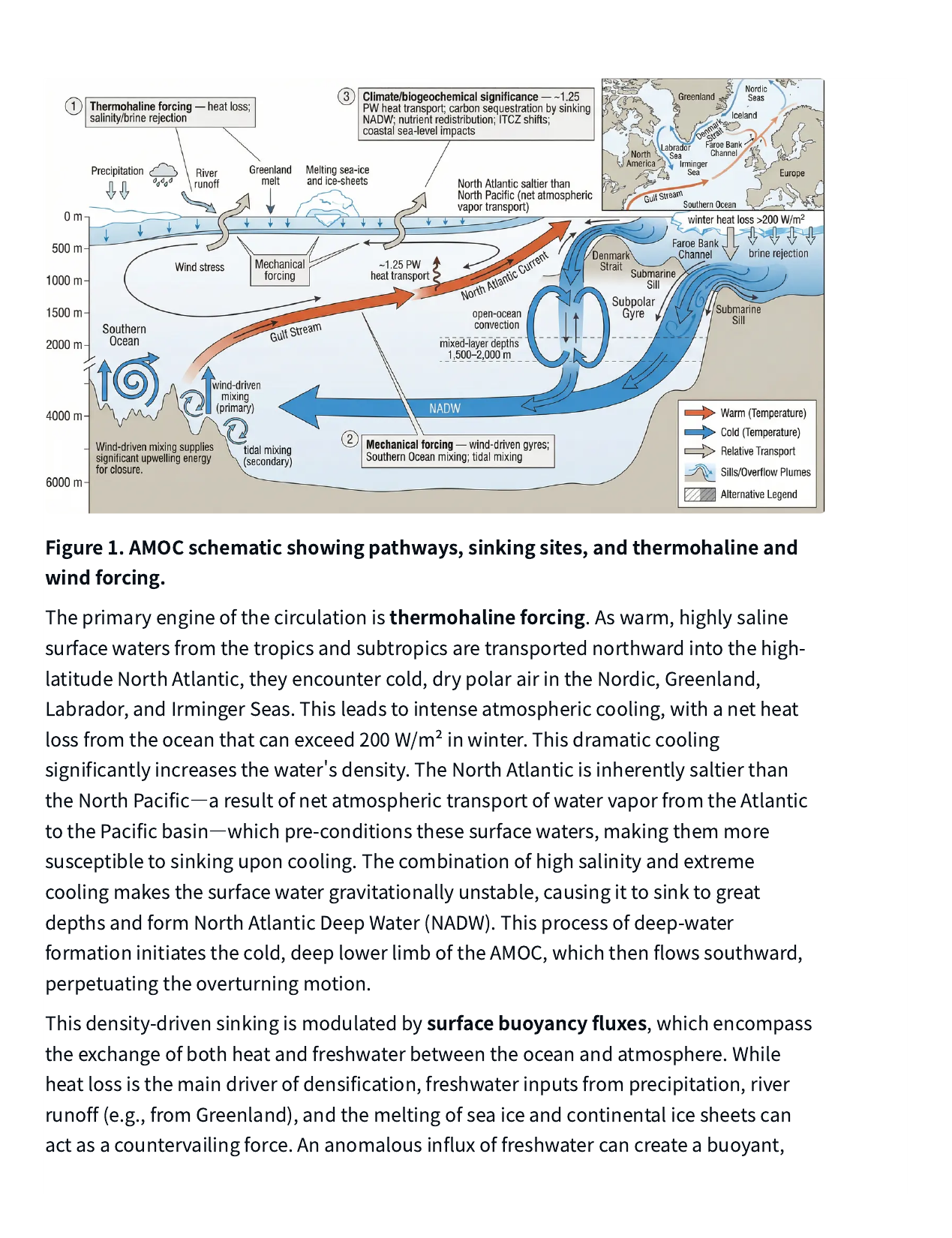} }
\end{center}
\vspace{-12pt}
\end{figure*}

\clearpage

\begin{figure*}[t]
\vspace{-6pt}
\begin{center}
\fbox{
\includegraphics[page=4,width=0.9\textwidth]{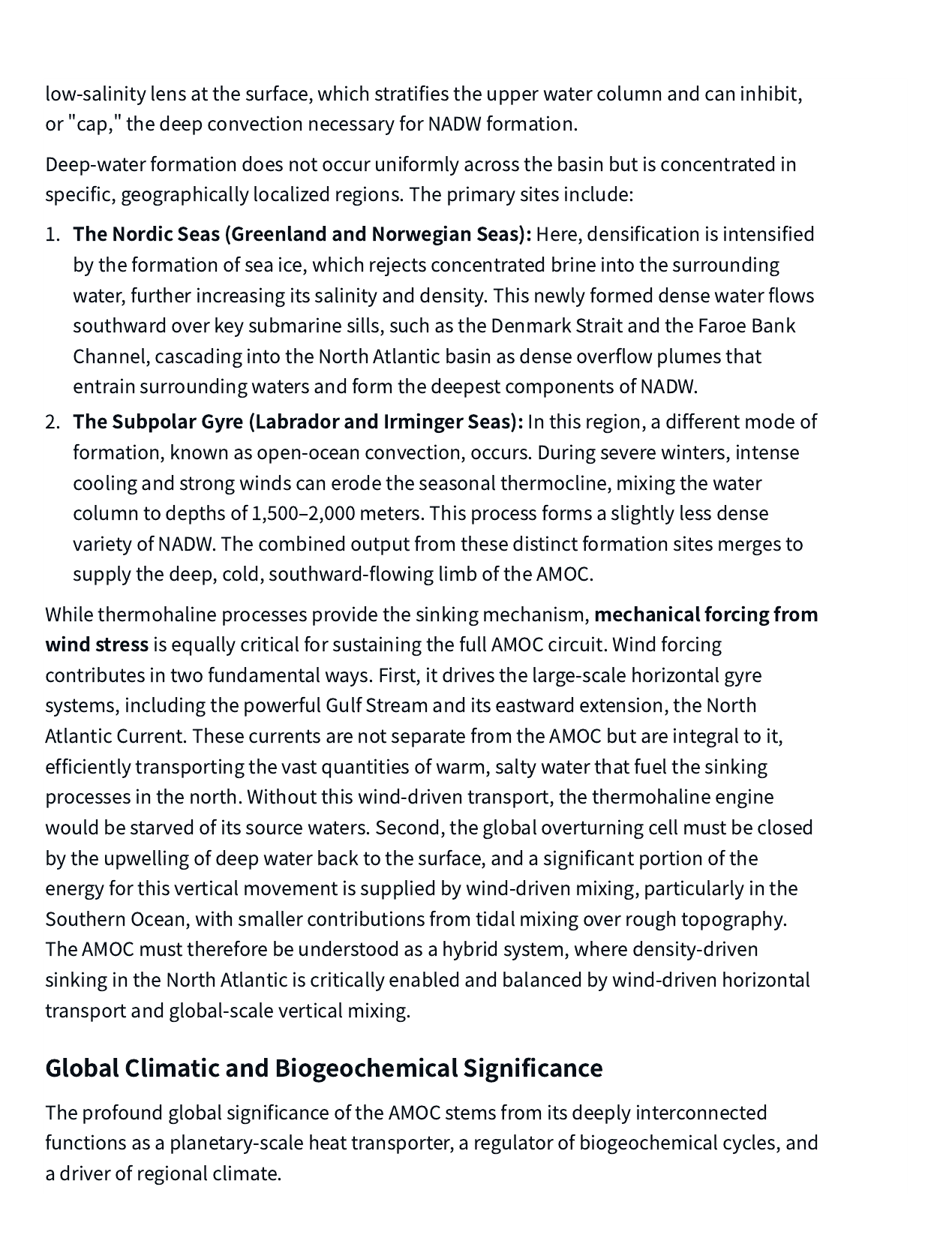} }
\end{center}
\vspace{-12pt}
\end{figure*}

\clearpage

\begin{figure*}[t]
\vspace{-6pt}
\begin{center}
\fbox{
\includegraphics[page=5,width=0.9\textwidth]{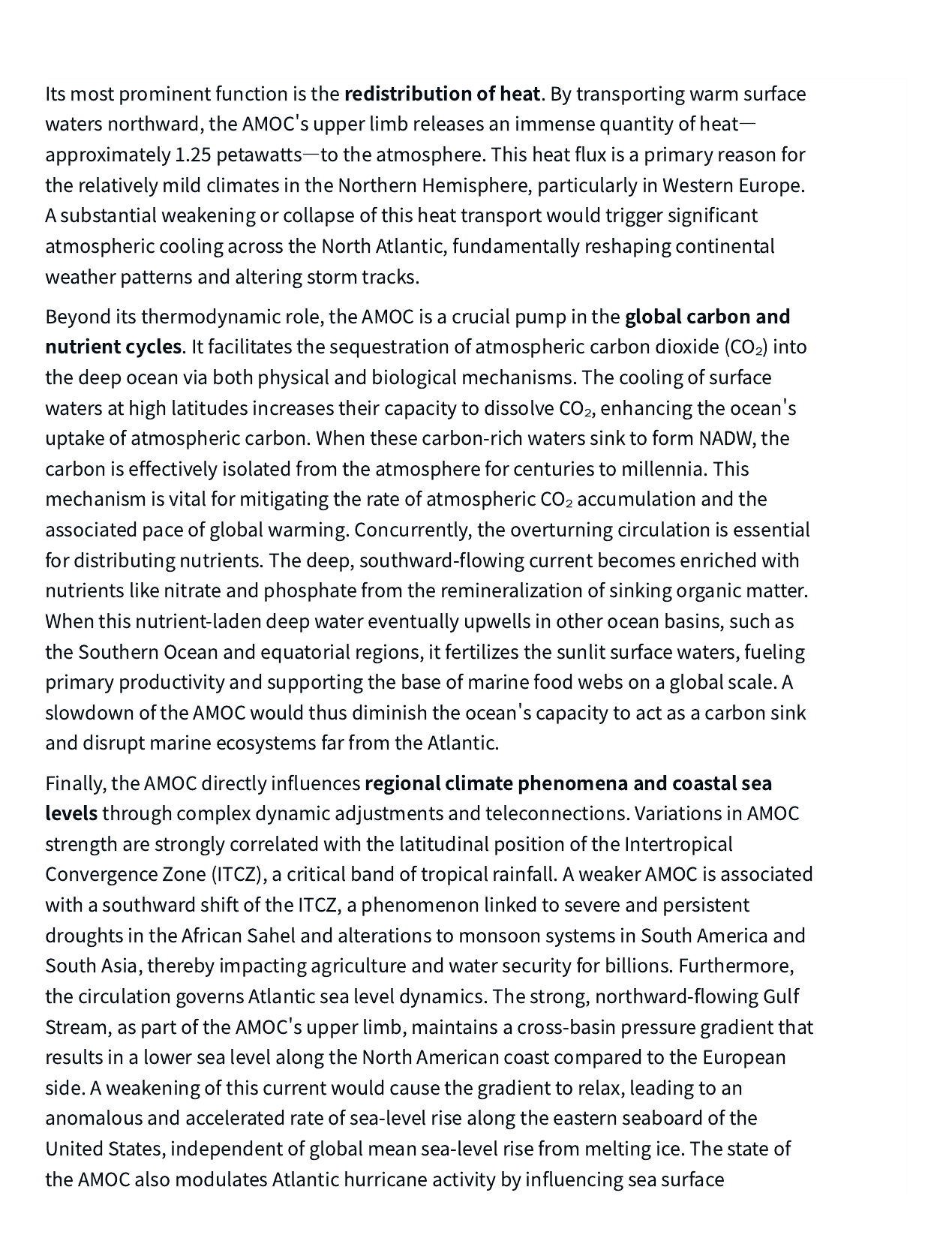} }
\end{center}
\vspace{-12pt}
\end{figure*}

\clearpage

\begin{figure*}[t]
\vspace{-6pt}
\begin{center}
\fbox{
\includegraphics[page=6,width=0.9\textwidth]{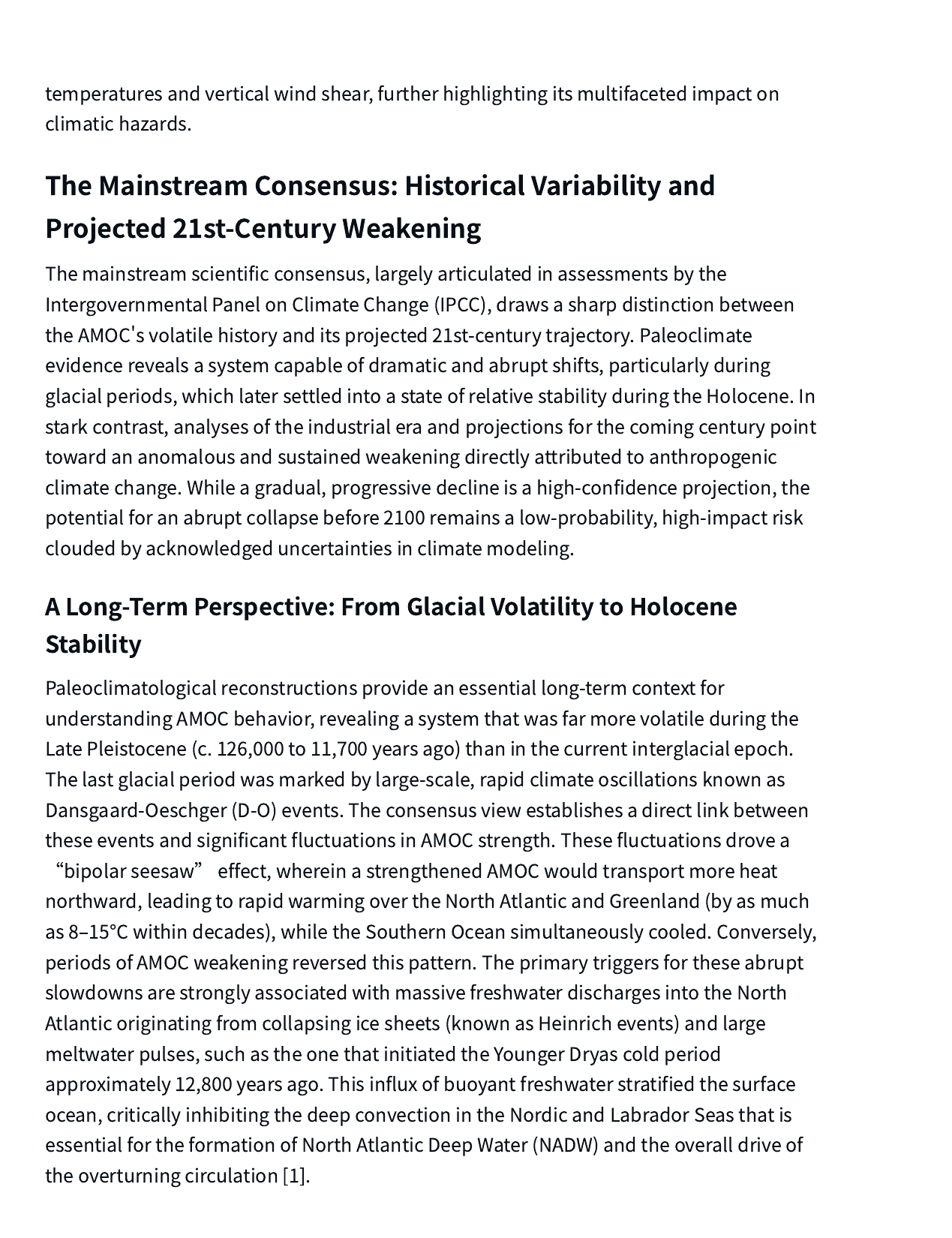} }
\end{center}
\vspace{-12pt}
\end{figure*}

\clearpage

\begin{figure*}[t]
\vspace{-6pt}
\begin{center}
\fbox{
\includegraphics[page=7,width=0.9\textwidth]{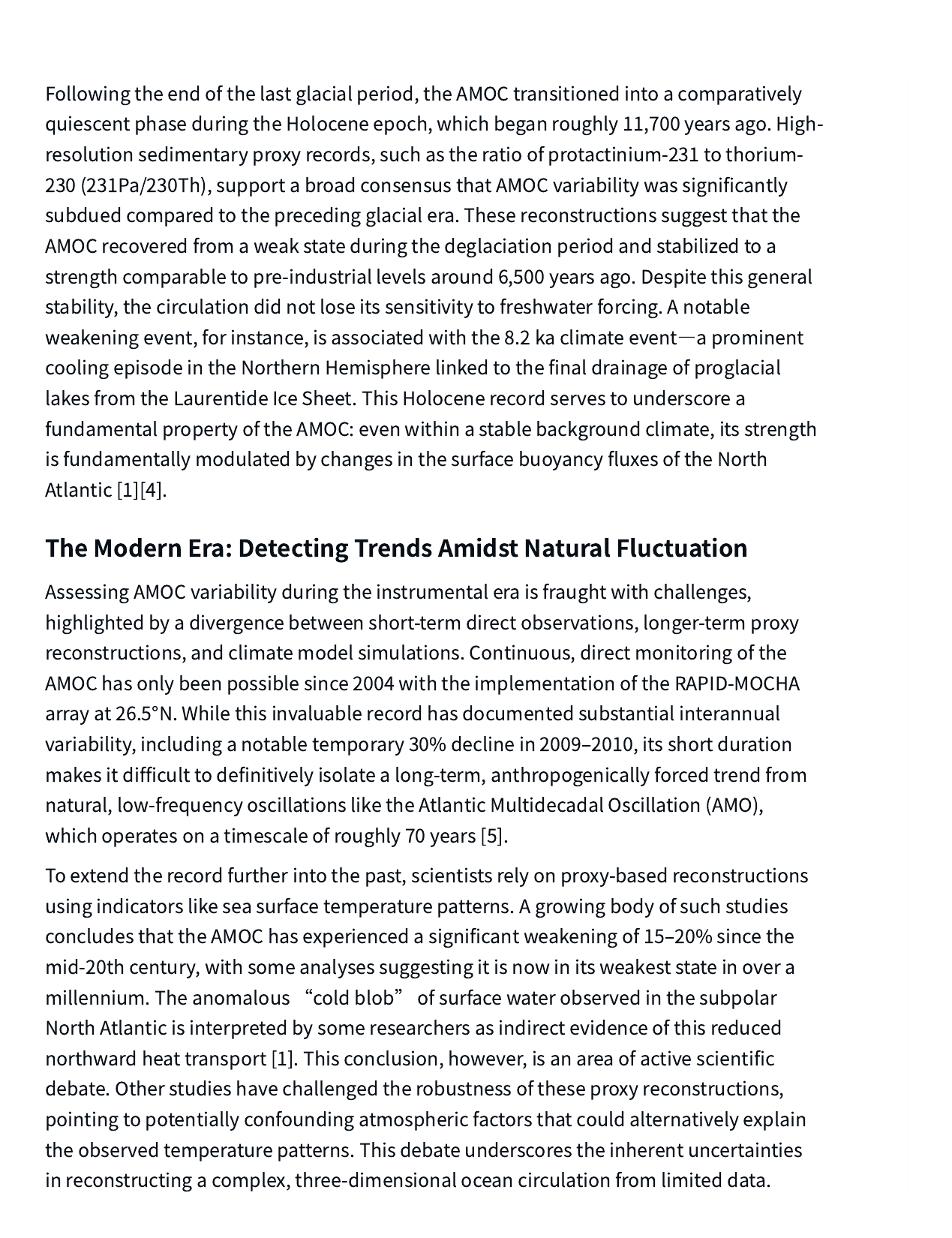} }
\end{center}
\vspace{-12pt}
\end{figure*}

\begin{figure*}[t]
\vspace{-6pt}
\begin{center}
\fbox{
\includegraphics[page=8,width=0.9\textwidth]{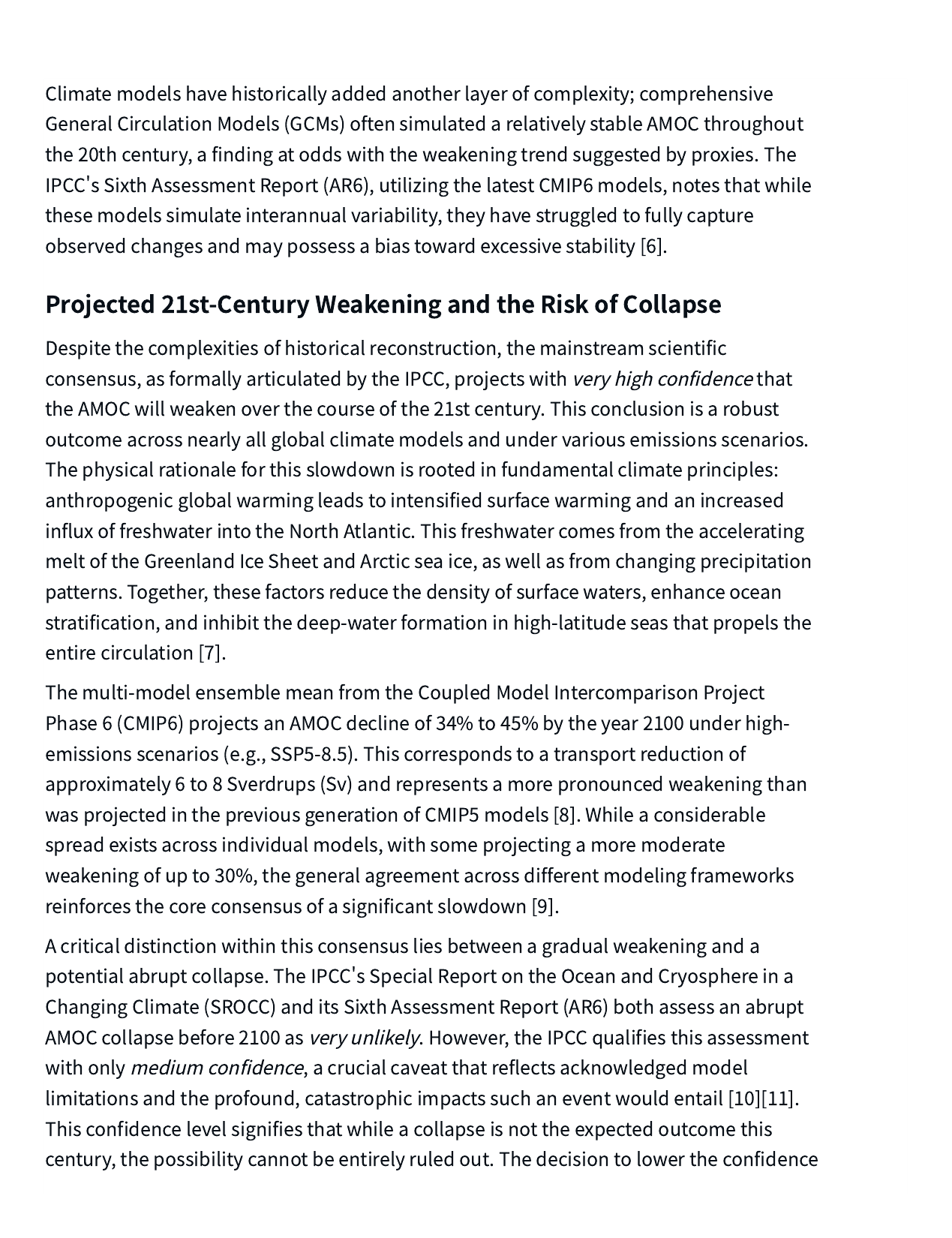} }
\end{center}
\vspace{-12pt}
\end{figure*}

\begin{figure*}[t]
\vspace{-6pt}
\begin{center}
\fbox{
\includegraphics[page=9,width=0.9\textwidth]{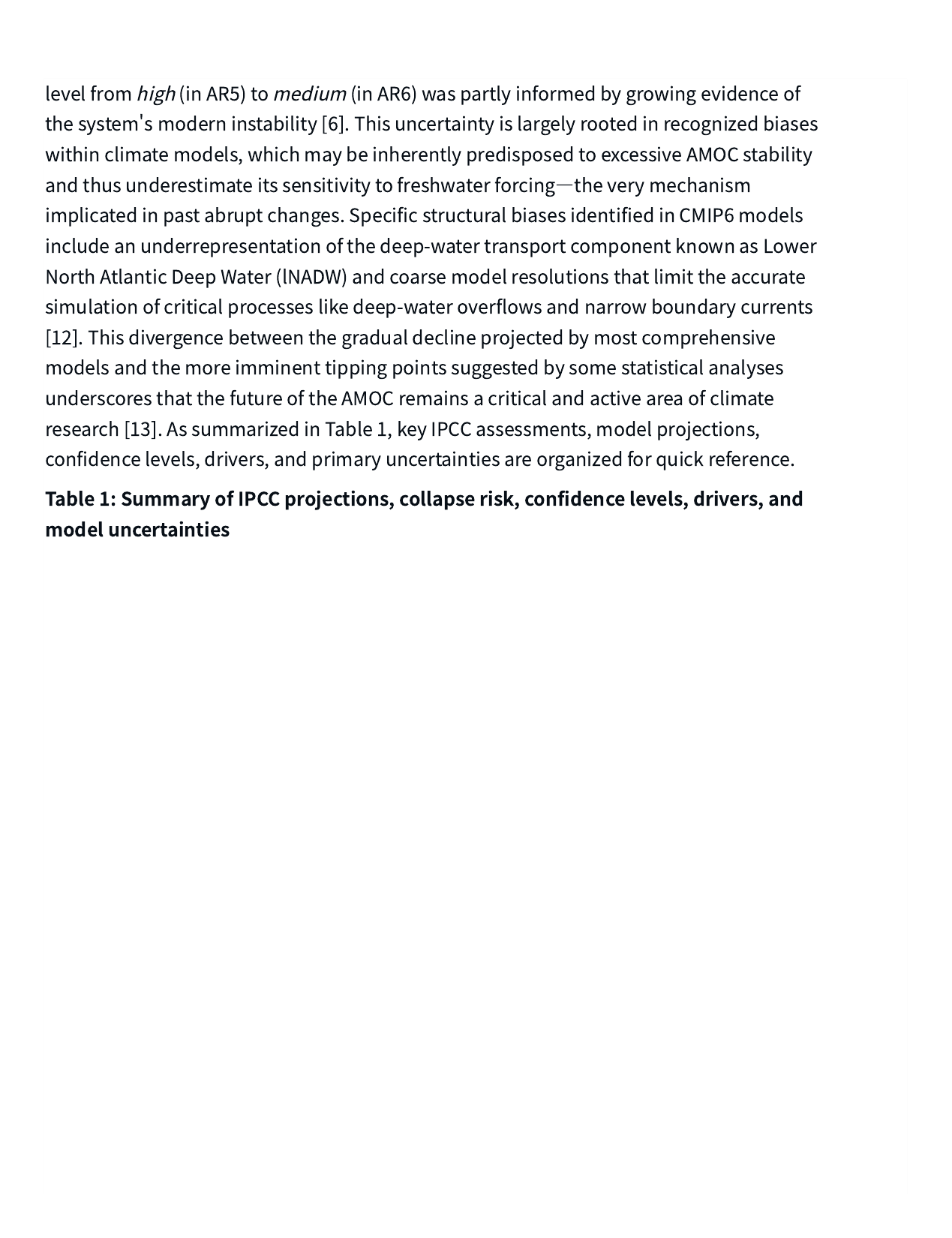} }
\end{center}
\vspace{-12pt}
\end{figure*}

\begin{figure*}[t]
\vspace{-6pt}
\begin{center}
\fbox{
\includegraphics[page=10,width=0.9\textwidth]{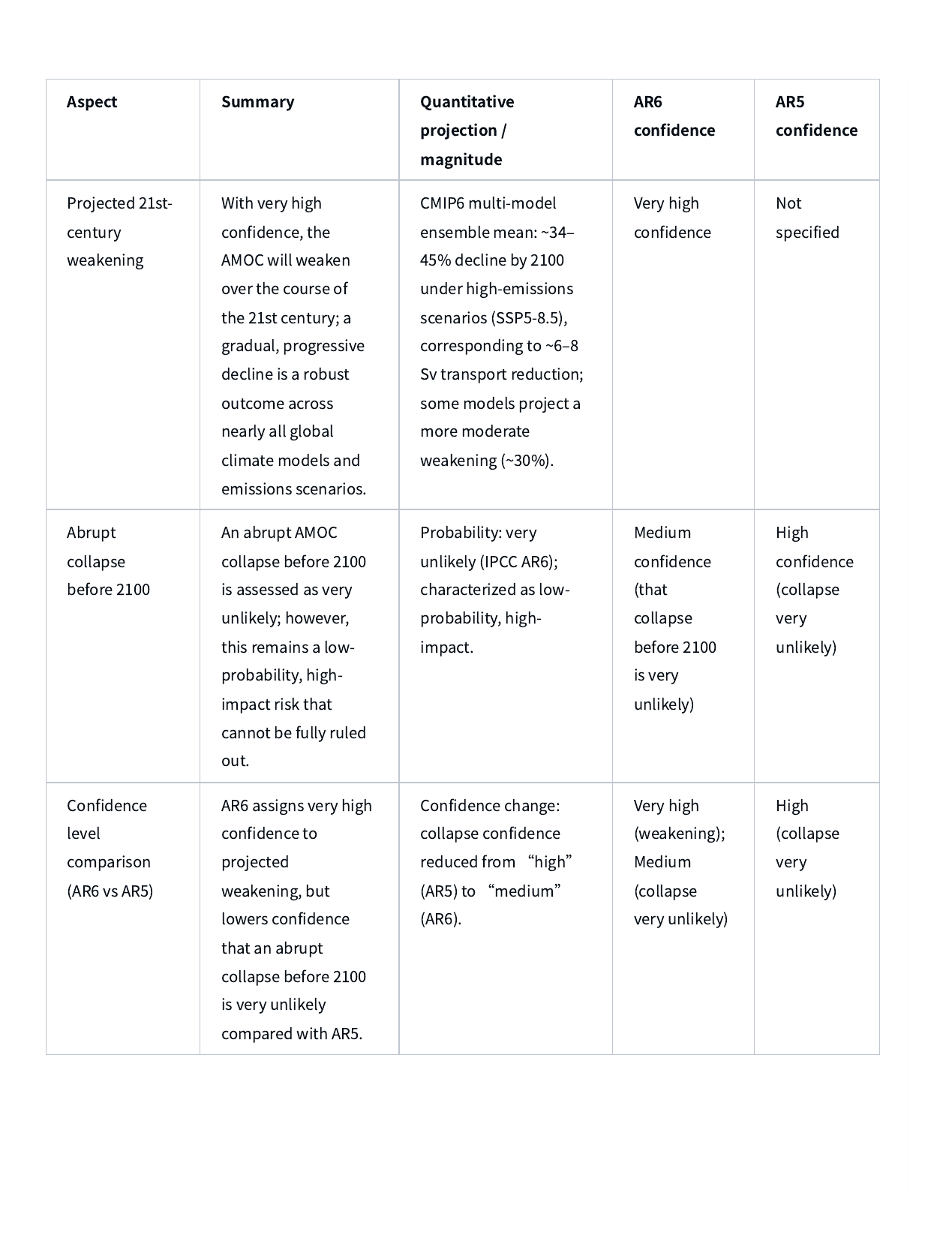} }
\end{center}
\vspace{-12pt}
\end{figure*}

\begin{figure*}[t]
\vspace{-6pt}
\begin{center}
\fbox{
\includegraphics[page=11,width=0.9\textwidth]{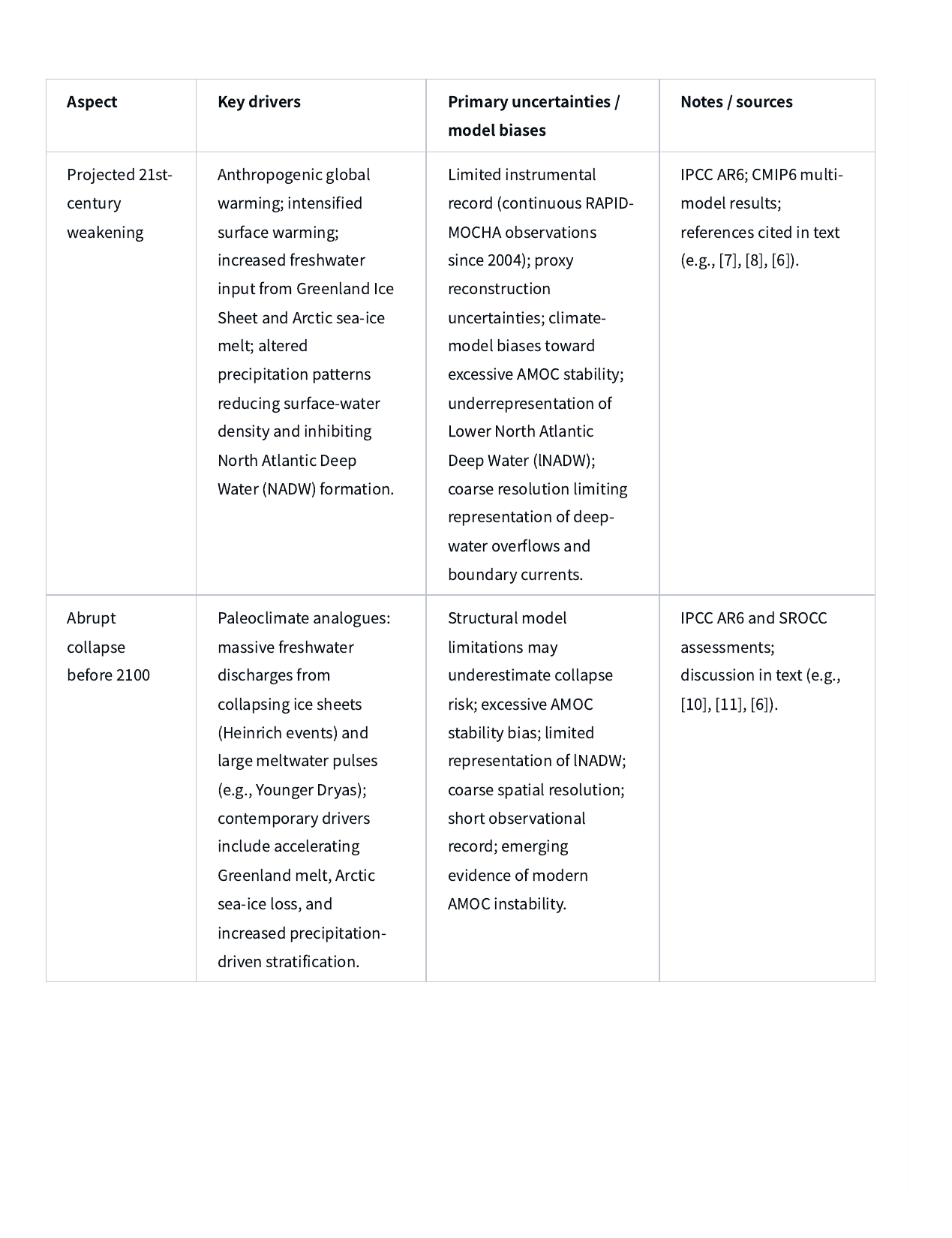} }
\end{center}
\vspace{-12pt}
\end{figure*}

\begin{figure*}[t]
\vspace{-6pt}
\begin{center}
\fbox{
\includegraphics[page=12,width=0.9\textwidth]{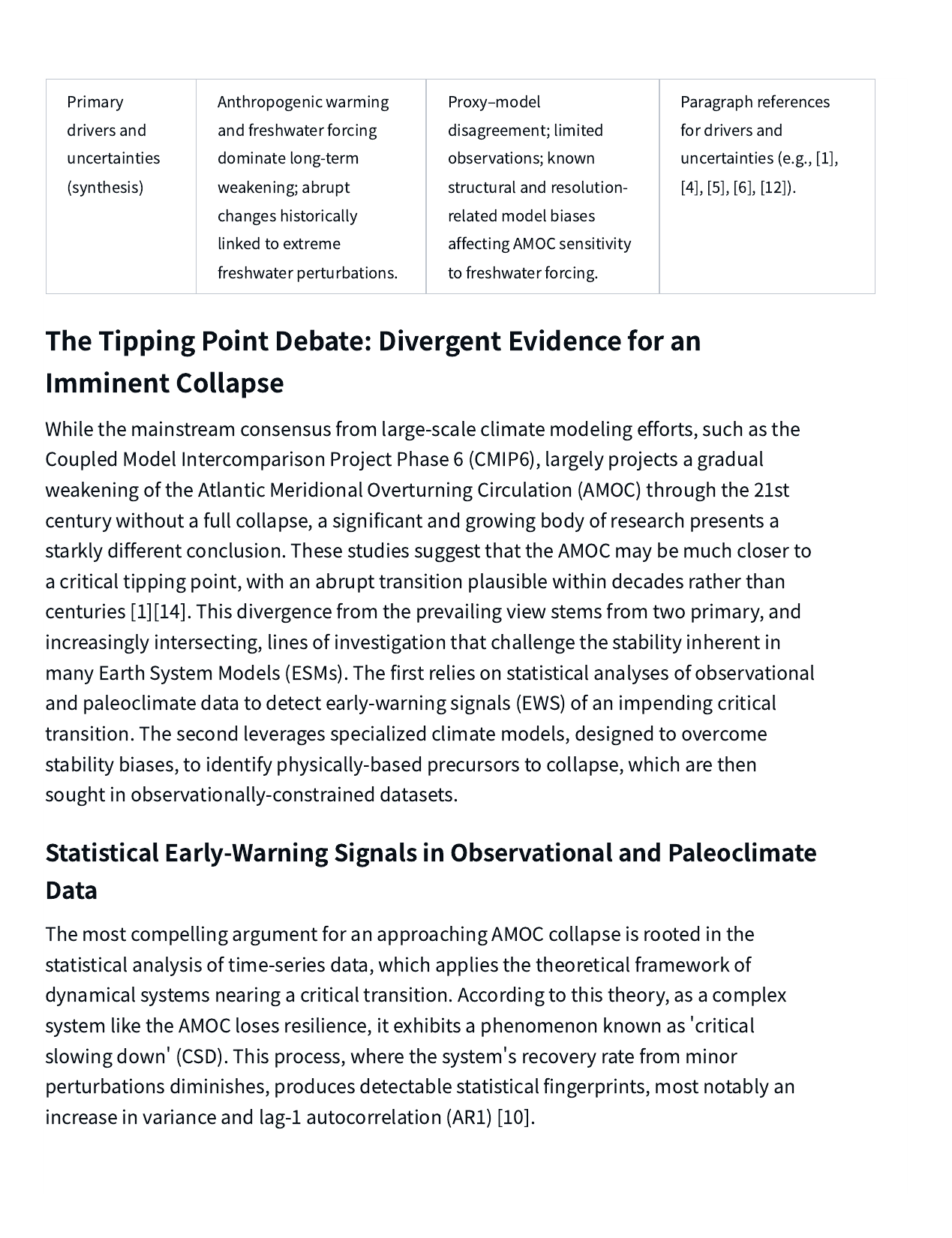} }
\end{center}
\vspace{-12pt}
\end{figure*}

\begin{figure*}[t]
\vspace{-6pt}
\begin{center}
\fbox{
\includegraphics[page=13,width=0.9\textwidth]{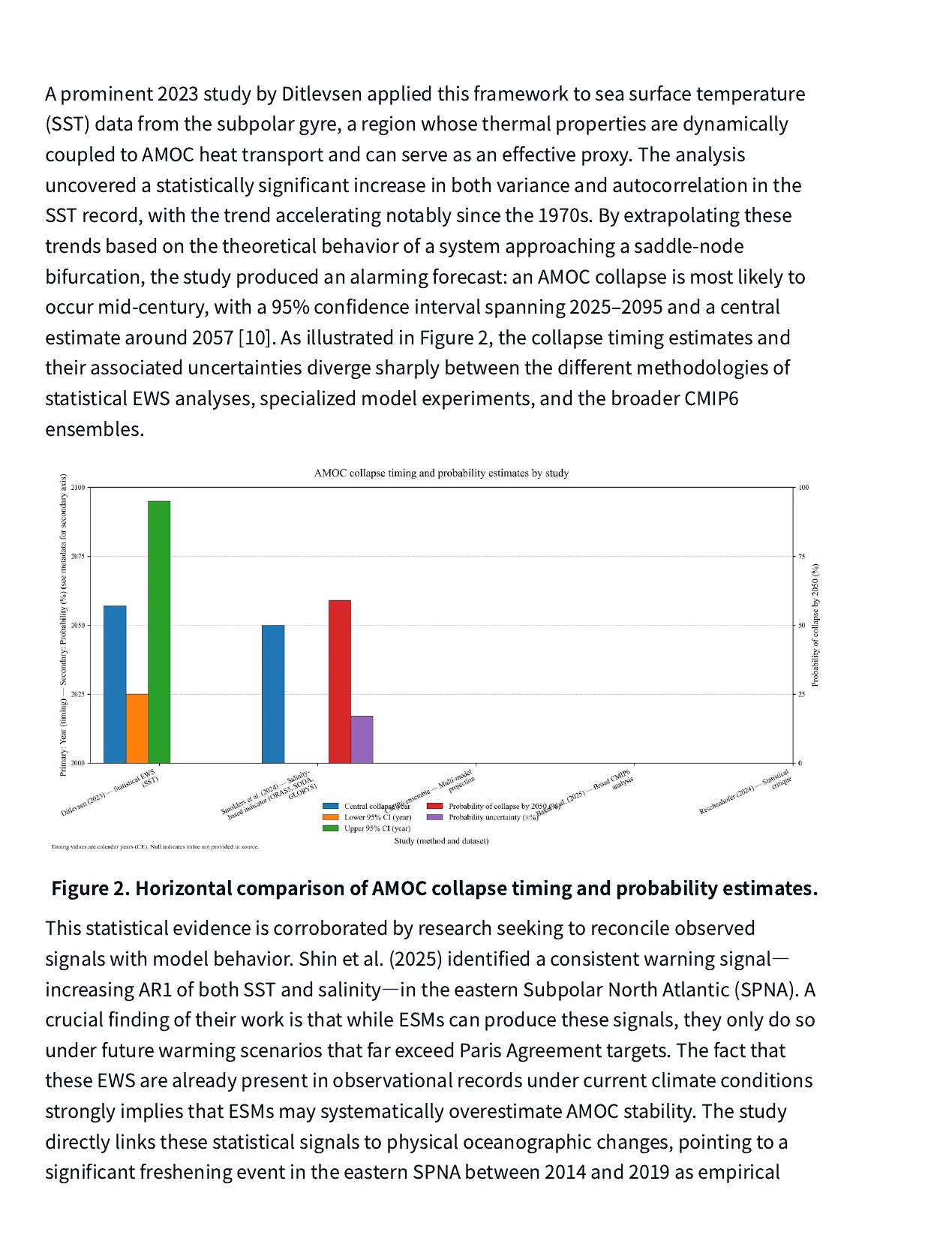} }
\end{center}
\vspace{-12pt}
\end{figure*}

\begin{figure*}[t]
\vspace{-6pt}
\begin{center}
\fbox{
\includegraphics[page=14,width=0.9\textwidth]{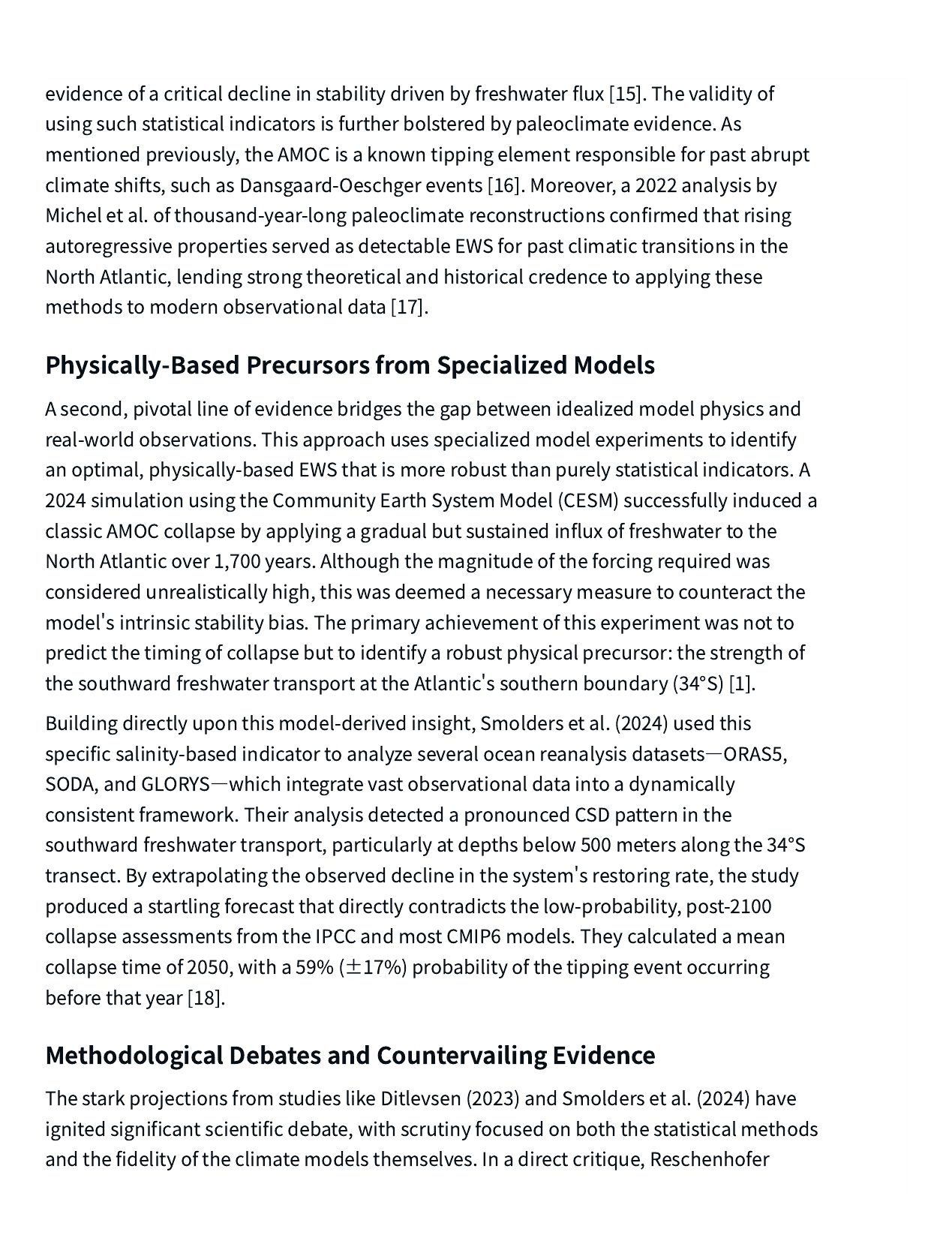} }
\end{center}
\vspace{-12pt}
\end{figure*}

\begin{figure*}[t]
\vspace{-6pt}
\begin{center}
\fbox{
\includegraphics[page=15,width=0.9\textwidth]{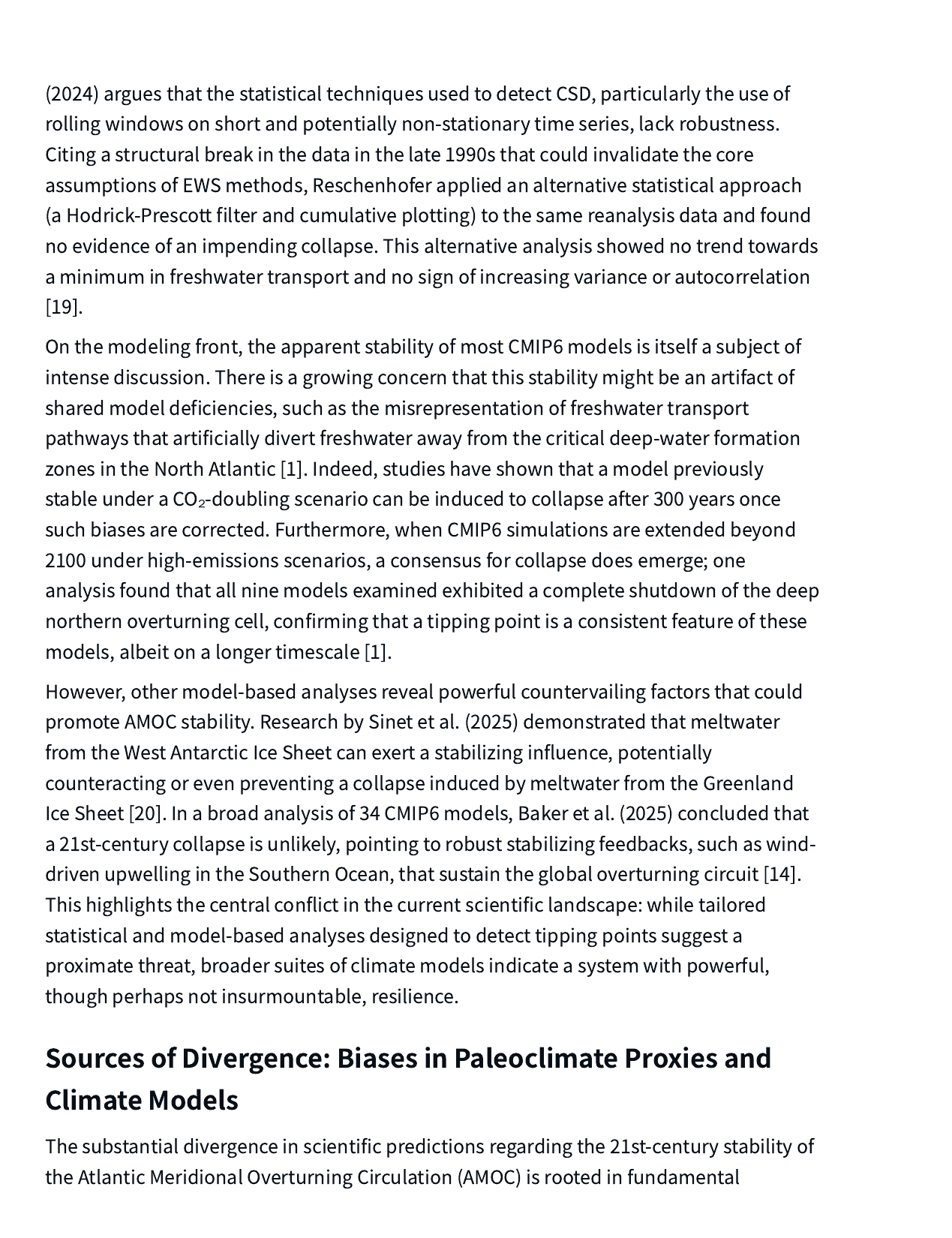} }
\end{center}
\vspace{-12pt}
\end{figure*}

\begin{figure*}[t]
\vspace{-6pt}
\begin{center}
\fbox{
\includegraphics[page=16,width=0.9\textwidth]{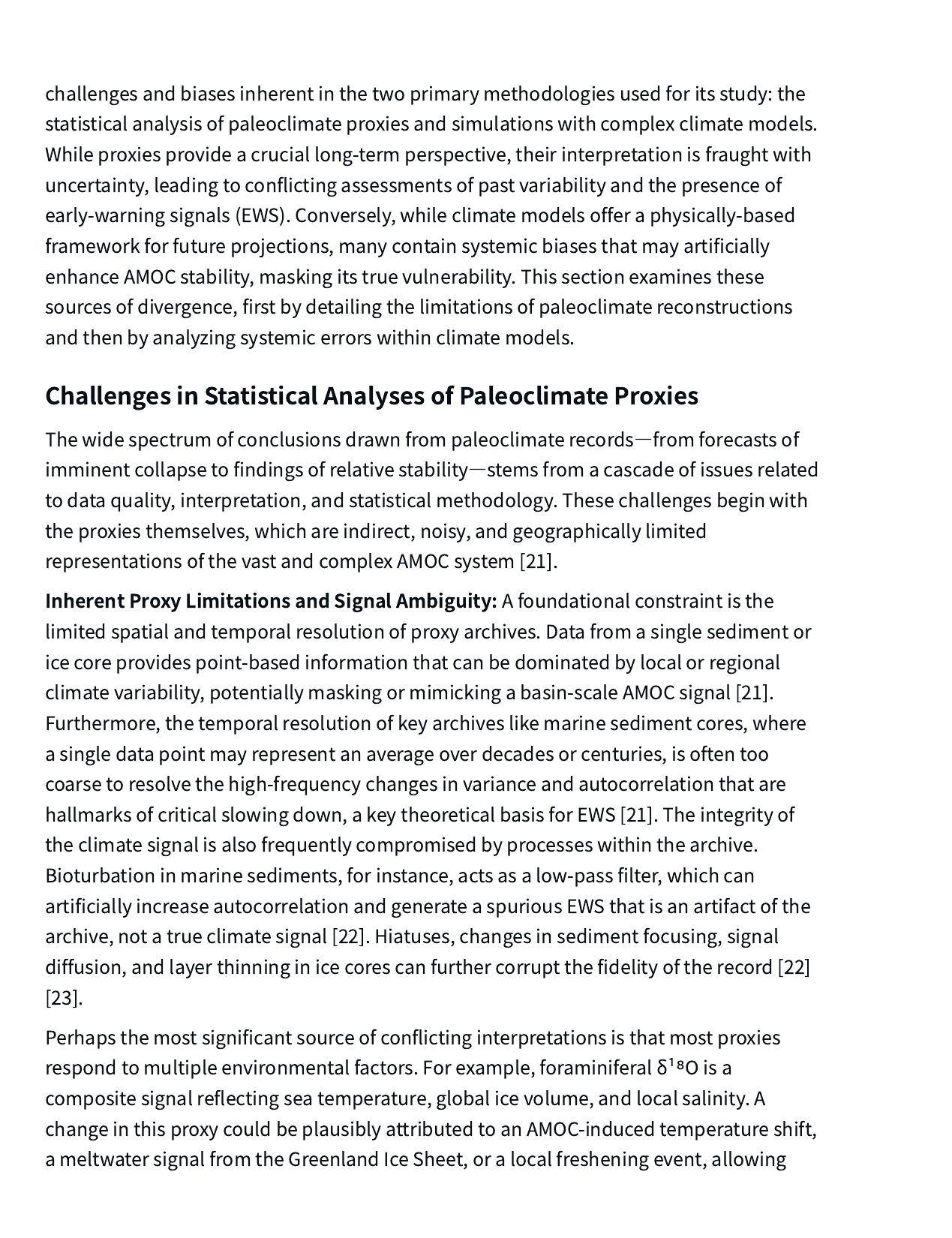} }
\end{center}
\vspace{-12pt}
\end{figure*}

\begin{figure*}[t]
\vspace{-6pt}
\begin{center}
\fbox{
\includegraphics[page=17,width=0.9\textwidth]{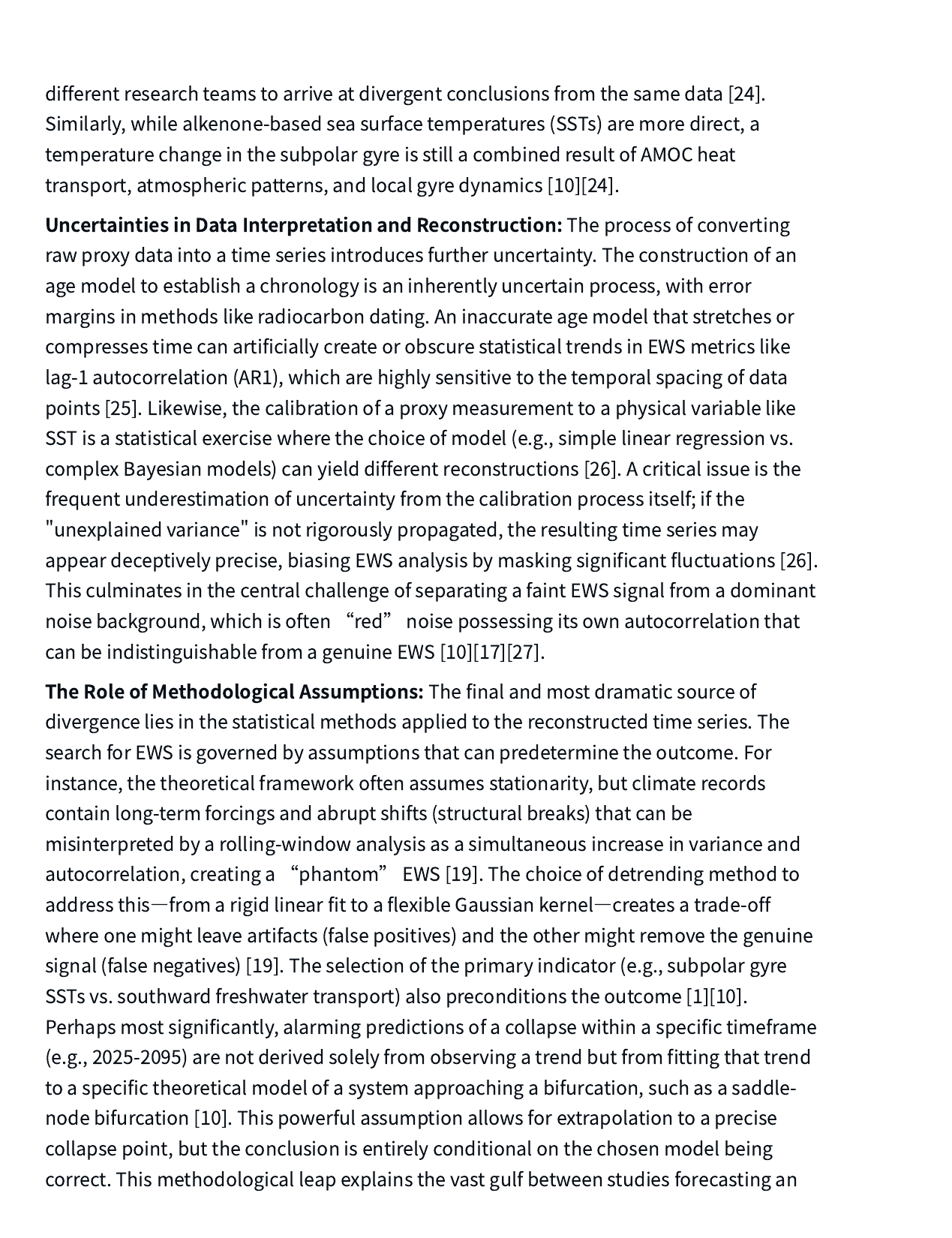} }
\end{center}
\vspace{-12pt}
\end{figure*}

\begin{figure*}[t]
\vspace{-6pt}
\begin{center}
\fbox{
\includegraphics[page=18,width=0.9\textwidth]{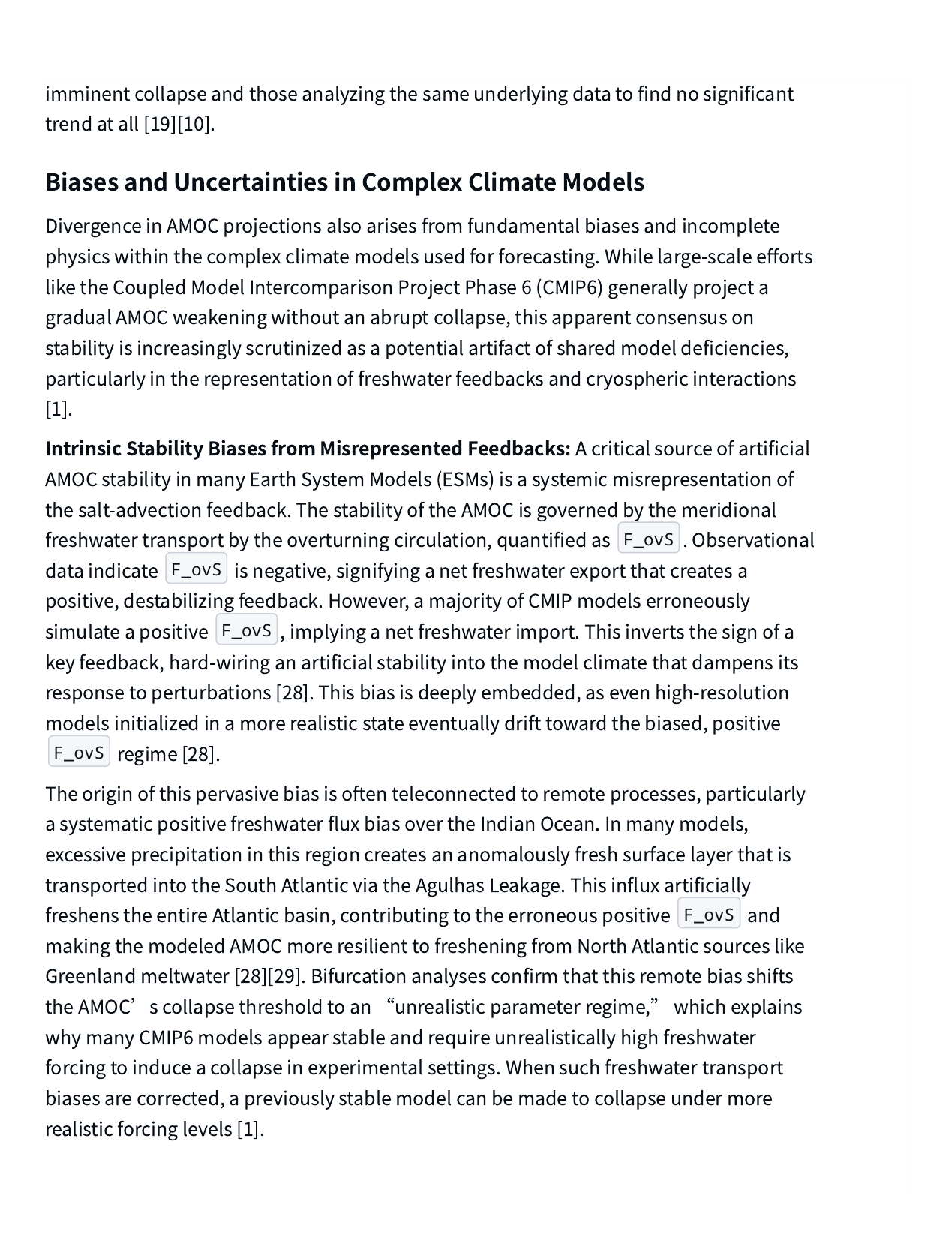}}
\end{center}
\vspace{-12pt}
\end{figure*}

\begin{figure*}[t]
\vspace{-6pt}
\begin{center}
\fbox{
\includegraphics[page=19,width=0.9\textwidth]{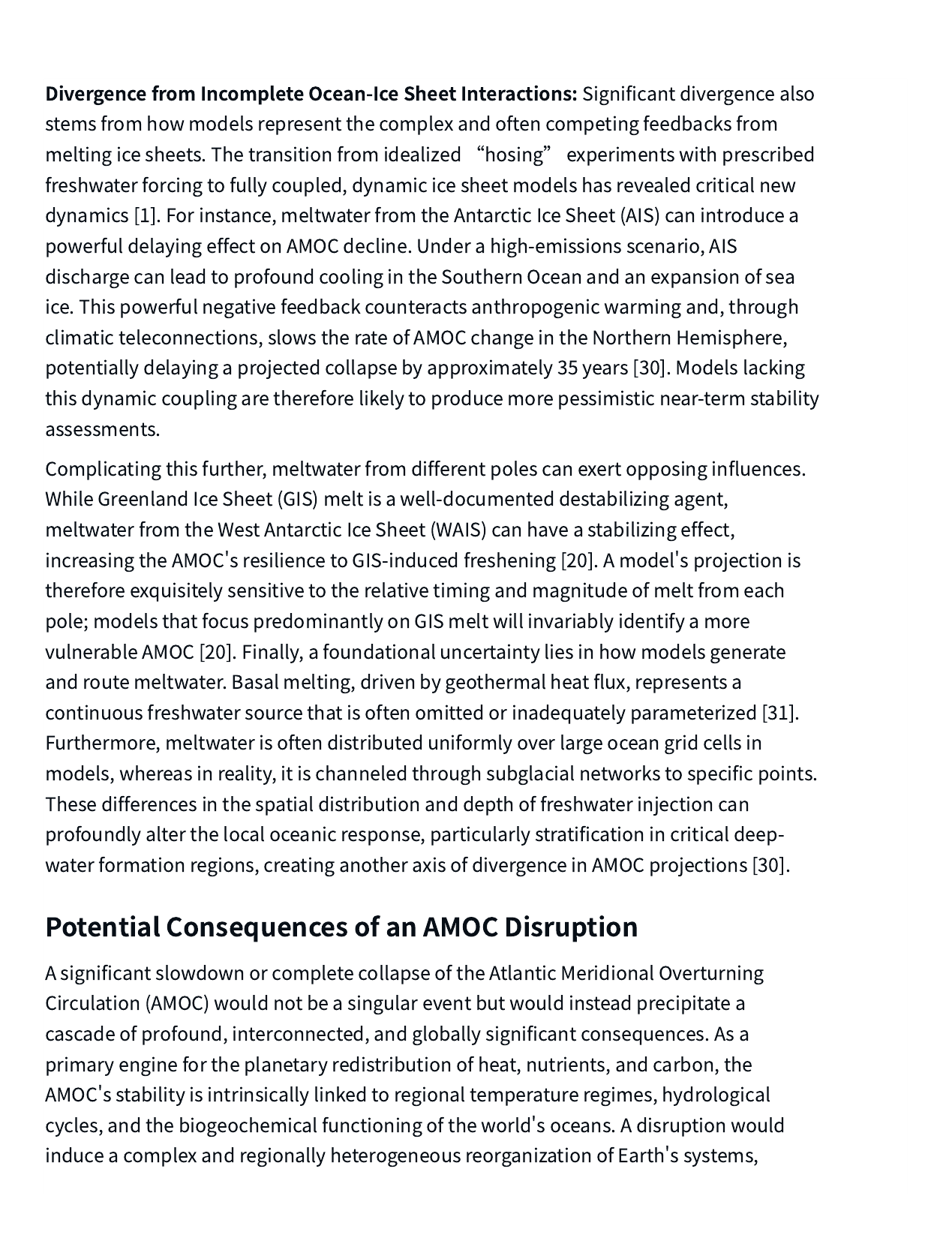} }
\end{center}
\vspace{-12pt}
\end{figure*}

\begin{figure*}[t]
\vspace{-6pt}
\begin{center}
\fbox{
\includegraphics[page=20,width=0.9\textwidth]{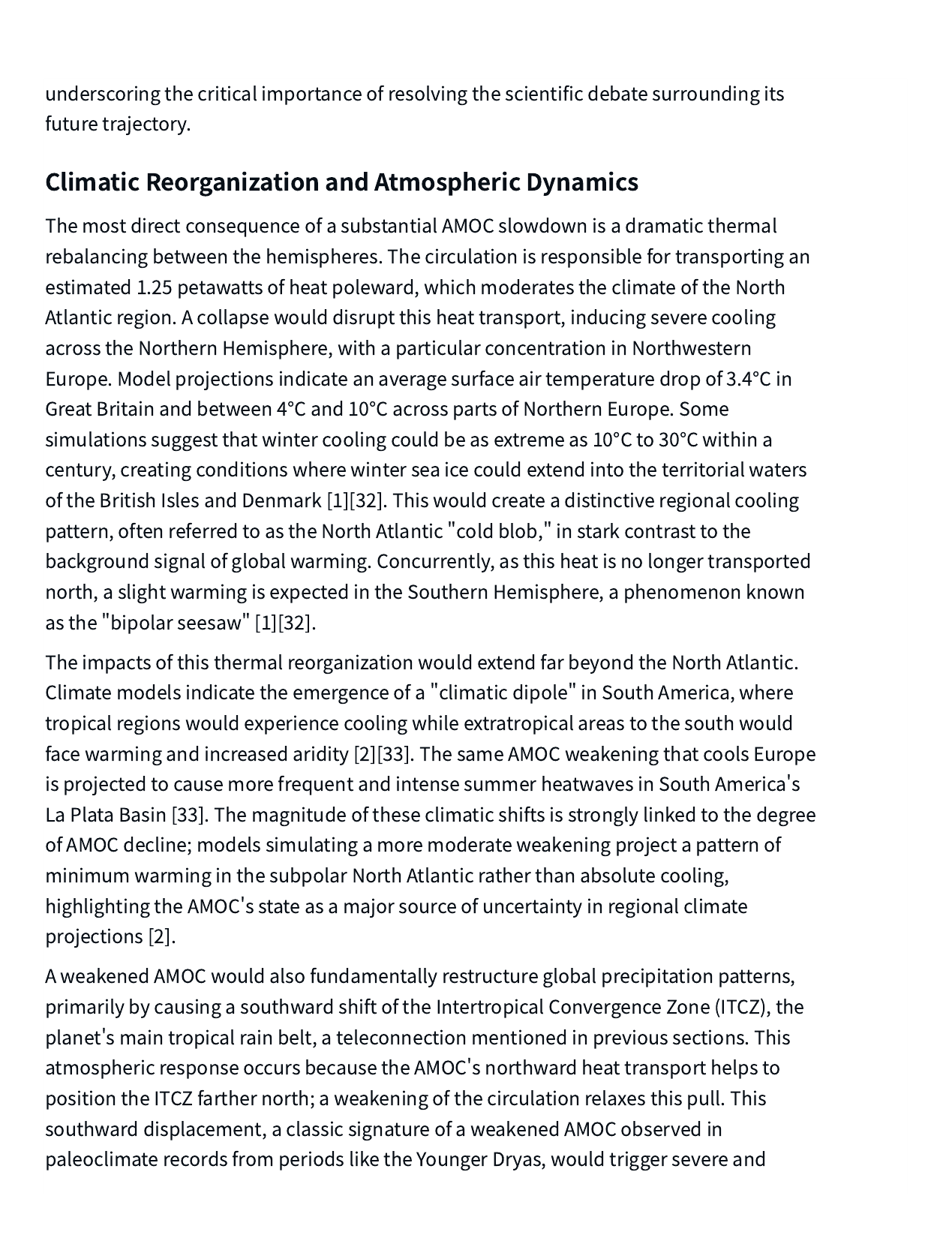} }
\end{center}
\vspace{-12pt}
\end{figure*}

\begin{figure*}[t]
\vspace{-6pt}
\begin{center}
\fbox{
\includegraphics[page=21,width=0.9\textwidth]{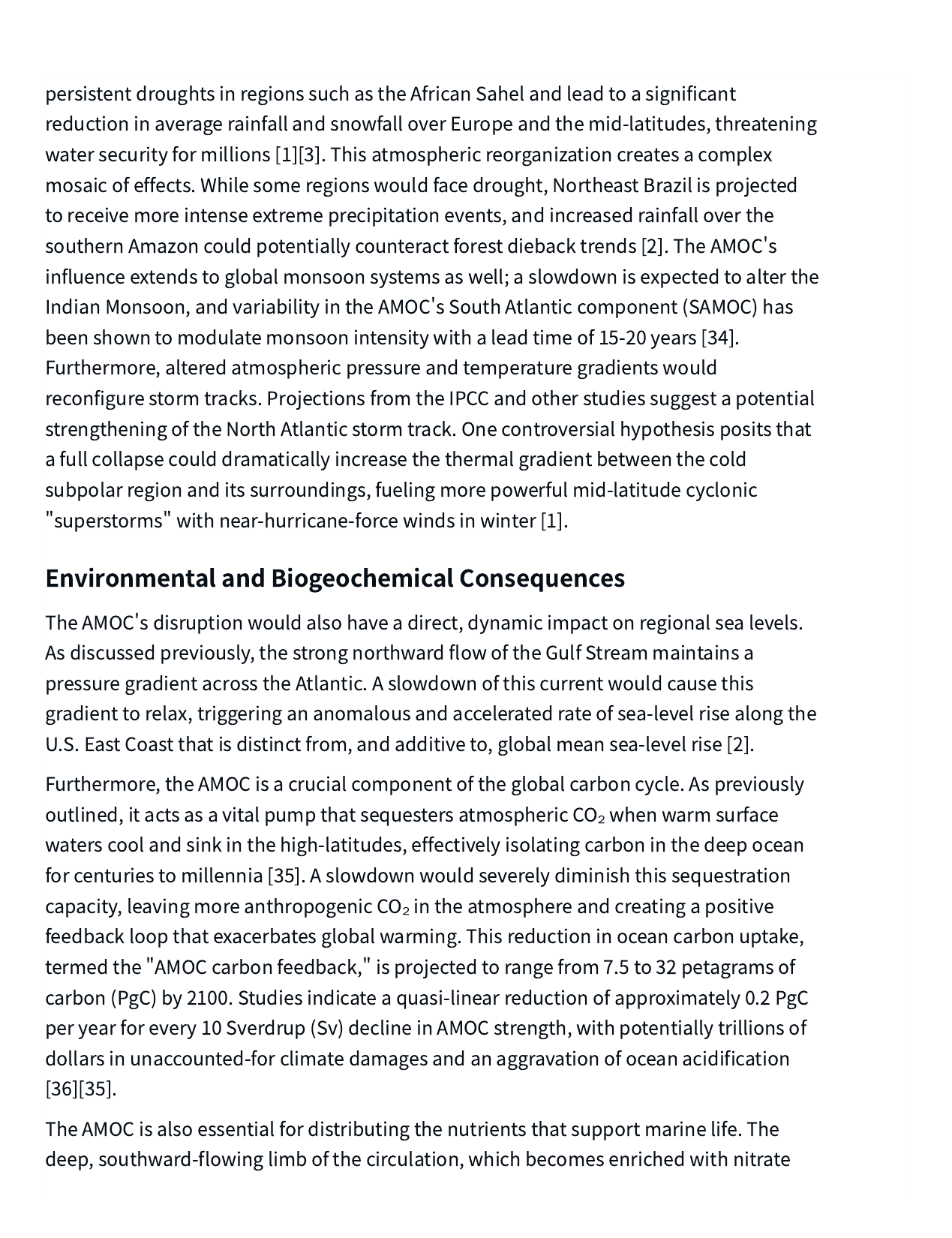} }
\end{center}
\vspace{-12pt}
\end{figure*}

\begin{figure*}[t]
\vspace{-6pt}
\begin{center}
\fbox{
\includegraphics[page=22,width=0.9\textwidth]{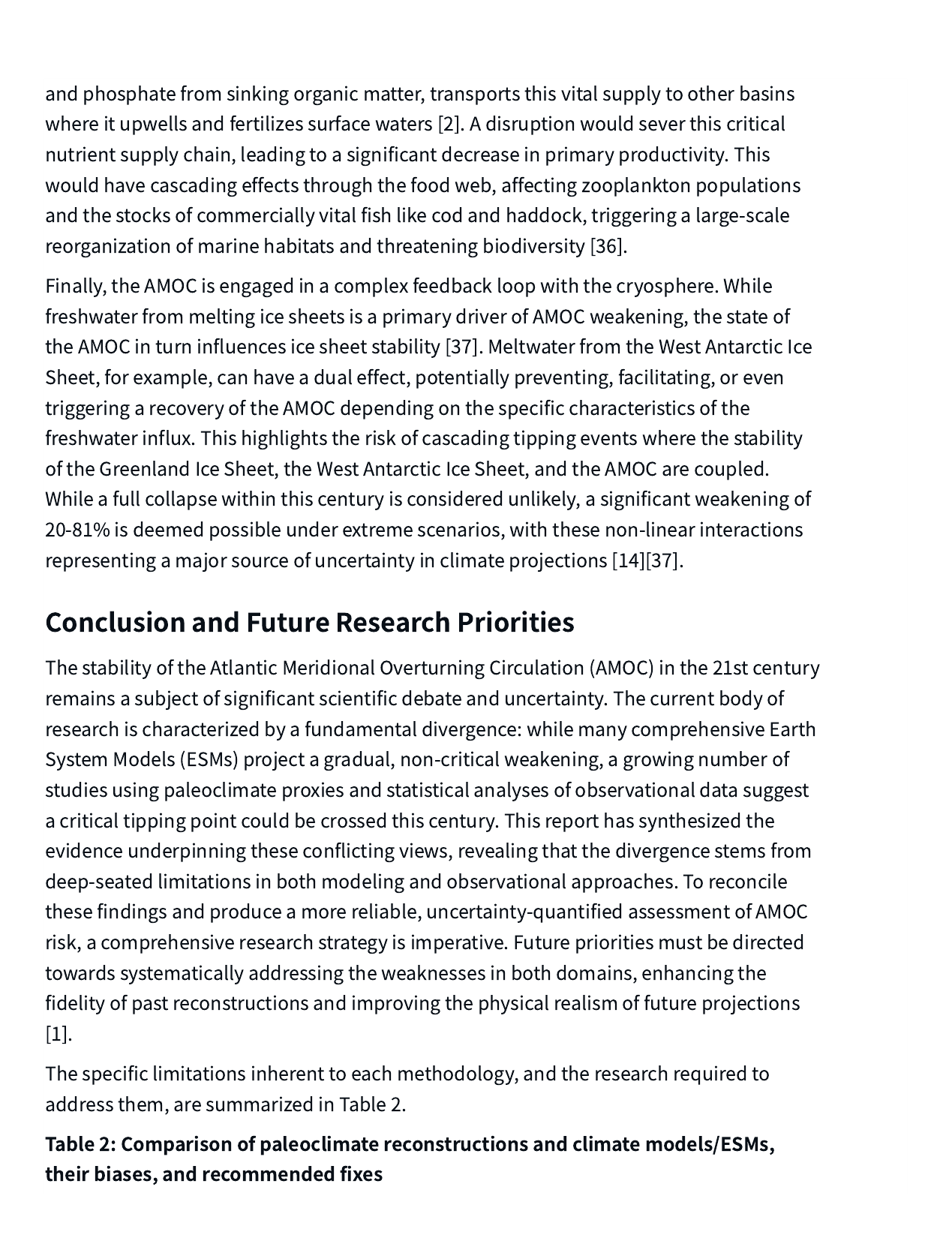} }
\end{center}
\vspace{-12pt}
\end{figure*}

\begin{figure*}[t]
\vspace{-6pt}
\begin{center}
\fbox{
\includegraphics[page=23,width=0.9\textwidth]{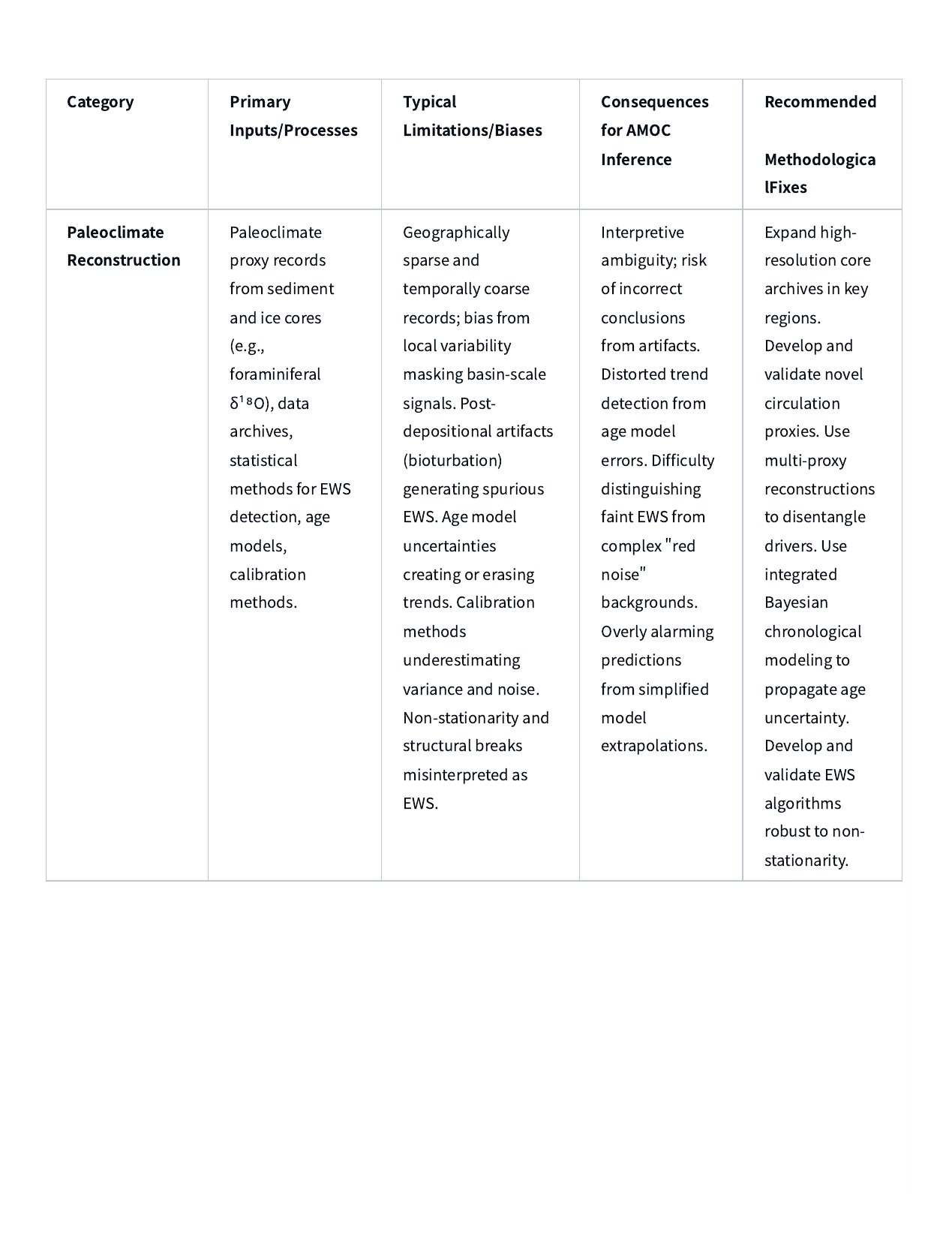} }
\end{center}
\vspace{-12pt}
\end{figure*}

\begin{figure*}[t]
\vspace{-6pt}
\begin{center}
\fbox{
\includegraphics[page=24,width=0.9\textwidth]{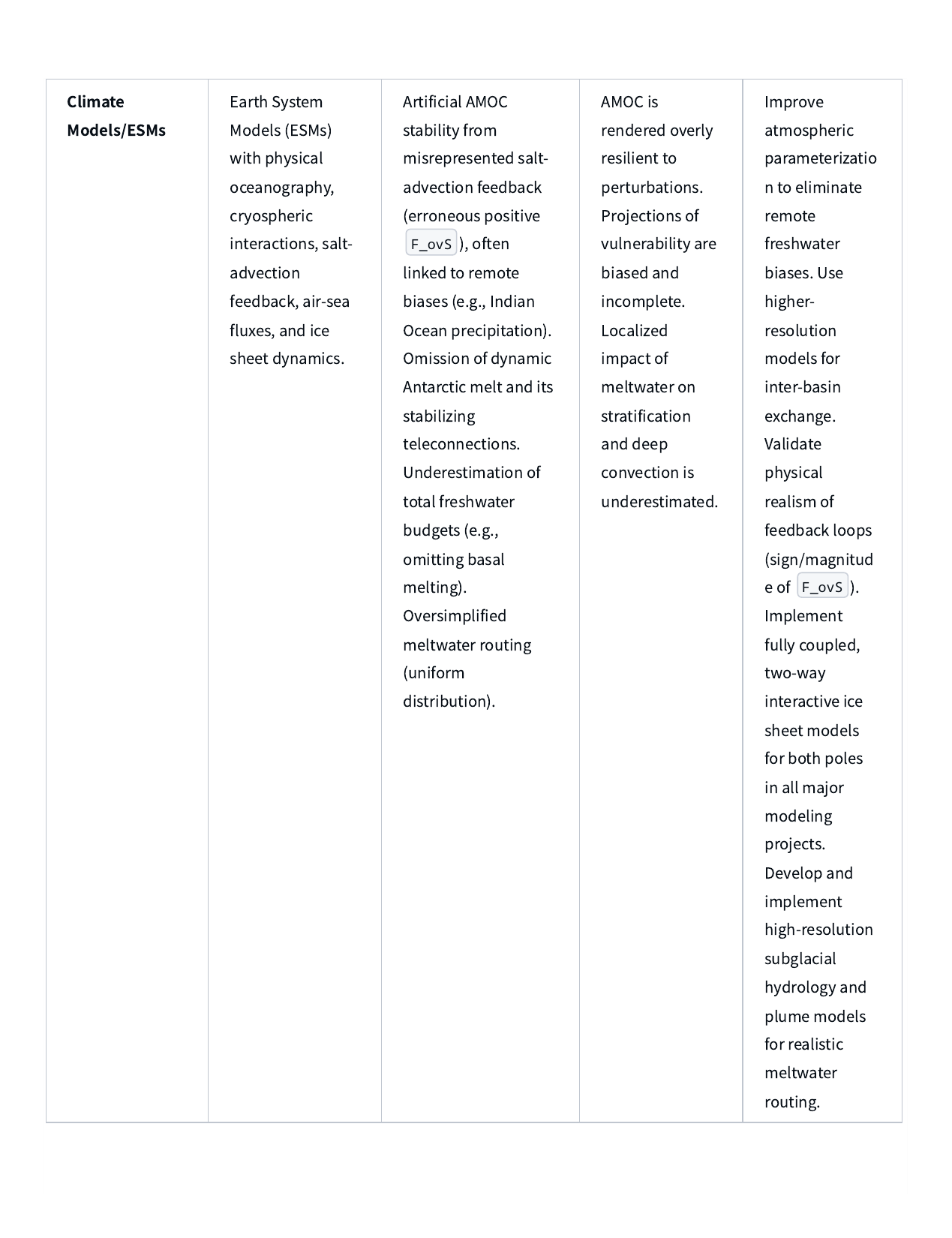} }
\end{center}
\vspace{-12pt}
\end{figure*}

\begin{figure*}[t]
\vspace{-6pt}
\begin{center}
\fbox{
\includegraphics[page=25,width=0.9\textwidth]{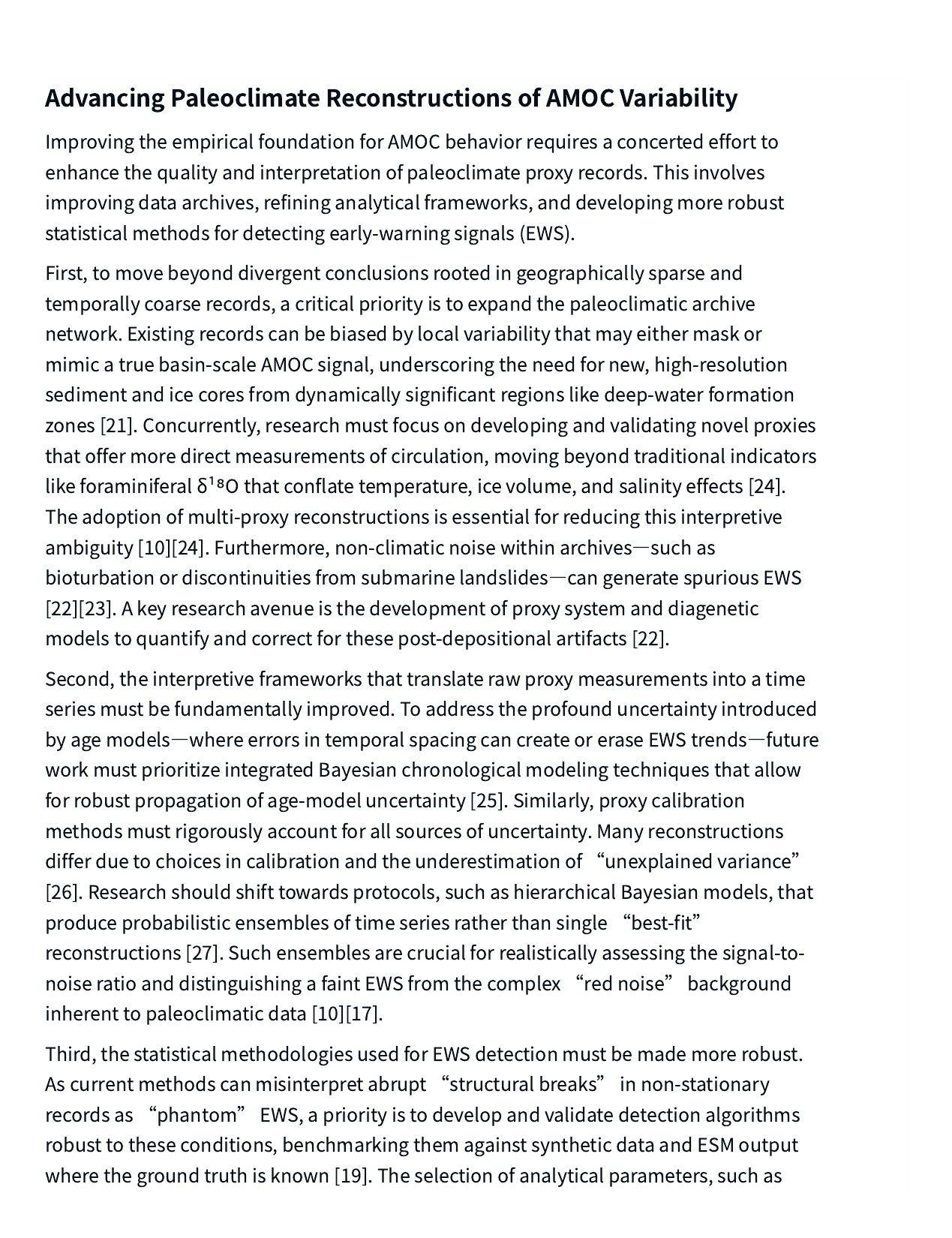} }
\end{center}
\vspace{-12pt}
\end{figure*}

\begin{figure*}[t]
\vspace{-6pt}
\begin{center}
\fbox{
\includegraphics[page=26,width=0.9\textwidth]{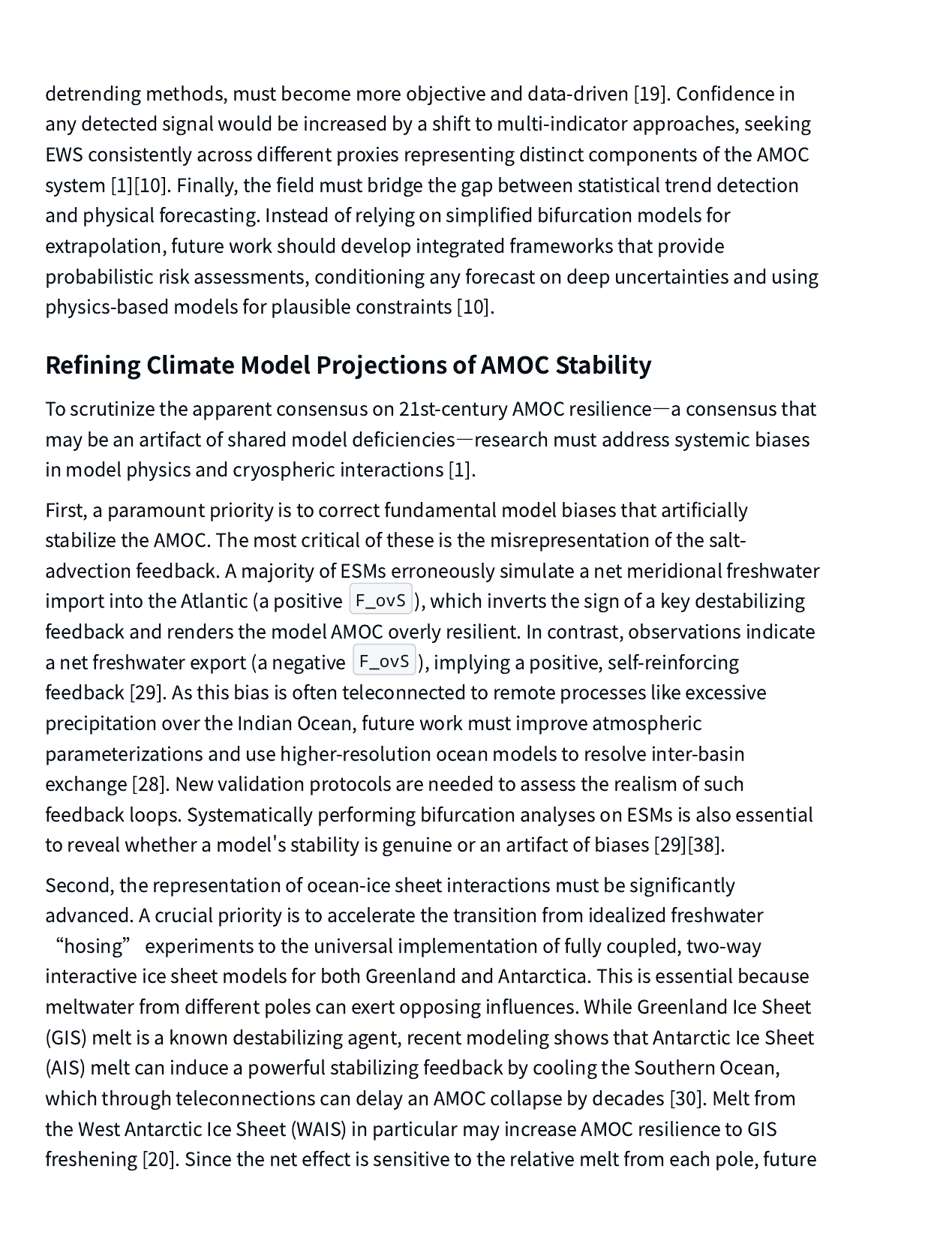} }
\end{center}
\vspace{-12pt}
\end{figure*}

\begin{figure*}[t]
\vspace{-6pt}
\begin{center}
\fbox{
\includegraphics[page=27,width=0.9\textwidth]{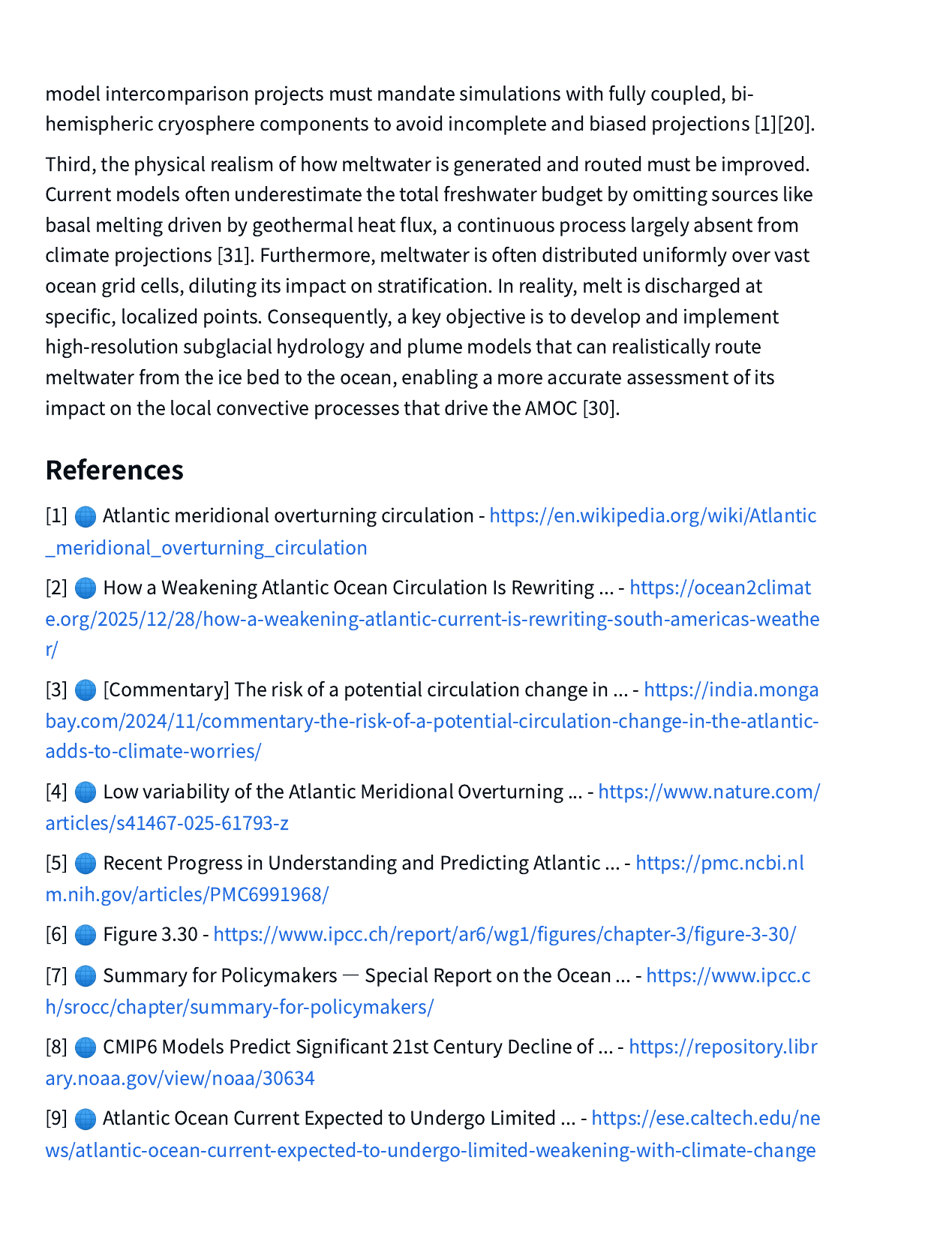} }
\end{center}
\vspace{-12pt}
\end{figure*}

\begin{figure*}[t]
\vspace{-6pt}
\begin{center}
\fbox{
\includegraphics[page=28,width=0.9\textwidth]{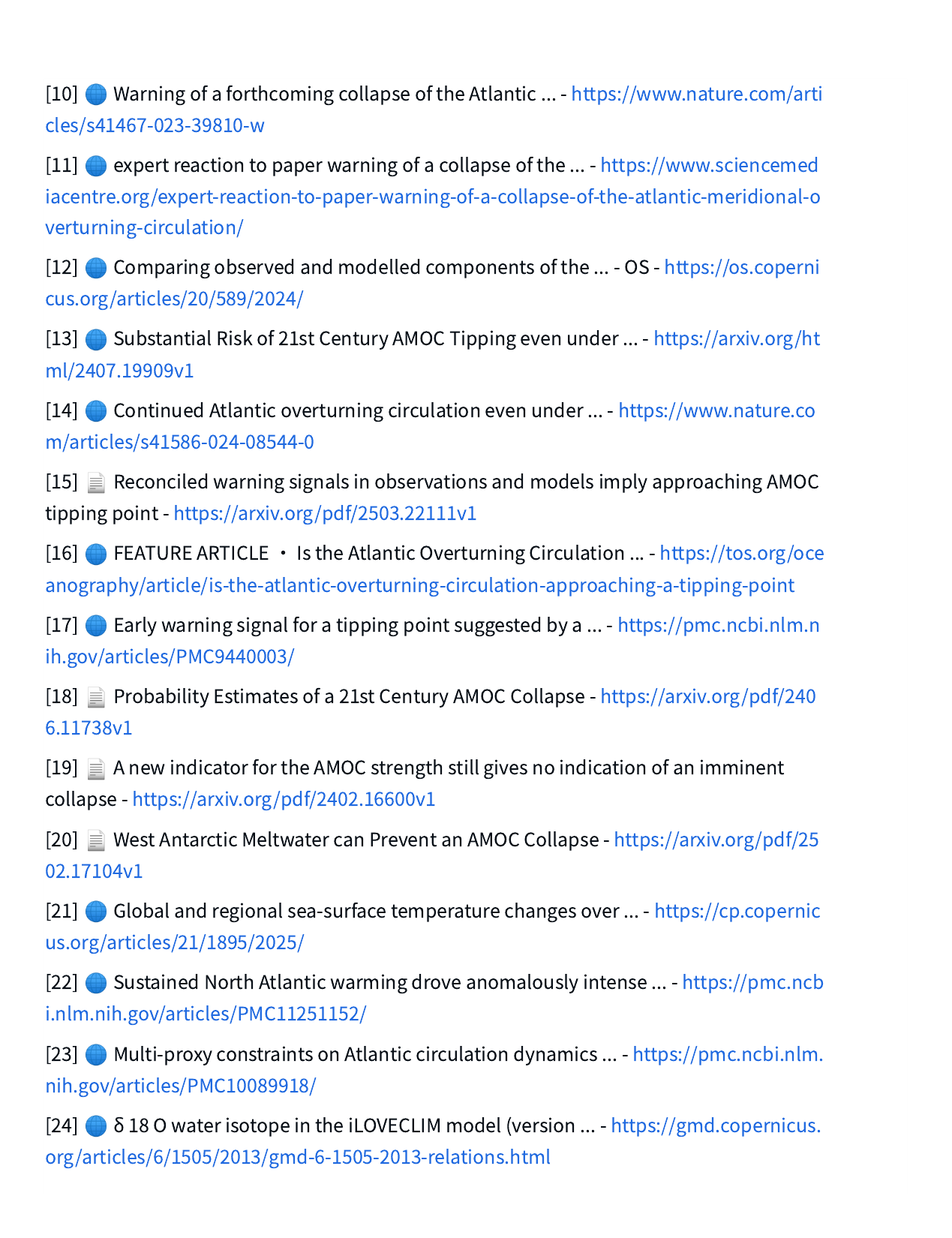} }
\end{center}
\vspace{-12pt}
\end{figure*}

\begin{figure*}[t]
\vspace{-6pt}
\begin{center}
\fbox{
\includegraphics[page=29,width=0.9\textwidth]{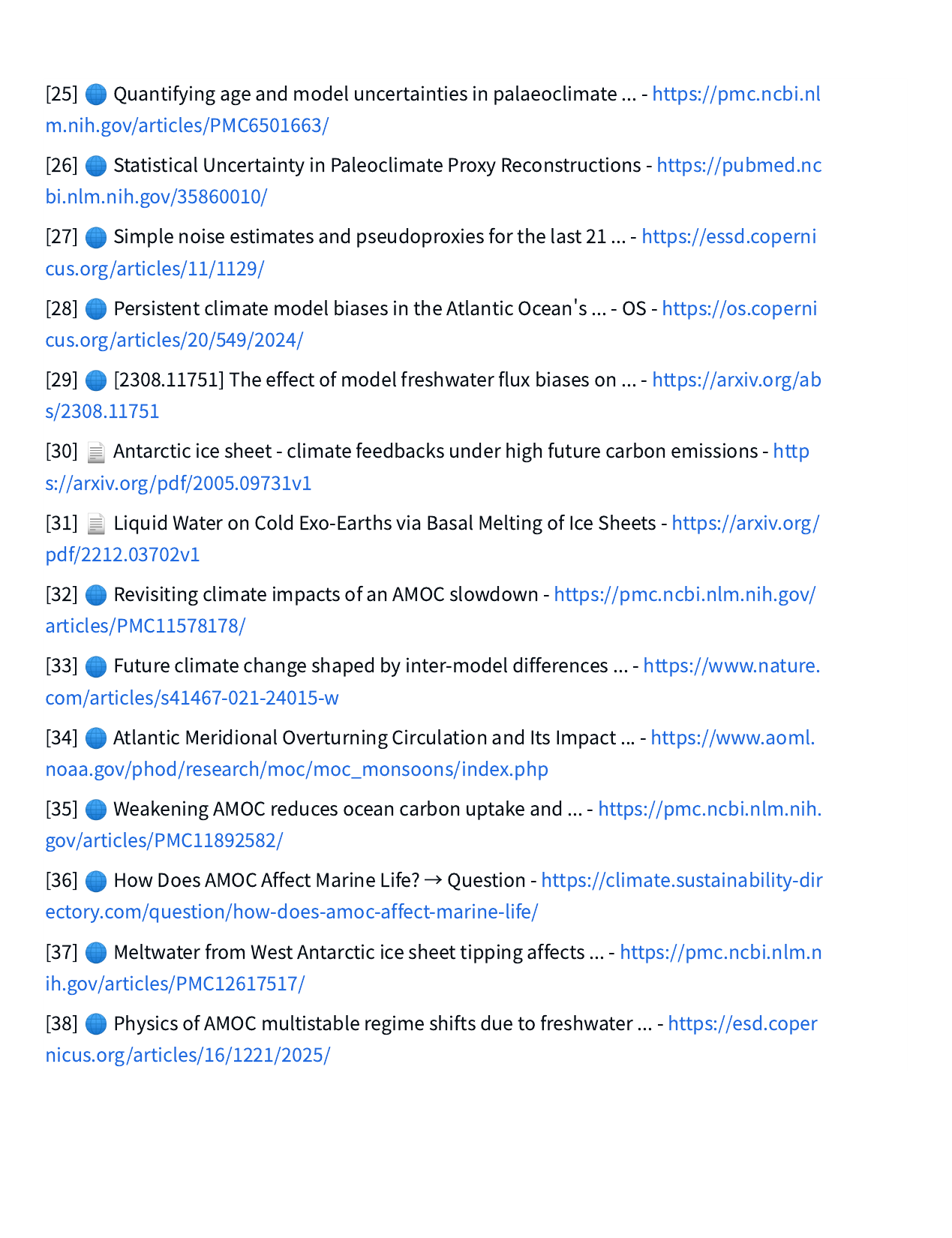} }
\end{center}
\vspace{-12pt}
\end{figure*}